\documentclass[11pt]{article}

\usepackage{arxivtemplate}

\usepackage{tikz,tcolorbox,float,tabularx,multirow,subcaption}

\theoremstyle{plain}
\newtheorem{fact}[theorem]{Fact}
\theoremstyle{remark}

\newcommand{\R}{\mathbb R}
\newcommand{\E}{\mathbb E}
\newcommand{\Prob}{\mathbb P}
\newcommand{\tr}{\operatorname{tr}}

\newcommand{\rank}{\operatorname{rank}}
\newcommand{\Cov}{\operatorname{Cov}}
\newcommand{\Var}{\operatorname{Var}}
\newcommand{\ER}{\operatorname{erank}}
\newcommand{\Rot}{\operatorname{rot}}
\newcommand{\op}{\mathrm{op}}

\title{Spectral and Subspace Dynamics of LoRA in Supervised Fine-Tuning and Reinforcement Learning}

\author{%
Chenruo Liu$^{1}$\thanks{Equal contribution.}\qquad
Zihan Wang$^{1}$\footnotemark[1]\qquad
Yijun Dong$^{2}$\qquad
Qi Lei$^{1}$\\[0.8ex]
$^{1}$New York University\\
$^{2}$Institute of Science and Technology Austria
}
\date{}

\begin{document}
\maketitle

\begin{abstract}
Low-rank adaptation (LoRA) is commonly used in both supervised fine-tuning (SFT) and reinforcement learning (RL), often under similar optimizer settings. Yet, how the different dynamics of the two stages manifest in the low-rank increment that LoRA optimizes remains unclear. 
  Tracking the singular spectrum and subspaces of LoRA updates on mathematical reasoning tasks, we find that RL concentrates the spectrum at every learning rate, while SFT expands it at moderate learning rates; moreover, RL rotates the singular subspaces much faster than SFT at small learning rates. Interpolating between SFT and RL one ingredient at a time, we pinpoint group-mean baseline subtraction as the source of both contrasts.
  A quadratic multi-index model updated via LoRA unveils the underlying mechanism: group-mean baseline subtraction tilts the balance between two competing forces, coherent fitting and gradient noise, toward spectral concentration and faster subspace rotation, while forcing the risk-favorable learning rate to be low.
  Guided by these dynamics, we tailor LoRA updates to each stage: in SFT, we flatten the spectrum of the full update to spread adaptation across singular directions, whereas in RL, we flatten the part of the update outside the current LoRA increment span to promote subspace rotation.
  Experiments on mathematical reasoning tasks across multiple models show that the respective configurations for SFT and RL yield the best performance among the compared methods in most evaluated settings.

\end{abstract}

\section{Introduction}
\label{sec:intro}

Post-training is an essential step in modern deep learning that adapts a powerful pre-trained model with general knowledge to specialized downstream tasks. 
A common post-training pipeline consists of two stages: supervised fine-tuning (SFT), which first equips the pre-trained model with primary domain knowledge through off-policy expert demonstrations, followed by reinforcement learning (RL), which further improves the model through its own on-policy responses guided by reward signals like human feedback~\citep{ouyang2022training,schulman2017proximal,rafailov2023direct} or verifiable correctness checks~\citep{shao2024deepseekmath,yu2025dapo}.
During post-training, parameter-efficient fine-tuning methods like low-rank adaptation (LoRA)~\citep{hu2021lora} are commonly incorporated to reduce the memory and compute footprint.
Specifically for LoRA, a frozen pretrained weight $W_0$ is adapted by a low-rank update $\Delta W_t=\frac{\alpha}{r}B_tA_t$, with trainable factors $B_t\in\mathbb{R}^{d_o\times r}$ and $A_t\in\mathbb{R}^{r\times d_i}$ of rank $r\ll\min\{d_o,d_i\}$, scaled by $\alpha$.
In practice, the LoRA factors $B_t, A_t$ are often trained by the same optimizer like Adam~\citep{kingma2014adam} under nearly identical settings in both SFT and RL.

However, SFT and RL demonstrate distinct behaviors during post-training, suggesting different dynamics that can be benefited from tailored optimization strategies.
For example, compared to SFT, RL tends to generalize better out of distribution~\citep{chu2025sft,kirk2024understanding}, forget less~\citep{chen2025retaining,shenfeld2026rl}, and update sparser subnetworks~\citep{mukherjee2026reinforcement}.
RL is also usually run with markedly smaller learning rates compared to SFT~\citep{touvron2023llama,shao2024deepseekmath}.
While numerous differences between SFT and RL have been observed through model outputs, hidden representations, and full weight matrices, in the context of LoRA, much less is known about how such differences manifest in the low-rank increment $\Delta W_t$ that LoRA actually optimizes.
Filling this gap is a prerequisite for tailoring LoRA optimization to the distinct dynamics of SFT and RL.

We investigate the dynamics of $\Delta W_t$ through two complementary lenses: the spread of singular values measured by the effective rank~\citep{roy2007effective} that determine how the knowledge required for the downstream task is distributed over the $r$ available directions, and the rotation of the left and right singular subspaces that determine which input and output directions are used.
We ask through these lenses:
\begin{tcolorbox}
    \emph{How do SFT and RL differ in the spectrum and singular subspaces of $\Delta W_t$; what is the mechanism that drives these differences; and how can LoRA optimization be tailored to each stage?}
\end{tcolorbox}

\subsection{Our Contributions}
\paragraph{Baseline subtraction separates the LoRA dynamics of SFT and RL (Section~\ref{sec:exp1}).}
Across various Llama and Qwen models on GSM8K~\citep{cobbe2021training} and MATH~\citep{hendrycks2021measuring}, we observed that RL with verifiable reward via group relative policy optimization (GRPO)~\citep{shao2024deepseekmath} concentrates the spectrum of $\Delta W_t$ at every learning rate; whereas SFT expands it at moderate learning rates that can facilitate desired downstream performance.
For the dynamics of the singular subspaces, we observed that RL with small learning rates rotates the singular subspace much faster than SFT.
Moreover, interpolating between SFT and RL one ingredient at a time, we find that both contrasts appear only once a group-mean baseline is subtracted from the reward, pinpointing baseline subtraction as the key mechanism that separates the LoRA dynamics of SFT and RL.

\paragraph{Baseline subtraction tilts the balance between coherent fitting and gradient noise (Section~\ref{sec:thm}).}
To understand these contrasts, we analyze a quadratic multi-index model in which a LoRA adapter is trained by stochastic gradient descent (SGD) on pairs of correlated inputs, each pair modeling a GRPO group of responses sampled for the same prompt.
Two forces compete in this model: coherent fitting that is linear in the learning rate and concentrates spectral energy on strongly responding directions, as well as feedback from gradient noise that is quadratic in the learning rate and redistributes the spectral energy more evenly.
Subtracting a group-mean baseline largely cancels the gradient component shared across a group's responses, which drives coherent fitting along pretrained directions and anchors successive updates to similar directions, while retaining most of the component that contrasts responses within the group, and damping the gradient noise from task-irrelevant directions. 
Coherent fitting thus persists through the contrasts while the redistributing feedback loses relative strength, and gradient noise, no longer anchored by the shared component, steers the update directions more, leading to spectral concentration and faster singular subspace rotation during RL.
Nevertheless, baseline subtraction can come at a cost for prediction: coherent fitting of the shared component reduces the risk more efficiently than fitting of the contrasts, cancellation of which shrinks the risk reduction from coherent fitting (linear in the learning rate) relative to the cost of gradient noise (quadratic in the learning rate).
The risk-favorable learning rate thus remains low, consistent with the common practice in RL.

\paragraph{The dynamics inform stage-specific LoRA optimization (Section~\ref{sec:exp2}).} 
Our analysis further shows which dynamics accompany the risk-favorable learning rate: training without a baseline (as in SFT) expands the spectrum of $\Delta W_t$, whereas training with a baseline (as in RL) concentrates the spectrum and rotates the left singular subspace faster.
Guided by these dynamics, we tailor the update of LoRA factors to each stage by combining two standard components, a norm-preserving spectral flattening as in Muon~\citep{jordan2024muon} and a per-factor choice between SGD and Adam update directions.
In SFT, we flatten the singular spectrum of the full update of the factor so that the adaptation is distributed more evenly across singular directions; whereas in RL, we flatten only the component of the update orthogonal to the current left singular subspace of $\Delta W_t$, which promotes subspace rotation while leaving the within-subspace component unchanged.
The resulting configurations achieve dominating downstream performance across most evaluated models and tasks.

\subsection{Related Work}
\label{sec:related_work}

\paragraph{Differences between supervised and reinforcement fine-tuning.} Prior work has characterized differences between SFT and RL at the output, representation, and parameter levels. At the output level, empirical studies report stronger out-of-distribution performance under RL than SFT~\citep{kirk2024understanding, chu2025sft}.
Subsequent work further highlights the dependence of SFT generalization on optimization, data quality, and model capability~\citep{ren2026rethinking}. Beyond performance, studies examine the diversity of generated outputs: RLHF reduces overall response diversity relative to SFT~\citep{kirk2024understanding}, while analyses distinguishing correct and incorrect reasoning trajectories find that SFT expands the former and RL compresses the latter~\citep{matsutani2026rl}. Moreover, a line of work examines how changes in the model's output distribution during fine-tuning relate to the preservation of pretrained capabilities. These studies illustrate how on-policy RL can better preserve distributional modes associated with prior knowledge than SFT~\citep{chen2025retaining}, and link RL's better retention to an implicit preference for output distributions closer to that of the initial model in KL divergence~\citep{shenfeld2026rl}.

At the hidden representation level, \citet{shi2026does} find that SFT rapidly introduces specialized features that stabilize early, whereas RL induces more restrained but continually evolving feature changes.

At the parameter update level, studies report greater sparsity in parameter updates under RL than SFT~\citep{mukherjee2026reinforcement}. Comparisons of update vectors across tasks reveal large absolute cosine similarities under SFT, whereas RL updates are nearly orthogonal~\citep{zhu2026sft}. Spectral analyses of the weight matrices further show smaller changes from the pretrained spectra and principal subspaces under RLVR than SFT~\citep{zhu2025path}. Our work also compares spectral and subspace dynamics under SFT and RL. In contrast, we focus on the LoRA update matrix itself, track how its spectral concentration and subspace rotation vary with learning rate and training step, interpolate between SFT and RL through intermediate training settings to identify where these differences arise, and use these findings to guide the choice of optimizer configurations for LoRA in each paradigm.

\paragraph{Spectral and subspace analyses for LoRA optimization.}
Prior work uses spectral and subspace analyses to study LoRA and improve it through subspace selection and the allocation of singular values.
From a subspace perspective, PiSSA initializes adapters with principal singular components of pretrained weights and reports faster convergence and better task performance than standard LoRA~\citep{meng2024pissa}.
MiLoRA instead initializes adapters with minor singular components to preserve pretrained knowledge, and its analysis shows less forgetting than PiSSA~\citep{wang2025milora}.
Beyond using pretrained weights alone, further work incorporates downstream task information to guide the initialization of LoRA subspaces.
LoRA-One initializes LoRA factors to align with the principal singular subspaces of the initial full gradient~\citep{zhang2025lora}.
\citet{xu2025understanding} study how LoRA factors align with task-relevant directions in a matrix factorization setting and propose a spectral initialization informed by both pretrained weights and the downstream task.
Moving from initialization to training, SDC-LoRA finds that LoRA updates can drift toward pretrained principal singular subspaces even with MiLoRA's minor-subspace initialization.
It further reduces forgetting by downscaling the gradient components aligned with these principal subspaces during training~\citep{zhang2026sdc}.

Another line of work examines the allocation of singular values in LoRA update matrices.
\citet{kang2026uniform} apply Muon to LoRA and observe more uniform growth of singular values, with effective ranks close to the nominal LoRA rank.
Spectral Surgery finds that some singular components of trained LoRA update matrices contribute little to task performance or are even detrimental.
It therefore reweights their singular values after training while keeping the learned singular directions fixed to improve task performance~\citep{tian2026spectral}.

While the above studies address general LoRA optimization and are primarily evaluated in supervised settings, recent work develops methods specifically for RLVR.
GeoRA argues that PiSSA's emphasis on pretrained principal directions can conflict with RLVR's need to preserve pretrained structure.
It therefore uses spectral and weight-magnitude information to construct a masked weight matrix and initializes LoRA adapters from its principal singular components~\citep{zhang2026geora}.
Similarly, \citet{zhang2026geometry} find that PiSSA and MiLoRA can become unstable under RLVR despite their benefits in SFT.
They link this instability to singular-value scaling at initialization and propose an orthonormal initialization that avoids this scaling. In contrast, we compare spectral and subspace dynamics under SFT and RL, and use the observed differences to guide the choice of optimizer configurations for LoRA during training in each setting.

\section{Empirical Characterization of LoRA Dynamics}
\label{sec:exp1}

\subsection{Experimental Setup}
\label{sec:exp1_setup}

\paragraph{Spectral and Subspace Diagnostics of LoRA.}
For a pretrained weight matrix
$W_0 \in \mathbb{R}^{d_o \times d_i}$,
LoRA~\citep{hu2021lora} parameterizes the adapted weight at
training step $t$ as
\[
    W_t = W_0 + \Delta W_t,
    \qquad
    \Delta W_t = \frac{\alpha}{r} B_t A_t,
\]
where $B_t \in \mathbb{R}^{d_o \times r}$ and
$A_t \in \mathbb{R}^{r \times d_i}$ are trainable,
$W_0$ remains frozen,
$r \ll \min\{d_o,d_i\}$ is the adapter rank,
and $\alpha$ controls the scaling.
For $t>0$, we write the singular value decomposition
of the LoRA increment as
$\Delta W_t = U_t \Sigma_t V_t^\top$,
where
$\Sigma_t = \operatorname{diag}(\sigma_{1,t},\ldots,\sigma_{r,t})$
with $\sigma_{1,t} \geq \cdots \geq \sigma_{r,t} \geq 0$, and the matrices
$U_t \in \mathbb{R}^{d_o \times r}$ and
$V_t \in \mathbb{R}^{d_i \times r}$
contain the corresponding orthonormal left and right
singular vectors, respectively.

We track both the singular values and singular subspaces
of the LoRA increment throughout training.
To characterize the singular value distribution of \(\Delta W_t\),
we apply the effective rank construction
of~\citet{roy2007effective} to the normalized squared singular values:
\[
\operatorname{erank}(t)
= \exp\left(
-\sum_{i=1}^{r}
\frac{\sigma_{i,t}^{2}}{\sum_{j=1}^{r}\sigma_{j,t}^{2}}
\log\frac{\sigma_{i,t}^{2}}{\sum_{j=1}^{r}\sigma_{j,t}^{2}}
\right)\in [1,r],
\]
where smaller values indicate a more uneven distribution of
$\sigma_{i,t}^2$, i.e., stronger concentration in a few dominant
singular directions.
Moreover, to characterize subspace rotation between training steps
$t_1$ and $t_2$, we measure the overlaps of their left and right singular subspaces:
\[
\begin{aligned}
    \operatorname{ovlp}_{L}(t_1,t_2)
    &= \frac{1}{r}
    \left\|U_{t_1}^{\top}U_{t_2}\right\|_{F}^{2} \in [0,1], \\
    \operatorname{ovlp}_{R}(t_1,t_2)
    &= \frac{1}{r}
    \left\|V_{t_1}^{\top}V_{t_2}\right\|_{F}^{2}\in [0,1],
\end{aligned}
\]
where, over the same training interval, smaller overlap values
indicate faster subspace rotation.

\paragraph{Interpolating between SFT and RL.}
Throughout this work, we use GRPO~\citep{shao2024deepseekmath}
as a representative RL algorithm.
To identify the sources of differences in effective rank
and subspace overlap between SFT and RL, we consider a ladder of four
training settings that interpolate between the two:
(i) SFT, which performs supervised training on gold completions;
(ii) Rejection Sampling Fine-Tuning (RFT), which replaces
gold completions with model-generated completions and trains
only on correct ones, corresponding to training weights
$r_i$ for binary rewards $r_i \in \{0,1\}$;
(iii) RFT with group-mean baseline subtraction (RFT-B), which
introduces baseline subtraction into RFT, changing the
weights from $r_i$ to
$r_i-\bar r$, where $\bar r$ is the mean reward over all
completions for the same prompt;
and (iv) GRPO, which extends RFT-B to online sampling and additionally
incorporates GRPO optimization mechanisms, including
policy ratio clipping and fixed-length loss normalization
following Dr.~GRPO~\citep{liu2025understanding},
without KL regularization.

\paragraph{Training and Evaluation Protocol.}
Our model suite includes the base and instruction-tuned
variants of Llama-3.2-1B, Llama-3.2-3B~\citep{grattafiori2024llama},
Qwen2.5-3B~\citep{qwen2025qwen25technicalreport}, and
Qwen2.5-Math-7B~\citep{yang2024qwen2},
yielding eight models in total.
We conduct experiments on the mathematical reasoning
benchmarks GSM8K~\citep{cobbe2021training} and
MATH~\citep{hendrycks2021measuring},
with the MATH results provided in Appendix~\ref{app:exp1}.
We use LoRA with rank $r=16$ and scaling parameter $\alpha=32$,
applied to all eligible linear modules in the transformer blocks.
By default, we use Kaiming uniform initialization for
$A_0$, with
$(A_0)_{ij} \overset{\mathrm{i.i.d.}}{\sim}
\mathcal{U}(-1/\sqrt{d_i},1/\sqrt{d_i})$,
and set $B_0=0$.
(Corresponding results with the reverse initialization,
using $A_0=0$ and Kaiming uniform initialization for $B_0$,
are provided in Appendix~\ref{app:exp1}.)
We use Adam~\citep{kingma2014adam} for optimization and consider learning rates
$\eta \in
\{5\times10^{-6},5\times10^{-5},
5\times10^{-4},5\times10^{-3}\}$.
For each model and learning rate, we use the same training
prompt pool, optimizer hyperparameters,
and batch size across the four training settings.
RFT and RFT-B additionally share the same offline
completions.
All diagnostics are computed separately for each
LoRA adapter and then averaged across all adapters.

\subsection{Spectral and Subspace Dynamics}
\label{sec:exp1_results}

\begin{figure*}[t]
    \centering
    \includegraphics[width=0.75\textwidth]{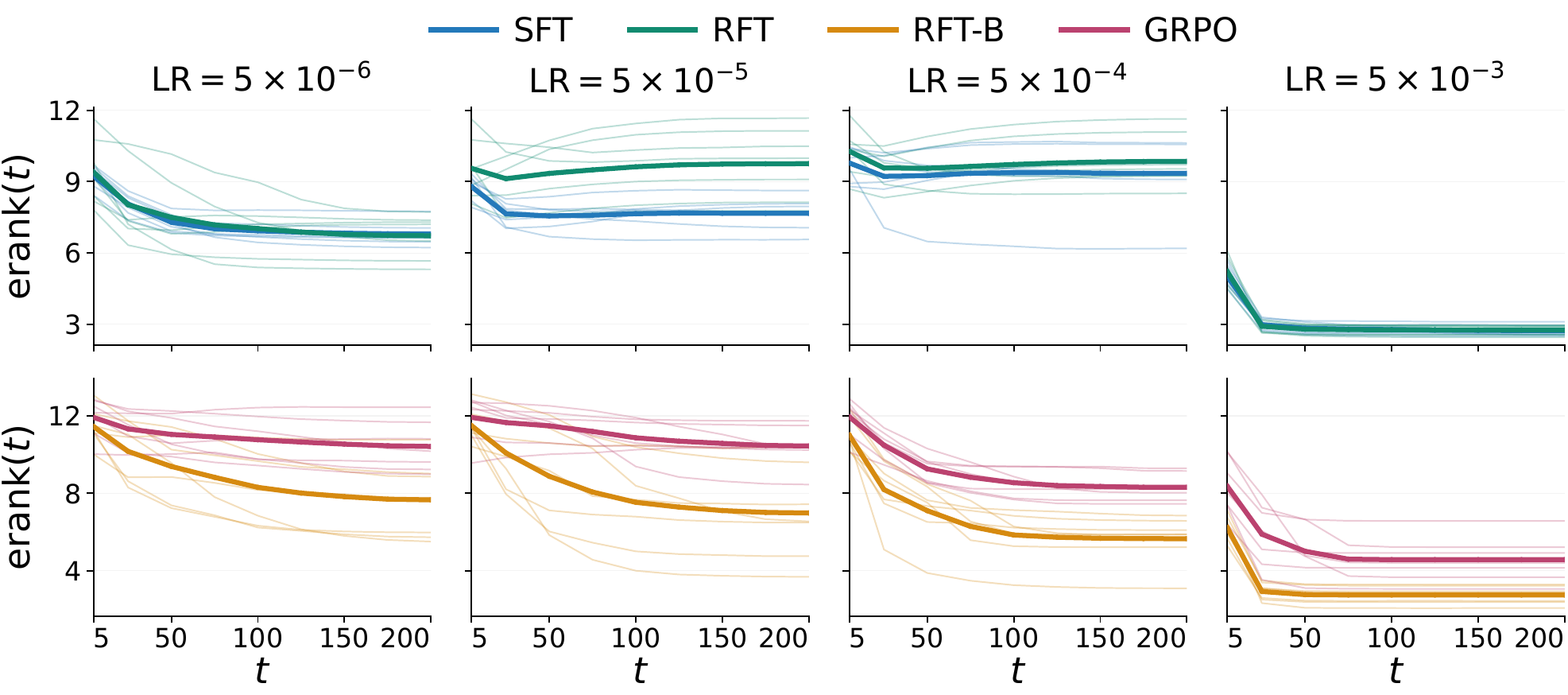}
\caption{\footnotesize
    On GSM8K under $B_0=0$ initialization, GRPO and RFT-B
    exhibit decreasing effective rank across all learning
    rates shown.
    In contrast, SFT and RFT exhibit an increasing trend
    from approximately step 25 at intermediate learning
    rates.
    Columns correspond to different learning rates;
    the top and bottom rows compare SFT/RFT and GRPO/RFT-B,
    respectively.
    Thin curves show individual models, with effective rank
    averaged over LoRA modules; thick curves show the mean
    across all models.
}
    \label{fig:erank-gsm8k-AKB0}
    \vspace{-1em}
\end{figure*}

\paragraph{Effective Rank across Learning Rates.}
In Figure~\ref{fig:erank-gsm8k-AKB0} and
Figures~\ref{fig:erank-math-AKB0},
\ref{fig:erank-gsm8k-BKA0}, and
\ref{fig:erank-math-BKA0} in Appendix~\ref{app:exp1},
GRPO consistently exhibits decreasing effective rank
as training progresses across all learning rates considered.
In contrast, the effective rank dynamics of SFT depend on
the learning rate: effective rank decreases at both small
and large learning rates, while at intermediate learning rates
from $5\times10^{-5}$ to $5\times10^{-4}$, it begins to increase
after approximately 25 training steps.
For models that benefit from SFT\footnote{
Not all models improve under SFT, as observed in our
experiments and in prior studies that fine-tune LLMs
on GSM8K and MATH~\citep{rathore2025much, karan2026finetuning},
consistent with findings that effective SFT depends on
the training data and base-model
capability~\citep{yin-etal-2025-marco,ren2026rethinking}.}, this intermediate regime also yields the best SFT performance
(Table~\ref{tab:optimizer_sft_sweep} in
Appendix~\ref{app:exp2}).
This observation is consistent with prior findings that both
excessively small and large learning rates can impair LoRA
SFT performance~\citep{hayou2024lora+,biderman2024lora},
and further suggests that a less concentrated singular value
distribution accompanies more effective adaptation in these settings.
Moreover, RFT-B mirrors GRPO in exhibiting decreasing
effective rank across all learning rates considered,
whereas RFT shares with SFT an intermediate learning rate
regime in which effective rank increases.
This comparison identifies group-mean baseline subtraction as a
key factor underlying the contrasting spectral dynamics.

\begin{figure*}[t]
    \centering
    \includegraphics[width=0.75\textwidth]{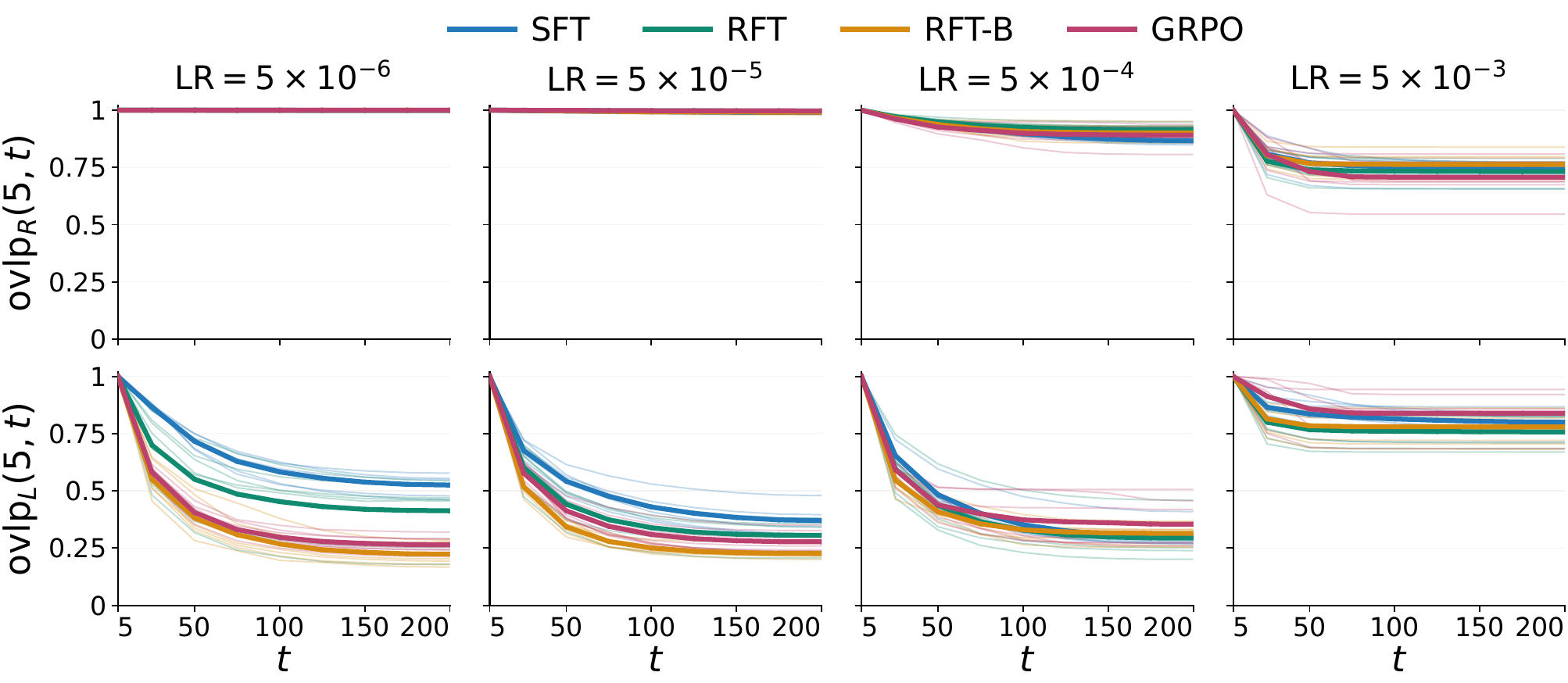}
    \caption{\footnotesize
    On GSM8K under $B_0=0$ initialization, GRPO and RFT-B
    exhibit a faster decrease in left-subspace overlap
    than SFT and RFT at small learning rates, indicating faster
    subspace rotation.
    This difference narrows as the learning rate increases.
The top and bottom rows show the right-subspace overlap
$\mathrm{ovlp}_R(5,t)$ and left-subspace overlap
$\mathrm{ovlp}_L(5,t)$, respectively, between the current
checkpoint at step $t$ and the reference checkpoint at step $5$.
    Columns correspond to different learning rates;
    each panel compares SFT, RFT, RFT-B, and GRPO.
    Thin curves show individual models, with overlap
    averaged over LoRA modules; thick curves show the
    mean across all models.
    }
    \label{fig:ovlp-gsm8k-AKB0}
\end{figure*}

\paragraph{Subspace Rotation across Learning Rates.}
Figure~\ref{fig:ovlp-gsm8k-AKB0} and
Figures~\ref{fig:ovlp-math-AKB0},
\ref{fig:ovlp-gsm8k-BKA0}, and
\ref{fig:ovlp-math-BKA0} in Appendix~\ref{app:exp1}
compare the subspace overlap dynamics across training settings.
First, the dominant rotation depends on the LoRA initialization:
under $B_0=0$, the decrease in overlap is generally more pronounced
for the left singular subspace, whereas under $A_0=0$, it is more
pronounced for the right singular subspace.
Second, at small learning rates, GRPO exhibits a markedly faster
decrease in overlap than SFT (in the left
subspace under $B_0=0$ and in the right subspace under $A_0=0$),
indicating more rapid rotation of the LoRA singular subspaces.
As the learning rate increases, this gap narrows.
Together with the use of smaller learning rates for RL than for
SFT in prior studies~\citep{touvron2023llama, shao2024deepseekmath},
this pattern suggests that promoting subspace
rotation may benefit RL adaptation.

Moreover, as with effective rank, GRPO and RFT-B exhibit similar
rotation dynamics, while SFT and RFT form another qualitatively
similar pair.
These comparisons point to baseline subtraction as the primary
source of the observed spectral and subspace differences in our
setting, rather than differences in completion sources between
SFT and RFT, or online sampling and other implementation details
distinguishing RFT-B from GRPO.
These findings motivate a theoretical understanding of how
baseline subtraction shapes spectral concentration and subspace
evolution, how these effects depend on the learning rate, and how
they relate to prediction risk.
\section{Theoretical Analysis of LoRA Dynamics}
\label{sec:thm}
The comparison between RFT and RFT-B in Section~\ref{sec:exp1} suggests that subtracting a group baseline can change how a LoRA adapter learns. We study this mechanism in a model that connects effective rank, singular-subspace rotation, and the learning rates favorable for prediction.

\paragraph{Model and grouped data.}
We use a quadratic multi-index model, whose prediction depends on several learned projections of the input:
$$
f_W(x)=\frac{\|W^\top x\|_2^2-\|W\|_F^2}{\sqrt2},\qquad W=W_0+BA.
$$
Quadratic models retain learning of both feature strengths and directions, making them useful for studying the two aspects of LoRA geometry considered here \citep{benarous2025learning}. We take unit LoRA scaling, initialize $B_0=0$ and $A_0$ with independent Gaussian entries, and train both factors by SGD. The labels are generated from a teacher weight $F_\delta=U_0\Sigma_0+\delta Z$, where $W_0=U_0\Sigma_0V_0^\top$ is the compact singular value decomposition and $Z$ is a random task perturbation drawn once, with components along pretrained directions and an orthogonal set of newly useful directions. At $\delta=0$, the teacher and pretrained student make identical predictions. $\delta$ therefore controls the size of the post-training change.

To represent the same-prompt groups in the GRPO formulation, we use a pair of correlated, marginally standard Gaussian inputs. This is the smallest group that contains both shared context and within-group variation. Positive correlation $\rho_a$ along pretrained directions models the common prompt and reusable knowledge. We model completion-specific contrasts by negative correlation $\rho_o$ along newly useful directions. The remaining $q$ task-irrelevant coordinates vary independently. The Gaussian approximation makes this correlation structure analytically tractable. Both training rules sample complete groups from the same fixed dataset.

Their difference is whether they remove the group's mean residual. Write $\varepsilon_\pm=f_W(x_\pm)-f_{F_\delta}(x_\pm)$ for the residuals on a pair $(x_+,x_-)$ and $\bar\varepsilon=(\varepsilon_++\varepsilon_-)/2$. Training without a baseline fits the individual residuals. Training with a group baseline first replaces each $\varepsilon_\pm$ by $\varepsilon_\pm-\bar\varepsilon$, changing the squared loss as follows:
$$
\underbrace{\frac14(\varepsilon_+^2+\varepsilon_-^2)}_{\text{without a baseline}}
\quad\longrightarrow\quad
\underbrace{\frac14\{(\varepsilon_+-\bar\varepsilon)^2+(\varepsilon_--\bar\varepsilon)^2\}}_{\text{with a group baseline}}.
$$
This residual operation isolates group baseline subtraction in a regression surrogate for GRPO. We evaluate both rules using the same ordinary population prediction risk.

\paragraph{Post-training regime.}
It is natural to assume that $\delta Z$ is sufficiently small relative to the pretrained representation, capturing post-training tasks that refine already useful features. We also require enough separation in feature strengths for directions to have different learning responses. To obtain a long training window, we let $q\to\infty$ grow polynomially faster than $r^2$, where $r$ is the LoRA rank. The window in the following theorem then grows without bound and its startup occupies a vanishing fraction. 

\begin{theorem}[Informal main theorem]
\label{thm:lora-joint-main}
Under the setting above, there are compatible stable learning-rate intervals such that every low/high pair $\eta_L<\eta_H$ in those intervals has the following properties. After $\widetilde O(r\sqrt q)$ startup updates and through $\Theta(q)$ updates:
\begin{enumerate}
\item \textbf{Effective rank.} Without a baseline, expected effective rank increases at every update at $\eta_H$ and decreases at $\eta_L$. With a group baseline, it decreases at both rates.
\item \textbf{Left singular-subspace rotation.} At either common rate, group baseline subtraction gives smaller expected $\operatorname{ovlp}_L(t,t+1)$ at every adjacent update, for the full rank-$r$ left singular space.
\item \textbf{Excess risk.} At the same update budget, the higher rate gives lower history-averaged risk without a baseline, whereas the lower rate gives lower risk with a group baseline. All four risks decrease. These preferences hold through $\Theta(qr)$ updates.
\end{enumerate}
\end{theorem}

See the formal statements, Theorems~\ref{thm:er}--\ref{thm:risk}, in Appendix~\ref{app:setting-theorems} and the complete proofs in Appendix~\ref{app:proofs}.

\paragraph{Mechanisms.}
The effective-rank trend reflects competition between fitting and redistribution across directions. Coherent fitting favors strongly responding directions, concentrating singular-value energy. Gradient noise, the variation of the stochastic gradient across sampled groups, produces feedback that counteracts this concentration. At leading order in the small task perturbation, the Gram-matrix update has cross terms linear in $\eta$ and gradient-square terms quadratic in $\eta$, so the feedback gains relative influence at larger rates. Normalizing the spectral energy removes uniform growth: effective rank increases only when relative energy becomes more evenly distributed.

Baseline subtraction acts differently on shared responses and within-group contrasts. Products of shared coordinates have correlation $\rho_a^2$ across the pair, whereas products of shared and newly useful coordinates have correlation $\rho_a\rho_o$. Subtracting the pair mean therefore retains fractions $\vartheta_a=(1-\rho_a^2)/2<1/2$ and $\vartheta_o=(1-\rho_a\rho_o)/2>1/2$ of their respective residual energies. Similar responses largely cancel, while contrasting responses survive more strongly and continue to drive fitting. When contrasts carry enough task energy, this dispersing feedback loses more relative strength than concentration after spectral normalization. A large learning rate is then needed for redistribution to dominate, allowing to increase effective rank without a baseline and decrease it with one. At the low rate, fitting dominates in both cases.

Rotation measures how much the span of the adapter's columns changes between checkpoints. Shared responses repeatedly pull updates toward similar directions, helping to stabilize this span across sampled groups. Baseline subtraction suppresses these responses more strongly than within-group contrasts, as reflected by $\vartheta_a<\vartheta_o$. Differences between sampled groups consequently have greater influence on the directions of subsequent updates. Then the span changes more between records, giving smaller left singular-subspace overlap.

For risk, the leading prediction gain is linear in the learning rate, while gradient noise contributes a quadratic cost. In this model, fitting the task component along pretrained directions gives more prediction progress per unit residual energy. Since $\vartheta_a<\vartheta_o$, baseline subtraction suppresses this shared component more strongly, reducing useful progress relative to the noise cost. The favorable learning-rate range therefore shifts downward, yielding the opposite rate preferences in the theorem.

\paragraph{Explaining the empirical observations.}
The theorem predicts effective-rank growth over a suitable range of learning rates, consistent with the intermediate-rate growth for SFT/RFT in Figure~\ref{fig:erank-gsm8k-AKB0} (Section~\ref{sec:exp1_results}). The theoretical high-rate regime has an upper limit, making a different trend at the largest tested rates. The leading-order rotation gap also narrows as the learning rate increases over the proved comparison range. Intuitively, larger rates amplify an additional source of rotation from gradient noise. At fixed adjacent updates, this contribution grows quadratically with the rate and is stronger without a baseline, which retains both the group-average and within-group components. It therefore offsets part of the extra rotation caused by baseline subtraction, consistent with the narrowing empirical separation in Figure~\ref{fig:ovlp-gsm8k-AKB0}.  Appendix~\ref{app:rotation} gives the coefficient comparison.

\paragraph{Implications for optimizer design.}
The risk comparison first supports different learning-rate choices for SFT and group-relative RL. In the model, the favorable higher rate without a baseline accompanies growing effective rank, whereas the favorable lower rate with a baseline accompanies greater left singular-subspace motion than without a baseline at that same rate. The geometry at these respective rates then motivates different update priorities: distributing adaptation across singular directions for SFT and promoting motion outside the current left singular subspace for RL. Section~\ref{sec:exp2_design} develops these ideas through a spectral transformation of the full update for SFT and of its component orthogonal to the current left singular subspace for RL. The specific optimizer combinations are evaluated in Section~\ref{sec:exp2_evaluation}.
\label{main:last}
\section{Dynamics-Informed Configurations for LoRA Optimization}
\label{sec:exp2}

\subsection{Configuration Design}
\label{sec:exp2_design}

Motivated by the empirical findings in
Section~\ref{sec:exp1} and the theoretical analysis in
Section~\ref{sec:thm}, we select dynamics-informed configurations
of standard optimizer components for LoRA in SFT and RL,
with GRPO as the RL algorithm.
Each configuration combines existing ingredients, namely the choice
of SGD or Adam update directions for each factor and a spectral
transformation of the update of $B$; the contribution lies in the
principled choice of combination for each stage.

We write an unmodified LoRA step as
$A_{t+1}=A_t+\eta D_t^A$ and
$B_{t+1}=B_t+\eta D_t^B$,
where $D_t^A$ and $D_t^B$ denote the signed update
directions produced by the base optimizers.
For a nonzero matrix $D$, we further define a
norm-preserving spectral flattening, a variant of the
Muon update~\citep{jordan2024muon}:
\[
\mathcal{M}(D)
:=
\frac{\|D\|_F}{\sqrt{\operatorname{rank}(D)}}
D\left[(D^\top D)^\dagger\right]^{1/2},
\]
where $\dagger$ denotes the Moore--Penrose pseudoinverse.
This transformation flattens the nonzero singular
spectrum by replacing each nonzero singular value
with $\|D\|_F/\sqrt{\operatorname{rank}(D)}$,
while preserving the update magnitude:
$\|\mathcal{M}(D)\|_F=\|D\|_F$.

\paragraph{Dynamics-Informed Configuration for SFT.}
We use an SGD update direction for $B$ with the spectral
transformation $\mathcal{M}$ applied to it, and an Adam
update direction for $A$:
\[
\begin{aligned}
B_{t+1} &= B_t+\eta\mathcal{M}(D_t^B),\\
A_{t+1} &= A_t+\eta D_t^A.
\end{aligned}
\]
Our experiments associate better SFT performance with the growth of effective rank, consistent with the joint geometry and risk trends in Section~\ref{sec:thm}.
The configuration therefore departs from standard Adam in two ways.
First, prior work finds that Muon promotes more uniform
growth across the singular values of the LoRA increment
$\Delta W_t$~\citep{kang2026uniform}.
We thus apply $\mathcal{M}$ to the update of $B$
to encourage spectral flattening.
Second, the mechanism of effective rank in Section~\ref{sec:thm} reflects competition between coherent fitting and feedback from gradient noise. We use SGD for $B$ so that $\mathcal M$ acts directly on the stochastic gradient's singular directions, with the aim of
encouraging the growth of the effective rank. For $A$, we retain the standard Adam update, whose
moment normalization can compensate for small nonzero
gradients and accelerate its early evolution.
Meanwhile, since $B_0=0$, unrolling the updates gives
\[
B_tA_t
=
\eta\Big[\sum_{s=0}^{t-1}\mathcal{M}(D_s^B)\Big]A_0
+\eta^2
\Big[\sum_{s=0}^{t-1}\mathcal{M}(D_s^B)\Big]
\Big[\sum_{s=0}^{t-1}D_s^A\Big].
\]
The updates to $A$ enter directly only
through the second term, leaving the spectral
flattening applied to the $B$ update directions
in the first term unchanged.

\paragraph{Dynamics-Informed Configuration for RL.}
For GRPO, our findings motivate promoting subspace rotation while preserving the within-subspace update. First, we use Adam to compute the base update directions
$D_t^A$ and $D_t^B$ for the LoRA factors $A$ and $B$,
respectively, retaining adaptive scaling for both factors.
Second, while applying $\mathcal{M}$ to the full $B$ update
distributes the update magnitude more evenly across singular
directions and may also promote subspace rotation,
the RL configuration further tailors the spectral
transformation to target subspace rotation more directly.
Define the orthogonal projector onto the current
column space of $B_t$ as
\(
P_t
:=
B_t(B_t^\top B_t)^\dagger B_t^\top.
\)
We apply $\mathcal{M}$ only to the component of
$D_t^B$ orthogonal to this space:
\[
\begin{aligned}
B_{t+1}
&=
B_t+\eta\left[
P_tD_t^B+
\mathcal{M}\!\left((I-P_t)D_t^B\right)
\right],\\
A_{t+1}
&=
A_t+\eta D_t^A.
\end{aligned}
\]
This update preserves the within-subspace component
$P_tD_t^B$ and flattens the spectrum of the orthogonal
component without changing its Frobenius norm.
The transformation thus specifically targets updates
that move $B_t$ beyond its current column space,
providing a direct way to shape subspace rotation.

\begin{table*}[t]
    \centering
        \caption{\footnotesize
    LoRA optimization on GSM8K with Llama-3.2-1B and Llama-3.2-3B.
The dynamics-informed configurations in the last row
    achieve the highest mean accuracies for both SFT (left)
    and RL (right).
    Methods are specified by the update directions for
    the two LoRA factors: Adam and SGD denote the corresponding
    base optimizer directions ($D_t^A$ or $D_t^B$),
    and $\mathcal{M}$ denotes the norm-preserving spectral
    flattening applied to the full direction.
    For RL, $\mathcal{M}_{\perp}(\mathrm{Adam})$ denotes
    $P_tD_t^B+\mathcal{M}((I-P_t)D_t^B)$ with an Adam
    base direction $D_t^B$, which preserves the within-subspace
    component while transforming only the orthogonal component.
   Values show mean test accuracy (\%) across three seeds,
    with standard errors in subscripts.
    For each configuration, results are reported at the
    learning rate with the highest mean accuracy in a completed sweep.
    Bold indicates the highest reported mean in each column.
    }
    \vspace{0.4em}
     {
    \small
    \setlength{\tabcolsep}{3pt}
    \renewcommand{\arraystretch}{1.15}
    \resizebox{\linewidth}{!}{%
\begin{tabular}{@{}llcc@{\hspace{1.5em}}llcc@{}}
        \toprule
        \multicolumn{4}{c}{SFT}
        & \multicolumn{4}{c}{RL} \\
        \cmidrule(lr){1-4}\cmidrule(lr){5-8}
        $A$ direction & $B$ direction
        & Llama-3.2-1B & Llama-3.2-3B
        & $A$ direction & $B$ direction
        & Llama-3.2-1B & Llama-3.2-3B \\
        \midrule

        Adam & Adam
        & $22.24_{\pm 0.33}$
        & $42.99_{\pm 0.38}$
        & Adam & Adam
        & $7.61_{\pm 1.24}$
        & $42.36_{\pm 3.25}$ \\

        Adam & $\mathcal{M}(\mathrm{Adam})$
        & $20.90_{\pm 0.76}$
        & $44.40_{\pm 0.93}$
        & Adam & $\mathcal{M}(\mathrm{Adam})$
        & $9.67_{\pm 1.22}$
        & $46.00_{\pm 0.77}$ \\

        Adam & $\mathcal{M}(\mathrm{SGD})$
        & $\mathbf{22.82}_{\pm 0.12}$
        & $\mathbf{45.72}_{\pm 0.61}$
        & Adam & $\mathcal{M}_{\perp}(\mathrm{Adam})$
        & $\mathbf{10.00}_{\pm 1.36}$
        & $\mathbf{47.66}_{\pm 1.85}$ \\

        \bottomrule
    \end{tabular}%
}
    }
    \label{tab:optimizer_gsm8k}
\end{table*}

\subsection{Empirical Evaluation}
\label{sec:exp2_evaluation}

\paragraph{Experimental Setup.}
We evaluate SFT and GRPO on GSM8K and MATH.
Our model suite includes Llama-3.2-1B, Llama-3.2-1B-Instruct,
Llama-3.2-3B, and Qwen2.5-3B.
Table~\ref{tab:optimizer_gsm8k} presents the GSM8K results
for Llama-3.2-1B and Llama-3.2-3B,
with additional results provided in Appendix~\ref{app:exp2}.
For both SFT and GRPO, we consider two baselines:
standard Adam for both LoRA factors $A$ and $B$;
and Adam base update directions $D_t^A$ and $D_t^B$,
with $\mathcal{M}$ applied to $D_t^B$.
We additionally evaluate the dynamics-informed
configurations for SFT and RL described in
Section~\ref{sec:exp2_design}.
These comparisons examine the effects of spectral
flattening, base optimizer choice, and selective
treatment of the orthogonal component.
Detailed experimental configurations are provided
in Appendix~\ref{app:exp2}.

\paragraph{SFT Results.}
As shown in Tables~\ref{tab:optimizer_gsm8k}
and~\ref{tab:optimizer_math}, while applying $\mathcal{M}$
to the full $B$ update with Adam directions does not
consistently improve over standard Adam,
the dynamics-informed configuration for SFT further replaces
the Adam direction for $B$ with SGD and achieves
the highest reported mean accuracy across all evaluated models and tasks.
These comparisons highlight the benefit of combining
spectral flattening with asymmetric optimization
of the two LoRA factors.

\paragraph{RL Results.}
The comparisons in Tables~\ref{tab:optimizer_gsm8k},
\ref{tab:optimizer_rl_gsm8k}, and~\ref{tab:optimizer_rl_math}
reveal a different preference for GRPO.
With Adam directions for both factors, both the configuration
that applies $\mathcal{M}$ to the full $B$ update and the
dynamics-informed RL configuration improve over standard Adam
in seven of the eight model--dataset settings.
Moreover, the dynamics-informed RL configuration achieves
the highest reported mean accuracy in six of the eight settings.
This supports the benefit of more selectively targeting
the update component responsible for subspace rotation.
Overall, the results favor different optimizer configurations for
SFT and RL, consistent with the distinct dynamics identified
in our analysis.
\section{Conclusion}
\label{sec:conclusion}

In this work, we identify distinct spectral and subspace
dynamics of LoRA under SFT and RL.
Our experiments show that RL effective rank consistently
decreases during training across the learning rates
considered, while for models that benefit from SFT,
the growth of effective rank accompanies better SFT
performance at intermediate learning rates.
In addition to this difference in spectral dynamics,
RL exhibits faster subspace rotation than SFT at small
learning rates, with the gap narrowing as the learning
rate increases.
Controlled comparisons identify group-mean baseline subtraction
as a key source of these differences.
Using a multi-index model, our theoretical framework analyzes how baseline subtraction changes coherent fitting and feedback from gradient noise, jointly shaping effective rank, subspace rotation, and learning-rate preferences in prediction risk. These insights guide the choice of two dynamics-informed
configurations for LoRA optimization, tailored to the distinct
training dynamics of SFT and RL.
Both configurations achieve the highest reported
mean accuracy in most evaluated settings.

\subsection*{AI use statement}

In this work, we used generative AI tools, including OpenAI's GPT-6 Astra, Anthropic's Opus 5.5, and Fable 5.1, to improve manuscript language, identify relevant literature, assist with code creation
and editing, and support the development of the theoretical framework and proofs.
We have not used generative AI tools to design the core framework, identify the empirical findings,
or devise the dynamics-informed configurations.
The authors checked the correctness of all AI-assisted revisions and outputs incorporated
into this work.
We take responsibility for the final content of this work, including text, claims, or artifacts produced with the aid of generative AI.
\subsubsection*{Acknowledgements}
QL acknowledges support of NSF
DMS-2523382.

\bibliographystyle{plainnat}
\bibliography{references}

\clearpage
\appendix

\section{Supplementary Experiments}
\label{app:supp_experiments}

\subsection{Additional Experiments for the Empirical Characterization of LoRA Dynamics}
\label{app:exp1}

\begin{figure}[H]
    \centering
    \includegraphics[width=0.75\textwidth]{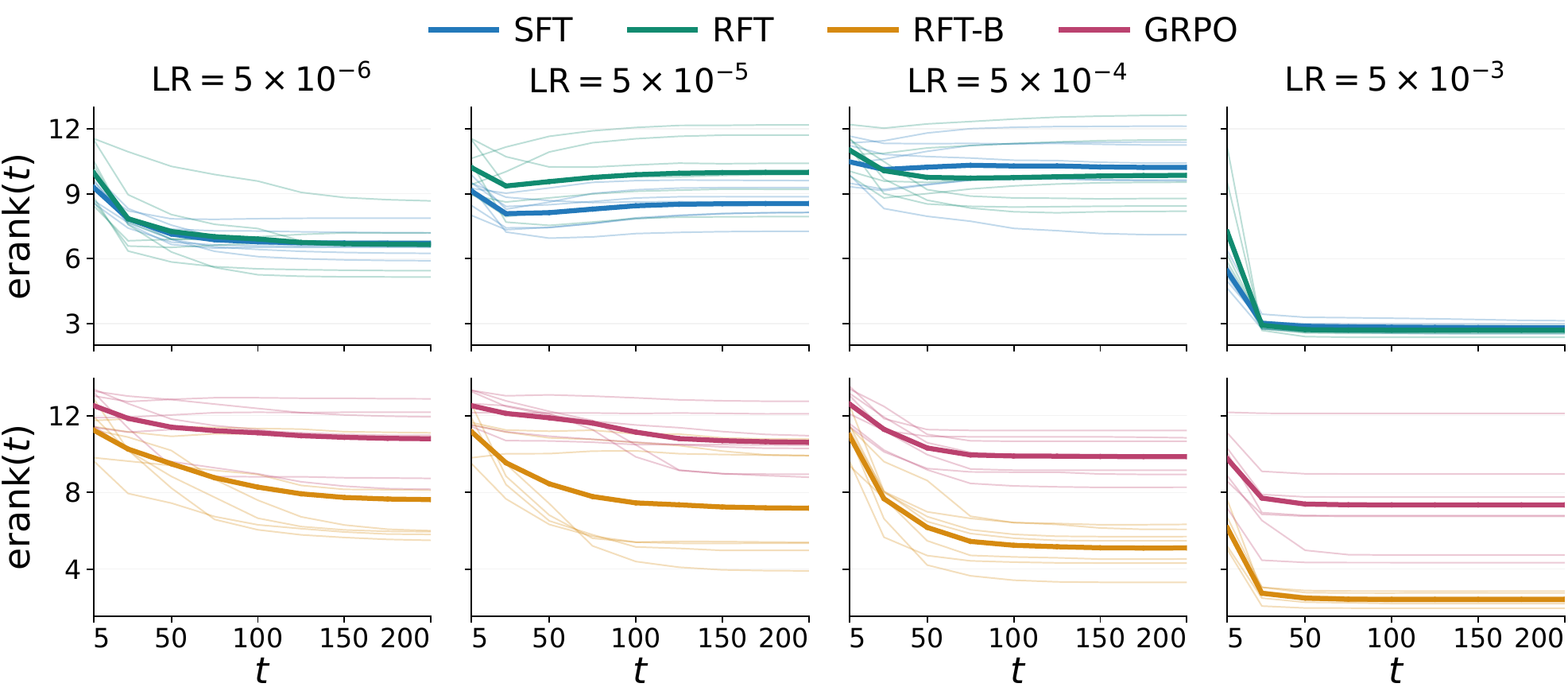}
    \caption{\footnotesize
On MATH under $B_0=0$ initialization, GRPO and RFT-B
    exhibit decreasing effective rank across all learning
    rates shown.
    In contrast, SFT and RFT exhibit an increasing trend
    from approximately step 25 at intermediate learning
    rates.
    Columns correspond to different learning rates;
    the top and bottom rows compare SFT/RFT and GRPO/RFT-B,
    respectively.
    Thin curves show individual models, with effective rank
averaged over LoRA modules; thick curves show the mean
across all models.
    }
    \label{fig:erank-math-AKB0}
\end{figure}

\begin{figure}[H]
    \centering
    \includegraphics[width=0.75\textwidth]{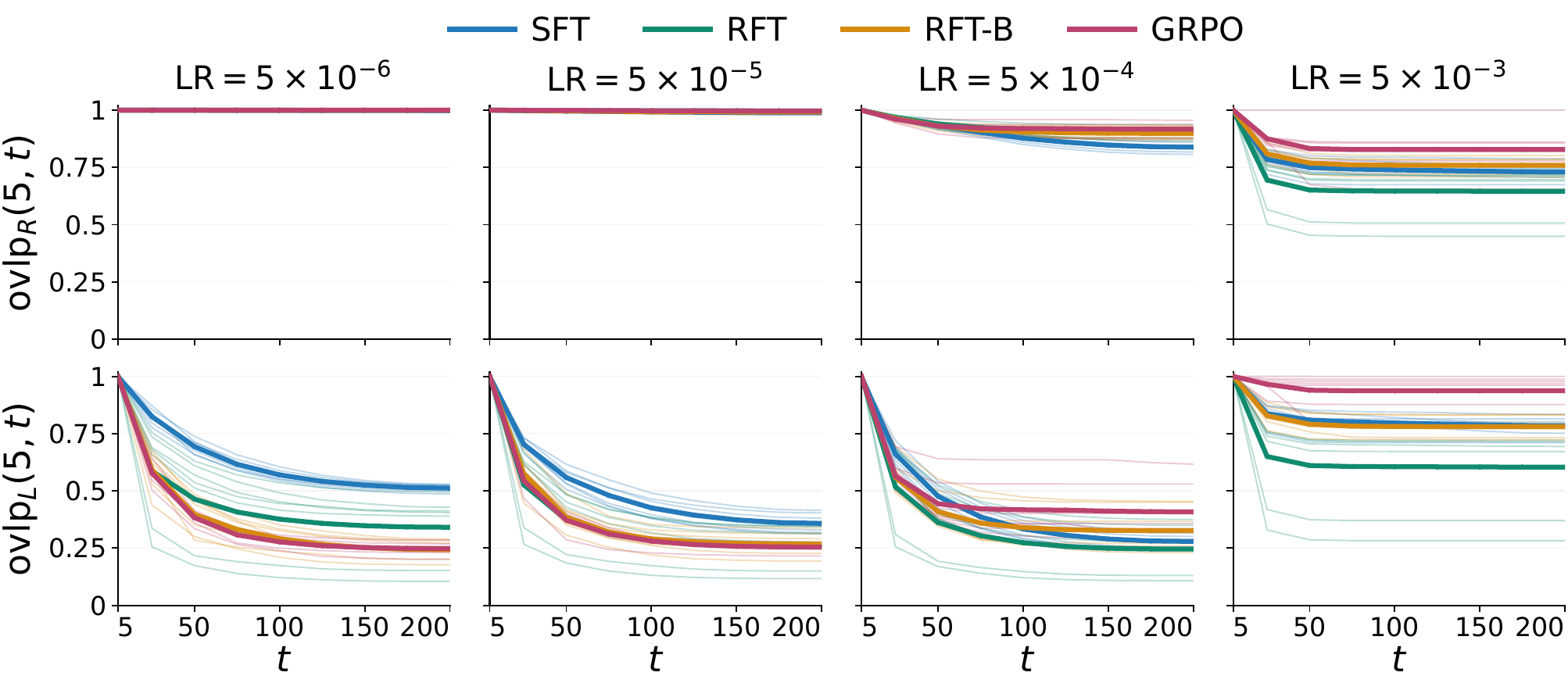}
    \caption{\footnotesize
    On MATH under $B_0=0$ initialization, GRPO and RFT-B
    exhibit a faster decrease in left-subspace overlap
    than SFT and RFT at small learning rates, indicating faster
    subspace rotation.
    This difference narrows as the learning rate increases.
The top and bottom rows show the right-subspace overlap
$\mathrm{ovlp}_R(5,t)$ and left-subspace overlap
$\mathrm{ovlp}_L(5,t)$, respectively, between the current
checkpoint at step $t$ and the reference checkpoint at step $5$.
    Columns correspond to different learning rates;
    each panel compares SFT, RFT, RFT-B, and GRPO.
    Thin curves show individual models, with overlap
    averaged over LoRA modules; thick curves show the
    mean across all models.
    }
    \label{fig:ovlp-math-AKB0}
\end{figure}

\paragraph{Additional Results under $B_0=0$ Initialization.}
Section~\ref{sec:exp1} reports the main spectral and subspace
dynamics on GSM8K under our default initialization, where
$A_0$ is Kaiming uniform initialized and $B_0=0$.
To examine whether these observations extend across datasets,
we additionally report the corresponding results on MATH.
As shown in Figures~\ref{fig:erank-math-AKB0}
and~\ref{fig:ovlp-math-AKB0}, the dynamics of effective rank and
subspace overlap exhibit qualitatively similar
patterns to those observed on GSM8K.

\begin{figure}[H]
    \centering
    \includegraphics[width=0.75\textwidth]{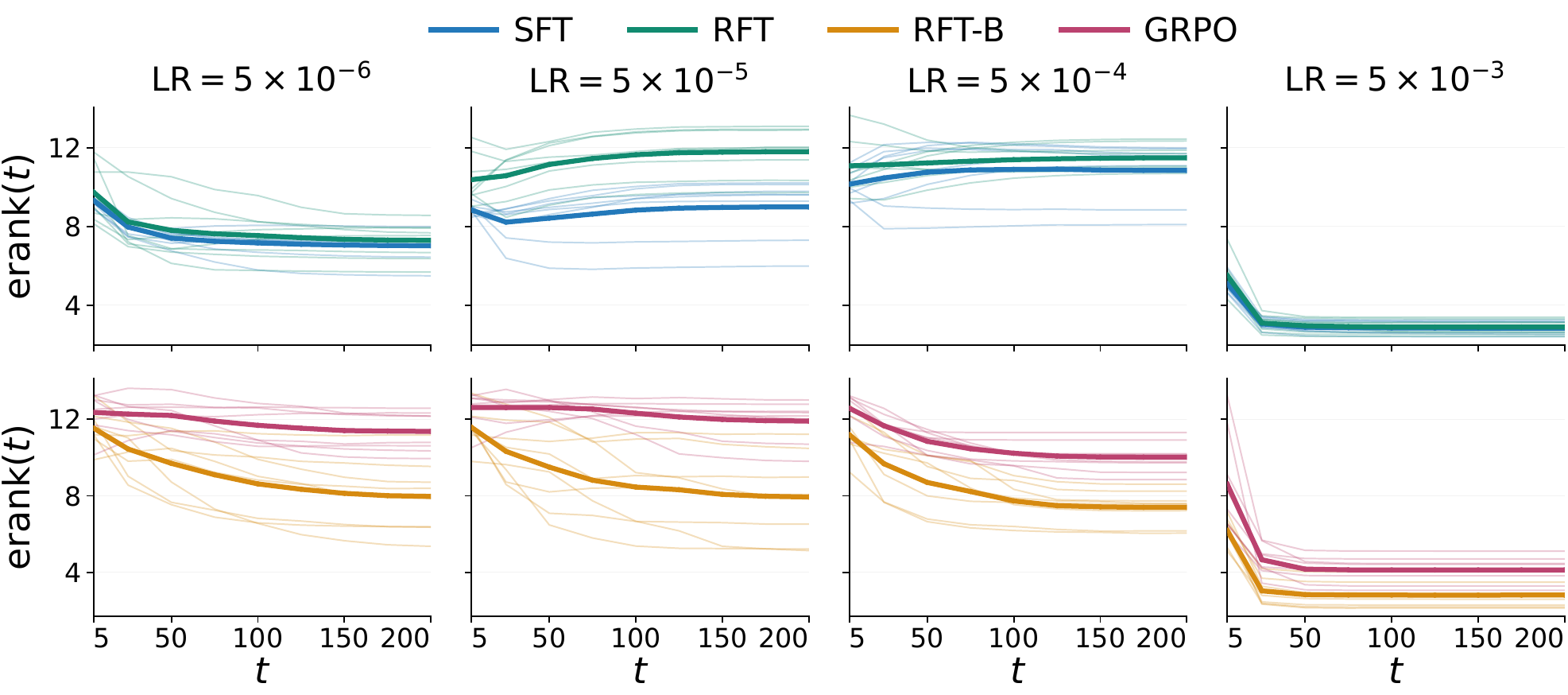}
    \caption{\footnotesize
On GSM8K under $A_0=0$ initialization, GRPO and RFT-B
    exhibit decreasing effective rank across all learning
    rates shown.
    In contrast, SFT and RFT exhibit an increasing trend
    from approximately step 25 at intermediate learning
    rates.
    Columns correspond to different learning rates;
    the top and bottom rows compare SFT/RFT and GRPO/RFT-B,
    respectively.
    Thin curves show individual models, with effective rank
averaged over LoRA modules; thick curves show the mean
across all models.
    }
    \label{fig:erank-gsm8k-BKA0}
\end{figure}

\begin{figure}[H]
    \centering
    \includegraphics[width=0.75\textwidth]{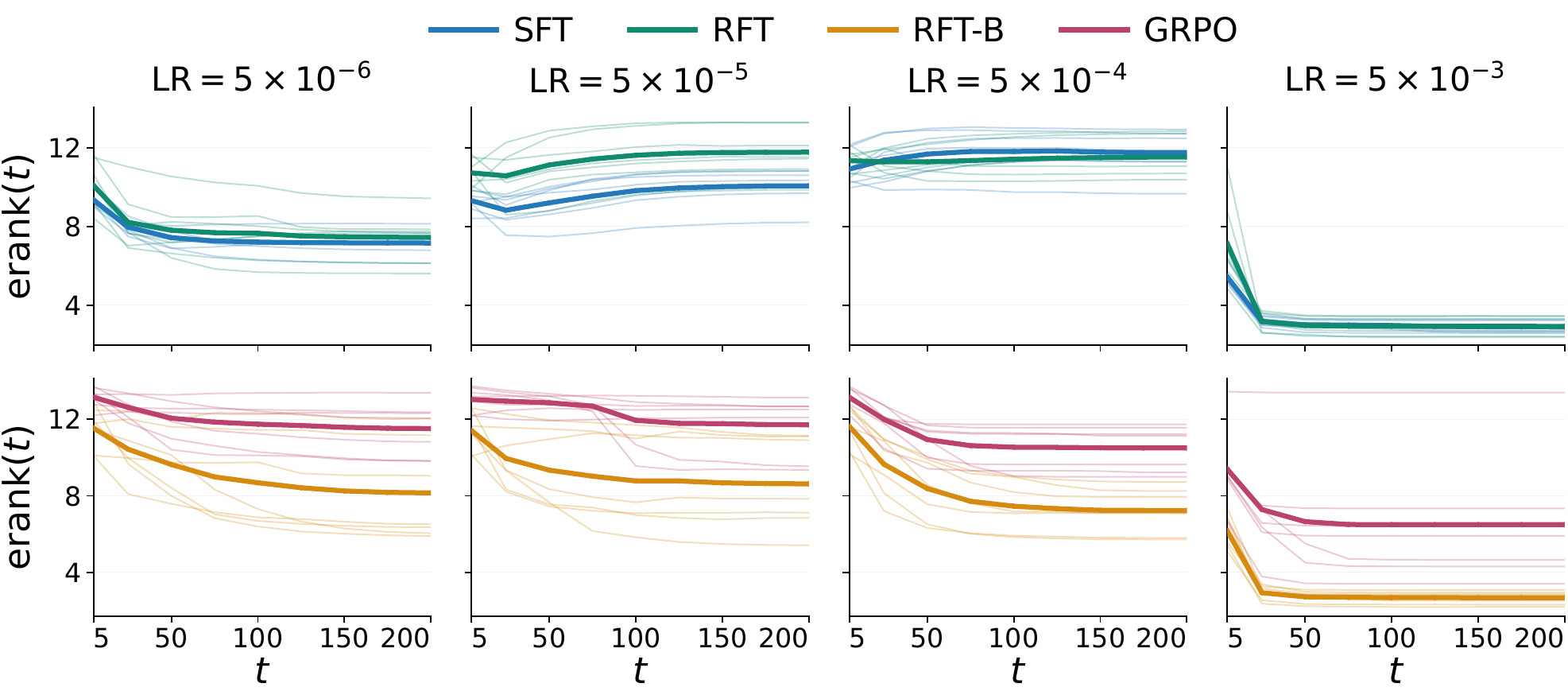}
    \caption{\footnotesize
    On MATH under $A_0=0$ initialization, GRPO and RFT-B
    exhibit decreasing effective rank across all learning
    rates shown.
    In contrast, SFT and RFT exhibit an increasing trend
    from approximately step 25 at intermediate learning
    rates.
    Columns correspond to different learning rates;
    the top and bottom rows compare SFT/RFT and GRPO/RFT-B,
    respectively.
    Thin curves show individual models, with effective rank
averaged over LoRA modules; thick curves show the mean
across all models.
    }
    \label{fig:erank-math-BKA0}
\end{figure}

\begin{figure}[H]
    \centering
    \includegraphics[width=0.75\textwidth]{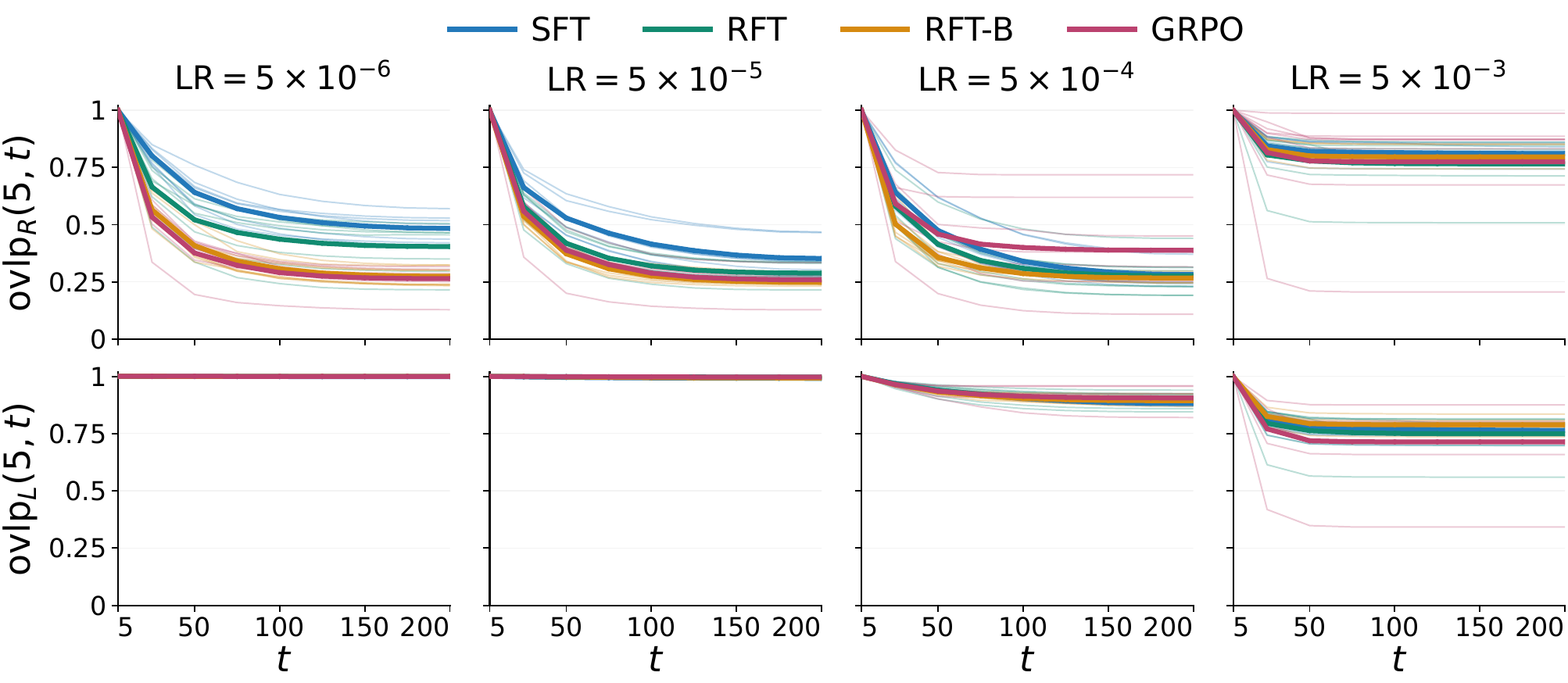}
    \caption{\footnotesize
    On GSM8K under $A_0=0$ initialization, GRPO and RFT-B
    exhibit a faster decrease in right-subspace overlap
    than SFT and RFT at small learning rates, indicating faster
    subspace rotation.
    This difference narrows as the learning rate increases.
The top and bottom rows show the right-subspace overlap
$\mathrm{ovlp}_R(5,t)$ and left-subspace overlap
$\mathrm{ovlp}_L(5,t)$, respectively, between the current
checkpoint at step $t$ and the reference checkpoint at step $5$.
    Columns correspond to different learning rates;
    each panel compares SFT, RFT, RFT-B, and GRPO.
    Thin curves show individual models, with overlap
    averaged over LoRA modules; thick curves show the
    mean across all models.
    }
    \label{fig:ovlp-gsm8k-BKA0}
\end{figure}

\begin{figure}[H]
    \centering
    \includegraphics[width=0.75\textwidth]{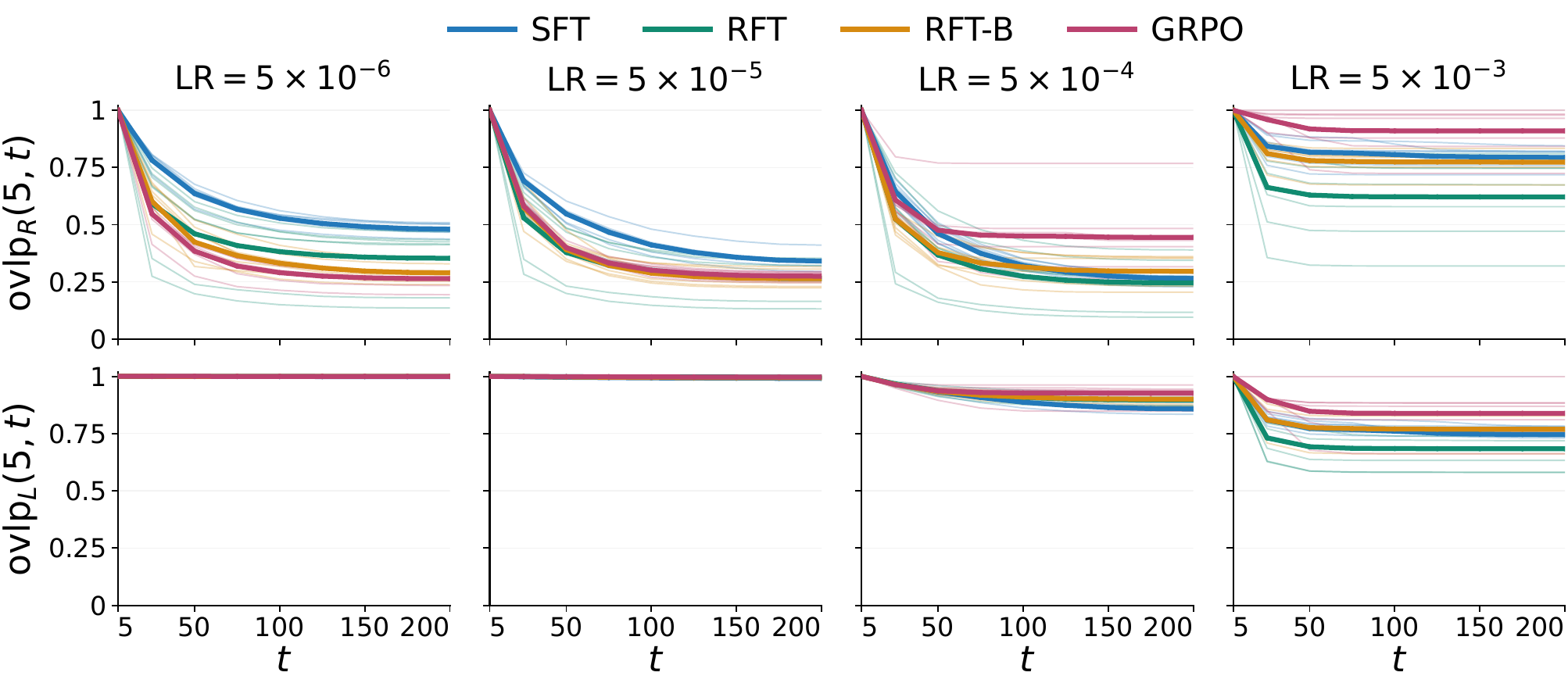}
    \caption{\footnotesize
    On MATH under $A_0=0$ initialization, GRPO and RFT-B
    exhibit a faster decrease in right-subspace overlap
    than SFT and RFT at small learning rates, indicating faster
    subspace rotation.
    This difference narrows as the learning rate increases.
The top and bottom rows show the right-subspace overlap
$\mathrm{ovlp}_R(5,t)$ and left-subspace overlap
$\mathrm{ovlp}_L(5,t)$, respectively, between the current
checkpoint at step $t$ and the reference checkpoint at step $5$.
    Columns correspond to different learning rates;
    each panel compares SFT, RFT, RFT-B, and GRPO.
    Thin curves show individual models, with overlap
    averaged over LoRA modules; thick curves show the
    mean across all models.
    }
    \label{fig:ovlp-math-BKA0}
\end{figure}

\paragraph{Reverse Initialization with $A_0=0$.}
We further examine whether the dynamics identified in
Section~\ref{sec:exp1} are robust to the choice of LoRA
factor initialization by reversing the default initialization:
we set $A_0=0$ and use Kaiming uniform initialization for $B_0$.
Figures~\ref{fig:erank-gsm8k-BKA0}--\ref{fig:ovlp-math-BKA0}
show the corresponding dynamics of effective rank and subspace overlap on GSM8K and MATH.
The main qualitative observations remain consistent with those
under the default initialization.
For effective rank, SFT exhibits a pattern that depends on the learning rate, with effective rank decreasing at small and large
learning rates while an increasing trend emerges at intermediate
learning rates; in contrast, GRPO predominantly exhibits
decreasing effective rank.
For subspace overlap, GRPO exhibits a markedly faster
decrease in overlap than SFT at small learning rates,
while this gap narrows as the learning rate increases.
Across both measures, SFT and RFT remain qualitatively similar,
while RFT-B and GRPO form another similar pair.
These results further support the conclusion from
Section~\ref{sec:exp1} that group-mean baseline subtraction is the primary
factor underlying the observed differences in spectral and
subspace dynamics.

\subsection{Additional Experiments for Dynamics-Informed Configurations}
\label{app:exp2}

\paragraph{Experimental Configurations.}
For the configuration experiments in Section~\ref{sec:exp2_evaluation},
we use fixed data partitions across all optimizer configurations.
For both GSM8K and MATH, we retain the original training and test
splits and reserve 1,000 examples from the training split for
validation.

We use LoRA with rank $r=16$, scaling parameter $\alpha=32$,
and dropout $0.05$. Adapters are applied to all eligible linear modules
in the transformer blocks, including the query, key, value, and output
projections in self-attention and the gate, up, and down
projections in the MLP blocks.
We use the standard initialization with Kaiming-uniform $A_0$
and $B_0=0$.
For configurations using Adam update directions, we sweep the
initial learning rate over
$\{5\times10^{-6},\,5\times10^{-5},\,5\times10^{-4},\,5\times10^{-3}\}$.
For the configuration using an SGD base update direction
for $B$ and an Adam direction for $A$, we use the larger
learning-rate grid
$\{5\times10^{-4},\,5\times10^{-3},\,10^{-2},\,5\times10^{-2}\}$
to accommodate the SGD direction.
The learning rate is linearly decayed during training with no
warmup. Adam uses
$(\beta_1,\beta_2)=(0.9,0.999)$ and $\epsilon=10^{-8}$. SGD uses zero momentum and
zero weight decay.

Each learning rate is evaluated with three random seeds,
$\{7,42,44\}$. Training is run for at most 1,000 optimizer steps.
We save a checkpoint and evaluate validation pass@1 every
50 optimizer steps, and stop training after three consecutive
validation evaluations without a strict improvement.
For each run, the checkpoint attaining the highest mean validation
pass@1 is restored for final test
evaluation. We then average test accuracy across the three seeds;
the results in Section~\ref{sec:exp2_evaluation} report the
learning rate with the highest mean accuracy over the completed
sweep, together with the standard error across seeds.

\paragraph{Additional Results.}
Table~\ref{tab:optimizer_math} reports SFT performance on MATH.
Tables~\ref{tab:optimizer_rl_gsm8k}
and~\ref{tab:optimizer_rl_math} report RL performance
on GSM8K and MATH, respectively.
Table~\ref{tab:optimizer_sft_sweep} provides the detailed
learning-rate sweeps for SFT on both datasets,
while Tables~\ref{tab:optimizer_rl_sweep}
and~\ref{tab:optimizer_rl_math_sweep} provide the corresponding
sweeps for RL on GSM8K and MATH, respectively.

\begin{table}[H]
    \centering
    \caption{\footnotesize
    SFT optimization on MATH with Llama-3.2-1B and Llama-3.2-3B.
    The dynamics-informed configuration in the last row
    achieves the highest mean accuracies.
    Methods are specified by the update directions for
    the two LoRA factors: Adam and SGD denote the corresponding
    base optimizer directions ($D_t^A$ or $D_t^B$),
    and $\mathcal{M}$ denotes the norm-preserving spectral
    flattening applied to the full direction.
    Values show mean test accuracy (\%) across three seeds,
    with standard errors in subscripts.
    For each configuration, results are reported at the
    learning rate with the highest mean accuracy in a completed sweep.
    Bold indicates the highest reported mean in each model column.
    }
    \vspace{1em}
    {
    \small
    \setlength{\tabcolsep}{3pt}
    \renewcommand{\arraystretch}{1.15}
    \begin{tabular*}{0.65\textwidth}
        {@{\extracolsep{\fill}}llcc@{}}
        \toprule
        $A$ direction & $B$ direction
        & Llama-3.2-1B & Llama-3.2-3B \\
        \midrule

        Adam & Adam
        & $5.66_{\pm 0.09}$
        & $10.83_{\pm 0.13}$ \\

        Adam & $\mathcal{M}(\mathrm{Adam})$
        & $5.93_{\pm 0.14}$
        & $10.53_{\pm 0.14}$ \\

        Adam & $\mathcal{M}(\mathrm{SGD})$
        & $\mathbf{6.01}_{\pm 0.25}$
        & $\mathbf{11.32}_{\pm 0.22}$ \\

        \bottomrule
    \end{tabular*}
    }
    \label{tab:optimizer_math}
\end{table}

\begin{table}[H]
    \centering
    \caption{\footnotesize
    RL optimization on GSM8K with Llama-3.2-1B,
    Llama-3.2-1B-Instruct, Llama-3.2-3B, and Qwen2.5-3B.
    The dynamics-informed configuration in the last row
    achieves the highest reported mean accuracy in each model column.
    Methods are specified by the update directions for
    the two LoRA factors: Adam denotes the corresponding
    base optimizer direction ($D_t^A$ or $D_t^B$),
    and $\mathcal{M}$ denotes the norm-preserving spectral
    flattening applied to the full direction.
    $\mathcal{M}_{\perp}(\mathrm{Adam})$ denotes
    $P_tD_t^B+\mathcal{M}((I-P_t)D_t^B)$ with an Adam
    base direction $D_t^B$, which preserves the within-subspace
    component while transforming only the orthogonal component.
    Values show mean test accuracy (\%) across three seeds,
    with standard errors in subscripts.
    For each configuration, results are reported at the
    learning rate with the highest mean accuracy in a completed sweep.
    Bold indicates the highest reported mean in each model column.
    }
    \vspace{1em}
    {
    \small
    \setlength{\tabcolsep}{3pt}
    \renewcommand{\arraystretch}{1.15}
    \begin{tabular*}{\textwidth}
        {@{\extracolsep{\fill}}llcccc@{}}
        \toprule
        $A$ direction & $B$ direction
        & Llama-3.2-1B & Llama-3.2-1B-Instruct
        & Llama-3.2-3B & Qwen2.5-3B \\
        \midrule

        Adam & Adam
        & $7.61_{\pm 1.24}$
        & $57.85_{\pm 0.39}$
        & $42.36_{\pm 3.25}$
        & $83.93_{\pm 0.34}$ \\

        Adam & $\mathcal{M}(\mathrm{Adam})$
        & $9.67_{\pm 1.22}$
        & $57.97_{\pm 0.18}$
        & $46.00_{\pm 0.77}$
        & $84.46_{\pm 0.38}$ \\

        Adam & $\mathcal{M}_{\perp}(\mathrm{Adam})$
        & $\mathbf{10.00}_{\pm 1.36}$
        & $\mathbf{58.23}_{\pm 0.72}$
        & $\mathbf{47.66}_{\pm 1.85}$
        & $\mathbf{84.51}_{\pm 0.62}$ \\

        \bottomrule
    \end{tabular*}
    }
    \label{tab:optimizer_rl_gsm8k}
\end{table}

\begin{table}[H]
    \centering
    \caption{\footnotesize
    RL optimization on MATH with Llama-3.2-1B,
    Llama-3.2-1B-Instruct, Llama-3.2-3B, and Qwen2.5-3B.
    The dynamics-informed configuration in the last row
    achieves the highest reported mean accuracy on two of the four models.
    Methods are specified by the update directions for
    the two LoRA factors: Adam denotes the corresponding
    base optimizer direction ($D_t^A$ or $D_t^B$),
    and $\mathcal{M}$ denotes the norm-preserving spectral
    flattening applied to the full direction.
    $\mathcal{M}_{\perp}(\mathrm{Adam})$ denotes
    $P_tD_t^B+\mathcal{M}((I-P_t)D_t^B)$ with an Adam
    base direction $D_t^B$, which preserves the within-subspace
    component while transforming only the orthogonal component.
    Values show mean test accuracy (\%) across three seeds,
    with standard errors in subscripts.
    For each configuration, results are reported at the
    learning rate with the highest mean accuracy in a completed sweep.
    Bold indicates the highest reported mean in each model column.
    }
    \vspace{1em}
    {
    \small
    \setlength{\tabcolsep}{3pt}
    \renewcommand{\arraystretch}{1.15}
    \begin{tabular*}{\textwidth}
        {@{\extracolsep{\fill}}llcccc@{}}
        \toprule
        $A$ direction & $B$ direction
        & Llama-3.2-1B & Llama-3.2-1B-Instruct
        & Llama-3.2-3B & Qwen2.5-3B \\
        \midrule

        Adam & Adam
        & $5.04_{\pm 0.28}$
        & $33.44_{\pm 0.52}$
        & $\mathbf{9.41}_{\pm 0.87}$
        & $63.03_{\pm 0.23}$ \\

        Adam & $\mathcal{M}(\mathrm{Adam})$
        & $5.43_{\pm 0.28}$
        & $\mathbf{34.16}_{\pm 0.61}$
        & $9.21_{\pm 0.44}$
        & $63.37_{\pm 0.21}$ \\

        Adam & $\mathcal{M}_{\perp}(\mathrm{Adam})$
        & $\mathbf{5.65}_{\pm 0.27}$
        & $33.71_{\pm 0.17}$
        & $8.21_{\pm 0.42}$
        & $\mathbf{63.54}_{\pm 0.21}$ \\

        \bottomrule
    \end{tabular*}
    }
    \label{tab:optimizer_rl_math}
\end{table}

\begin{table}[H]
    \centering
    \caption{\footnotesize
    Detailed learning-rate sweeps for SFT on GSM8K and MATH.
    Methods are specified by the update directions for
    the two LoRA factors: Adam and SGD denote the corresponding
    base optimizer directions ($D_t^A$ or $D_t^B$),
    and $\mathcal{M}$ denotes the norm-preserving spectral
    flattening applied to the full direction.
    Values show mean test accuracy (\%) across three seeds,
    with standard errors in subscripts.
    Bold indicates the highest mean accuracy across learning
    rates for each update configuration and model--dataset pair.
    }
    \vspace{1em}
    {
    \small
    \setlength{\tabcolsep}{4pt}
    \renewcommand{\arraystretch}{1.12}
    \begin{tabular*}{0.95\textwidth}
        {@{\extracolsep{\fill}}lllcccc@{}}
        \toprule
        & & & \multicolumn{2}{c}{GSM8K}
        & \multicolumn{2}{c}{MATH} \\
        \cmidrule(lr){4-5}\cmidrule(lr){6-7}
        $A$ direction & $B$ direction & LR
        & Llama-3.2-1B & Llama-3.2-3B
        & Llama-3.2-1B & Llama-3.2-3B \\
        \midrule

        Adam & Adam
        & $5\times10^{-6}$
        & $10.87_{\pm 1.20}$ & $32.35_{\pm 0.37}$
        & $5.06_{\pm 0.02}$ & $10.19_{\pm 0.21}$ \\

        & & $5\times10^{-5}$
        & $20.60_{\pm 0.61}$ & $\mathbf{42.99}_{\pm 0.38}$
        & $5.46_{\pm 0.12}$ & $10.57_{\pm 0.21}$ \\

        & & $5\times10^{-4}$
        & $\mathbf{22.24}_{\pm 0.33}$ & $42.68_{\pm 0.92}$
        & $\mathbf{5.66}_{\pm 0.09}$
        & $\mathbf{10.83}_{\pm 0.13}$ \\

        & & $5\times10^{-3}$
        & $2.65_{\pm 0.08}$ & $0.58_{\pm 0.58}$
        & $3.43_{\pm 0.52}$ & $2.83_{\pm 1.41}$ \\

        \addlinespace[2pt]

        Adam & $\mathcal{M}(\mathrm{Adam})$
        & $5\times10^{-6}$
        & $11.22_{\pm 0.20}$ & $30.65_{\pm 0.62}$
        & $4.96_{\pm 0.14}$ & $10.44_{\pm 0.16}$ \\

        & & $5\times10^{-5}$
        & $\mathbf{20.90}_{\pm 0.76}$
        & $\mathbf{44.40}_{\pm 0.93}$
        & $5.45_{\pm 0.17}$ & $10.39_{\pm 0.36}$ \\

        & & $5\times10^{-4}$
        & $20.67_{\pm 0.32}$ & $43.47_{\pm 0.54}$
        & $\mathbf{5.93}_{\pm 0.14}$
        & $\mathbf{10.53}_{\pm 0.14}$ \\

        & & $5\times10^{-3}$
        & $2.43_{\pm 0.27}$ & $2.45_{\pm 0.24}$
        & $4.15_{\pm 0.15}$ & $3.83_{\pm 0.38}$ \\

        \addlinespace[2pt]

        Adam & $\mathcal{M}(\mathrm{SGD})$
        & $5\times10^{-4}$
        & $13.01_{\pm 0.17}$ & $32.12_{\pm 0.42}$
        & $5.19_{\pm 0.21}$ & $10.40_{\pm 0.06}$ \\

        & & $5\times10^{-3}$
        & $22.52_{\pm 0.99}$ & $\mathbf{45.72}_{\pm 0.61}$
        & $5.85_{\pm 0.30}$ & $10.67_{\pm 0.25}$ \\

        & & $1\times10^{-2}$
        & $\mathbf{22.82}_{\pm 0.12}$ & $44.73_{\pm 0.74}$
        & $\mathbf{6.01}_{\pm 0.25}$
        & $\mathbf{11.32}_{\pm 0.22}$ \\

        & & $5\times10^{-2}$
        & $20.82_{\pm 1.03}$ & $40.97_{\pm 1.61}$
        & $5.82_{\pm 0.57}$ & $9.88_{\pm 0.18}$ \\

        \bottomrule
    \end{tabular*}
    }
    \label{tab:optimizer_sft_sweep}
\end{table}

\begin{table}[H]
    \centering
    \caption{\footnotesize
    Detailed learning-rate sweeps for RL on GSM8K.
    Methods are specified by the update directions for
    the two LoRA factors: Adam and SGD denote the corresponding
    base optimizer directions ($D_t^A$ or $D_t^B$),
    and $\mathcal{M}$ denotes the norm-preserving spectral
    flattening applied to the full direction.
    For RL, $\mathcal{M}_{\perp}(\mathrm{Adam})$ denotes
    $P_tD_t^B+\mathcal{M}((I-P_t)D_t^B)$ with an Adam
    base direction $D_t^B$, which preserves the within-subspace
    component while transforming only the orthogonal component.
    Values show mean test accuracy (\%) across three seeds,
    with standard errors in subscripts.
    Bold indicates the highest mean accuracy across learning
    rates for each update configuration and model.
    }
    \vspace{0.6em}
    {
    \small
    \setlength{\tabcolsep}{3pt}
    \renewcommand{\arraystretch}{1.12}
    \begin{tabular*}{\textwidth}
        {@{\extracolsep{\fill}}lllcccc@{}}
        \toprule
        $A$ direction & $B$ direction & LR
        & Llama-3.2-1B & Llama-3.2-1B-Instruct
        & Llama-3.2-3B & Qwen2.5-3B \\
        \midrule

        Adam & Adam
        & $5\times10^{-6}$
        & $2.65_{\pm 0.15}$
        & $\mathbf{57.85}_{\pm 0.39}$
        & $27.02_{\pm 3.31}$
        & $82.82_{\pm 1.11}$ \\

        & & $5\times10^{-5}$
        & $\mathbf{7.61}_{\pm 1.24}$
        & $57.24_{\pm 0.29}$
        & $\mathbf{42.36}_{\pm 3.25}$
        & $\mathbf{83.93}_{\pm 0.34}$ \\

        & & $5\times10^{-4}$
        & $1.34_{\pm 0.77}$
        & $2.75_{\pm 2.21}$
        & $0.15_{\pm 0.15}$
        & $0.96_{\pm 0.68}$ \\

        & & $5\times10^{-3}$
        & $0.00_{\pm 0.00}$
        & $0.00_{\pm 0.00}$
        & $0.00_{\pm 0.00}$
        & $0.00_{\pm 0.00}$ \\

        \addlinespace[2pt]

        Adam & $\mathcal{M}(\mathrm{Adam})$
        & $5\times10^{-6}$
        & $2.91_{\pm 0.21}$
        & $\mathbf{57.97}_{\pm 0.18}$
        & $14.30_{\pm 1.12}$
        & $82.61_{\pm 0.48}$ \\

        & & $5\times10^{-5}$
        & $\mathbf{9.67}_{\pm 1.22}$
        & $57.24_{\pm 0.59}$
        & $\mathbf{46.00}_{\pm 0.77}$
        & $\mathbf{84.46}_{\pm 0.38}$ \\

        & & $5\times10^{-4}$
        & $0.35_{\pm 0.32}$
        & $3.87_{\pm 1.52}$
        & $0.00_{\pm 0.00}$
        & $39.60_{\pm 19.67}$ \\

        & & $5\times10^{-3}$
        & $0.00_{\pm 0.00}$
        & $0.00_{\pm 0.00}$
        & $0.00_{\pm 0.00}$
        & $0.00_{\pm 0.00}$ \\

        \addlinespace[2pt]

        Adam & $\mathcal{M}_{\perp}(\mathrm{Adam})$
        & $5\times10^{-6}$
        & $2.88_{\pm 0.19}$
        & $56.86_{\pm 0.23}$
        & $19.69_{\pm 1.40}$
        & $83.50_{\pm 0.24}$ \\

        & & $5\times10^{-5}$
        & $\mathbf{10.00}_{\pm 1.36}$
        & $\mathbf{58.23}_{\pm 0.72}$
        & $\mathbf{47.66}_{\pm 1.85}$
        & $\mathbf{84.51}_{\pm 0.62}$ \\

        & & $5\times10^{-4}$
        & $1.01_{\pm 0.51}$
        & $0.48_{\pm 0.40}$
        & $1.42_{\pm 0.71}$
        & $51.76_{\pm 19.42}$ \\

        & & $5\times10^{-3}$
        & $0.00_{\pm 0.00}$
        & $0.00_{\pm 0.00}$
        & $0.00_{\pm 0.00}$
        & $0.00_{\pm 0.00}$ \\

        \bottomrule
    \end{tabular*}
    }
    \label{tab:optimizer_rl_sweep}
\end{table}

\begin{table}[H]
    \centering
    \caption{\footnotesize
    Detailed learning-rate sweeps for RL on MATH.
    Methods are specified by the update directions for
    the two LoRA factors: Adam and SGD denote the corresponding
    base optimizer directions ($D_t^A$ or $D_t^B$),
    and $\mathcal{M}$ denotes the norm-preserving spectral
    flattening applied to the full direction.
    For RL, $\mathcal{M}_{\perp}(\mathrm{Adam})$ denotes
    $P_tD_t^B+\mathcal{M}((I-P_t)D_t^B)$ with an Adam
    base direction $D_t^B$, which preserves the within-subspace
    component while transforming only the orthogonal component.
    Values show mean test accuracy (\%) across three seeds,
    with standard errors in subscripts.
    Bold indicates the highest mean accuracy across learning
    rates for each update configuration and model.
    }
    \vspace{0.6em}
    {
    \small
    \setlength{\tabcolsep}{3pt}
    \renewcommand{\arraystretch}{1.12}
    \begin{tabular*}{\textwidth}
        {@{\extracolsep{\fill}}lllcccc@{}}
        \toprule
        $A$ direction & $B$ direction & LR
        & Llama-3.2-1B & Llama-3.2-1B-Instruct
        & Llama-3.2-3B & Qwen2.5-3B \\
        \midrule

        Adam & Adam
        & $5\times10^{-6}$
        & $4.79_{\pm 0.02}$
        & $\mathbf{33.44}_{\pm 0.52}$
        & $7.28_{\pm 0.19}$
        & $62.78_{\pm 0.40}$ \\

        & & $5\times10^{-5}$
        & $\mathbf{5.04}_{\pm 0.28}$
        & $32.55_{\pm 0.27}$
        & $\mathbf{9.41}_{\pm 0.87}$
        & $\mathbf{63.03}_{\pm 0.23}$ \\

        & & $5\times10^{-4}$
        & $0.00_{\pm 0.00}$
        & $1.17_{\pm 1.17}$
        & $0.01_{\pm 0.01}$
        & $0.59_{\pm 0.59}$ \\

        & & $5\times10^{-3}$
        & $0.00_{\pm 0.00}$
        & $0.00_{\pm 0.00}$
        & $0.00_{\pm 0.00}$
        & $0.00_{\pm 0.00}$ \\

        \addlinespace[2pt]

        Adam & $\mathcal{M}(\mathrm{Adam})$
        & $5\times10^{-6}$
        & $4.89_{\pm 0.01}$
        & $32.91_{\pm 0.21}$
        & $7.01_{\pm 0.07}$
        & $61.80_{\pm 0.64}$ \\

        & & $5\times10^{-5}$
        & $\mathbf{5.43}_{\pm 0.28}$
        & $\mathbf{34.16}_{\pm 0.61}$
        & $\mathbf{9.21}_{\pm 0.44}$
        & $\mathbf{63.37}_{\pm 0.21}$ \\

        & & $5\times10^{-4}$
        & $4.05_{\pm 0.37}$
        & $5.94_{\pm 4.28}$
        & $2.18_{\pm 1.17}$
        & $29.09_{\pm 11.67}$ \\

        & & $5\times10^{-3}$
        & $0.00_{\pm 0.00}$
        & $0.00_{\pm 0.00}$
        & $0.00_{\pm 0.00}$
        & $0.00_{\pm 0.00}$ \\

        \addlinespace[2pt]

        Adam & $\mathcal{M}_{\perp}(\mathrm{Adam})$
        & $5\times10^{-6}$
        & $4.71_{\pm 0.06}$
        & $\mathbf{33.71}_{\pm 0.17}$
        & $7.30_{\pm 0.18}$
        & $62.28_{\pm 0.14}$ \\

        & & $5\times10^{-5}$
        & $\mathbf{5.65}_{\pm 0.27}$
        & $32.72_{\pm 0.29}$
        & $\mathbf{8.21}_{\pm 0.42}$
        & $\mathbf{63.54}_{\pm 0.21}$ \\

        & & $5\times10^{-4}$
        & $3.06_{\pm 0.41}$
        & $2.50_{\pm 1.29}$
        & $1.43_{\pm 1.43}$
        & $19.72_{\pm 10.13}$ \\

        & & $5\times10^{-3}$
        & $0.00_{\pm 0.00}$
        & $0.00_{\pm 0.00}$
        & $0.00_{\pm 0.00}$
        & $0.00_{\pm 0.00}$ \\

        \bottomrule
    \end{tabular*}
    }
    \label{tab:optimizer_rl_math_sweep}
\end{table}

\subsection{Dataset and Model Licenses}
\label{app:licenses}

GSM8K and MATH are distributed under the MIT License.
The base and instruction-tuned variants of
Llama-3.2-1B and Llama-3.2-3B are released under
the Llama 3.2 Community License Agreement.
The corresponding variants of Qwen2.5-3B and
Qwen2.5-Math-7B are released under the Qwen Research
License Agreement and the Apache License 2.0, respectively.

\subsection{Computing Infrastructure}
\label{app:compute}

Experiments were conducted on an institutional
high-performance computing cluster using NVIDIA H200 GPUs.
\section{Complete Theoretical Results}
\label{app:setting-theorems}

\subsection{Setting}
\label{main:setting}
\paragraph{Model and task.}
The student weight has dimensions $W_t\in\R^{d_o\times d_i}$, with $d_o=k+p+q$.
The quadratic surrogate takes $x\in\R^{d_o}$ and has $d_i$ feature columns.
Its rank-$r$ adapter has unit gain $\alpha/r=1$.
The LoRA gain is therefore fixed, and $\alpha,\beta\in\{+,-\}$ subsequently index positions within a pair:

\begin{equation*}
W_t:=W_0+B_tA_t,\qquad f_W(x):=\frac{\langle WW^\top,xx^\top-I\rangle_F}{\sqrt2},\qquad B_t\in\R^{d_o\times r},\quad A_t\in\R^{r\times d_i}.
\end{equation*}
Write $W_0=U_0\Sigma_0V_0^\top$, where $U_0,V_0$ have $k$ orthonormal columns and $\Sigma_0$ is positive diagonal.
The $p$-column isometry $U_{\perp}$ has range orthogonal to $U_0$.
The teacher has $k$ columns and is
\begin{equation*}
F_\delta:=U_0\Sigma_0+\delta(U_0Z_a+U_{\perp}Z_o),\qquad Z_a:=\frac{\sigma_a}{\sqrt k}\Xi_a,\quad Z_o:=\frac{\sigma_o}{\sqrt p}\Xi_o,\qquad \sigma_a,\sigma_o>0.
\end{equation*}
The entries of $\Xi_a\in\R^{k\times k}$ and $\Xi_o\in\R^{p\times k}$ are independent, centered, variance-one, and uniformly sub-Gaussian.
They are drawn once, independently of initialization and data.
Labels equal $f_{F_\delta}(x)$.
Initialize $B_0=0$ and independent $(A_0)_{ij}\sim \mathcal{N}(0,1/d_i)$.

\paragraph{Fixed paired data.}
Generate $n$ independent Gaussian pairs once.
Both positions have marginal $\mathcal{N}(0,I)$, with
\begin{equation*}
\E x_+x_-^\top=\rho_aU_0U_0^\top+\rho_oU_{\perp}U_{\perp}^\top,\qquad 0<\rho_a<1,\quad -1<\rho_o<0.
\end{equation*}
Thus the remaining $q$ nuisance coordinates are independent across positions.
Each update samples one complete stored pair uniformly with replacement.

\paragraph{The two training rules.}
We call training without a baseline the \emph{raw} rule ($m=R$) and training with a group baseline the \emph{centered} rule ($m=C$).
For $\varepsilon_\pm:=f_W(x_\pm)-f_{F_\delta}(x_\pm)$, use either raw or residual-centered loss and simultaneous factor SGD:
\begin{equation}
\begin{gathered}\ell_R:=\frac{\varepsilon_+^2+\varepsilon_-^2}{4},\qquad \ell_C:=\frac{(\varepsilon_+-\varepsilon_-)^2}{8},\\B_{t+1}=B_t-\eta\nabla_B\ell,\qquad
A_{t+1}=A_t-\eta\nabla_A\ell.\end{gathered}
\label{eq:losses}
\end{equation}
Centering subtracts the mean residual, without additional gradient rescaling.
All runs share the realized teacher, initialization $A_0$, fixed training set, and low/high rate pair.

\paragraph{Observables.}
For $\Delta W_t=B_tA_t\ne0$, define the effective rank used in the paper:
\begin{equation*}
\ER(\Delta W_t):=\exp\left(-\sum_i p_i\log p_i\right),\qquad p_i:=\frac{\sigma_i(\Delta W_t)^2}{\sum_j\sigma_j(\Delta W_t)^2}.
\end{equation*}
Specifically, zero weights contribute zero entropy and set $\ER(0)=1$ on exceptional histories.
For $t\ge1$, write $\Delta W_t=U_t\Sigma_tV_t^\top$ for its rank-$r$ thin singular value decomposition, with any fixed measurable orthonormal rank-$r$ completion on deficient histories.
Define
\begin{equation*}
\operatorname{ovlp}_L(t,t'):=\frac1r\|U_t^\top U_{t'}\|_F^2,\qquad \Rot(t,t'):=1-\operatorname{ovlp}_L(t,t').
\end{equation*}
The ordinary population prediction risk is
\begin{equation*}
\mathcal R_\delta(W):=\frac12\E_x\{f_W(x)-f_{F_\delta}(x)\}^2=\frac12\|WW^\top-F_\delta F_\delta^\top\|_F^2.
\end{equation*}

The test input is an independent standard Gaussian.
Fact~\ref{fact:gaussian-isometry} proves the risk identity.
For any run-dependent object $X$, the subscript $X_{m,j,t}$ records the rule $m\in\{R,C\}$, rate $j\in\{L,H\}$, and update $t$, in that order.
The expectation $\E_{\mathrm{hist}}$ averages the sampled pair indices conditional on the realized initialization $A_0$, task, and training set $\mathcal D_{\mathrm{train}}$.
Unqualified expectations in the theorems average all four sources of randomness.

\subsection{Assumptions}
All population parameters are fixed before drawing the task, initialization, or training set.
The following statements specify the primitive assumptions.

\begin{assumption}[Dimensions and training-set size]
\label{ass:dimensions}
Along the rank limit,
\begin{equation*}
k/r\to \gamma_k>1,\qquad d_i/r\to \gamma_d>\gamma_k,\qquad p\asymp r,\qquad q=\lceil r^{a_q}\rceil\ (a_q>2),\qquad n\asymp\gamma_n qr\ (\gamma_n>0).
\end{equation*}
\end{assumption}
These scalings separate the startup, geometry, and risk windows and control the number of repeated samples.
For example, $k=\lceil2r\rceil$, $d_i=\lceil3r\rceil$, $p=r$, $q=\lceil r^3\rceil$, and $n=qr$ satisfy this assumption.
Initialization bounds use the aspect-ratio gaps. The trajectory and training-set bounds also use $a_q>2$ and the stated sample size.

\paragraph{Group coefficients.}
Define
\begin{equation*}
\begin{gathered}
f:=\frac{\sigma_a^2}{\sigma_a^2+\sigma_o^2},\qquad
\vartheta_a:=\frac{1-\rho_a^2}{2},\qquad
\vartheta_o:=\frac{1-\rho_a\rho_o}{2},\\
\bar\vartheta_1:=f\vartheta_a+(1-f)\vartheta_o,\qquad
\bar\vartheta_2:=f\vartheta_a^2+(1-f)\vartheta_o^2,\qquad
g:=\frac{\bar\vartheta_2}{\bar\vartheta_1}.
\end{gathered}
\end{equation*}

\begin{assumption}[Group-response balance]
\label{ass:rates}
The group parameters satisfy $1/2<g<3/4$.
\end{assumption}

Here $f$ is the limiting within-pretrained response-energy
fraction, while $\vartheta_a<1/2<\vartheta_o$ are the energy fractions retained
by centering in the within-pretrained and mixed channels.
The ratio $g$ averages these retention factors using the
retained energy as weights.
Its lower bound separates the
raw and centered effective-rank transition rates, while its upper bound is
a sufficient restriction for compatibility with greater
centered rotation in suitable spectral regimes.
A concrete group-parameter example is $f=1/5$, $\vartheta_a=1/5$, and $\vartheta_o=7/10$, which gives $g=2/3$.
It is realized by $\rho_a=\sqrt{3/5}$, $\rho_o=-2/(5\rho_a)$, and $\sigma_o^2=4\sigma_a^2$.

\paragraph{Pretrained spectrum.}
Set
\begin{equation*}
M_a:=\frac{r}{d_i}\Sigma_0^2,\qquad
\mu_{s,r}:=\frac1k\tr(M_a^s),\qquad s\ge1.
\end{equation*}

\begin{assumption}[Pretrained spectral dispersion]
\label{ass:spectrum}\label{main:spectrum}
The eigenvalues of $M_a$ lie in a fixed interval
$[\lambda_-,\lambda_+]\subset(0,\infty)$, and their
empirical distribution converges to a law $H$.
Writing $\mu_s=\int\lambda^s\,dH(\lambda)$, assume
\begin{equation}
\mu_1\mu_3-\mu_2^2
\ge C_{\mathrm{sp}}\lambda_+\mu_1^3,
\end{equation}
where $C_{\mathrm{sp}}>0$ is sufficiently large,
depending only on the fixed aspect ratios and group
parameters.
\end{assumption}

Under the energy-weighted distribution
$dH_E(\lambda)=\lambda\,dH(\lambda)/\mu_1$, this condition
is equivalently
$\Var_{H_E}(\lambda)
\ge C_{\mathrm{sp}}\lambda_+\mu_1$.
It requires sufficient spectral dispersion relative to
the mean and maximum spectral scales, without prescribing
a particular spectral density or eigenspace.
Section~\ref{app:spectrum} gives a two-point law satisfying this assumption. Its smallest support point is fixed before the rank grows.
The spectral assumption supplies the coefficient margins, while the dimension and training-set conditions control the sampled trajectories and transfer errors.

\subsection{Parameter choices}
\label{main:theorem}
Write $v_H:=\mu_2/\mu_1^2$ and define
\begin{equation*}
\begin{gathered}
T_C:=\frac{2(v_H+f\vartheta_a/\bar\vartheta_1)}{v_H+\gamma_k},\qquad
T_R:=\frac{2(v_H+f)}{v_H+\gamma_k},\\
u_{\max}:=\frac{1+\min\{2g,4(1-g)\}}2,\qquad
\theta:=\frac{1+\max\{2-u_{\max},u_{\max}/(2g)\}}2,\\
I_H:=(2-\theta,u_{\max}),\qquad
I_L(u_H):=(T_C-u_H,\ \min\{\theta,T_R-u_H\}).
\end{gathered}
\end{equation*}

The following proposition provides one common low/high rate pair for all three observables.
\begin{proposition}[Nonempty learning-rate intervals]
\label{prop:common-parameters}
Under Assumptions~\ref{ass:dimensions}--\ref{ass:spectrum}, $I_H$ and every $I_L(u_H)$ with $u_H\in I_H$ are nonempty.
Every pair chosen from them satisfies
\begin{equation*}
0<u_L<\theta<1<u_H<u_{\max}<T_C,
\qquad T_C<u_L+u_H<T_R<2.
\end{equation*}
\end{proposition}
\begin{proof}
Since $1/2<g<3/4$ and $\bar\vartheta_1>\vartheta_a$,
\begin{equation*}
1<2-\theta<u_{\max}<\min\{2g\theta,4(1-g)\},
\qquad T_C<T_R<2.
\end{equation*}
Moreover $\mu_3\le\lambda_+\mu_2$ and Assumption~\ref{ass:spectrum} give
\begin{equation*}
v_H\ge\frac{\mu_1\mu_3-\mu_2^2}{\lambda_+\mu_1^3}
\ge C_{\mathrm{sp}},\qquad
T_C-u_{\max}
=\frac{(2-u_{\max})v_H+2f\vartheta_a/\bar\vartheta_1-\gamma_k u_{\max}}{v_H+\gamma_k}>0,
\end{equation*}
where it suffices that $C_{\mathrm{sp}}>
\max\{1,(\gamma_k u_{\max}-2f\vartheta_a/\bar\vartheta_1)/(2-u_{\max})\}$.
For $u_H\in I_H$,
\begin{equation*}
0<T_C-u_H<2-u_H<\theta,\qquad
T_C-u_H<T_R-u_H.
\end{equation*}
These inequalities prove the claims.
\end{proof}
Thus the low and high rates can be fixed before the task, initialization, or training set is drawn.
The strict gaps provide room for the first-order and transfer errors.

\paragraph{Parameter choices.}
Fix $u_H\in I_H$ and $u_L\in I_L(u_H)$, and use
\begin{equation*}
\eta_j:=\frac{u_j}{\gamma_k\mu_{1,r}qr},\qquad j=L,H.
\end{equation*}
The finite-dimensional mean $\mu_{1,r}$ retains the original spectral scale.
For this fixed pair, take $C_0,J>0$ sufficiently large and
$c_N,c_{\mathrm{risk}}>0$ sufficiently small, and set
\begin{equation*}
\begin{gathered}
\Lambda_r:=\log(qr+2),\quad
t_0:=\lceil C_0r\sqrt q\,\Lambda_r^J\rceil,\\
N:=\lfloor c_N q\rfloor,\qquad
N_{\mathrm{risk}}:=\lfloor c_{\mathrm{risk}}qr\rfloor,\\
\delta_*(r,q):=\min\left\{
\frac{\Lambda_r^J}{q^{7/2}r\Lambda_r^3},\;
\frac1{q^3r^3\Lambda_r^4}\right\}.
\end{gathered}
\end{equation*}
Fix $0<\delta\le\delta_*(r,q)$ and take $\kappa>0$ sufficiently small.
The choices are deterministic and shared by the three theorems, in the following order:
\begin{enumerate}
\item Fix $(u_L,u_H)$ in the stated intervals.
\item Fix $(C_0,J,c_N,c_{\mathrm{risk}})$ from the first-order margins for this pair.
\item Choose $0<\delta\le\delta_*(r,q)$.
\item Use a common sufficiently small $\kappa>0$.
\end{enumerate}
The constants depend only on the fixed population parameters and the selected rates, while $\delta_*$ additionally has the displayed dependence on $r$ and $q$.

Proposition~\ref{prop:task-radius} verifies this explicit sufficient task radius,
and the end of Section~\ref{app:foundation} verifies the common window choices.

\subsection{Main theorems}
The first theorem identifies the sign of the effective-rank increment for each of the four runs.
\begin{theorem}[Stepwise expected effective rank]
\label{thm:er}
In the setting of Section~\ref{main:setting}, under Assumptions~\ref{ass:dimensions}--\ref{ass:spectrum} and the parameter choices above, for sufficiently large $r$ and every $t_0\le t<N$,
\begin{equation*}
\E[\ER(\Delta W_{R,H,t+1})-\ER(\Delta W_{R,H,t})]\ge\kappa r\eta_H,
\end{equation*}
\begin{equation*}
\E[\ER(\Delta W_{R,L,t+1})-\ER(\Delta W_{R,L,t})]\le-\kappa r\eta_L,
\end{equation*}
\begin{equation*}
\E[\ER(\Delta W_{C,j,t+1})-\ER(\Delta W_{C,j,t})]\le-\kappa r\eta_j,\qquad j=L,H.
\end{equation*}
The expectation includes initialization, task, fixed training set, and sampling histories, including exceptional histories.
\end{theorem}
Since $t_0/N\asymp r\Lambda_r^J/\sqrt q\to0$, the startup is asymptotically shorter than the window.
Summing the increments gives a fixed positive absolute effective-rank change and the change divided by $r$ may vanish.

The next theorem compares the full left singular subspace at a common rate.
\begin{theorem}[Stepwise expected full left singular-subspace rotation]
\label{thm:rotation}
In the setting of Section~\ref{main:setting}, under Assumptions~\ref{ass:dimensions}--\ref{ass:spectrum} and the parameter choices above, for sufficiently large $r$, every $t_0\le t<N$ and either
common rate satisfy
\begin{equation*}
\E\Rot_{C,j}(t,t+1)-\E\Rot_{R,j}(t,t+1)\ge\kappa\eta_j,\qquad j=L,H.
\end{equation*}
The expectation has the same scope as Theorem~\ref{thm:er}.
\end{theorem}
This compares the complete rank-$r$ LoRA left spaces at every adjacent update after startup.

The last theorem determines which of the two rates gives more prediction progress at a fixed update budget.
\begin{theorem}[Opposite learning-rate preferences in ordinary risk]
\label{thm:risk}
In the setting of Section~\ref{main:setting}, under Assumptions~\ref{ass:dimensions}--\ref{ass:spectrum} and the parameter choices above, for sufficiently large $r$, there is an event of
probability $1-o(1)$ over initialization, task, and training set such that, conditional on every realization in that event, for every $0\le t<N_{\mathrm{risk}}$, the per-update claims are
\begin{equation*}
\E_{\mathrm{hist}}[\mathcal R_\delta(W_{m,j,t+1})-\mathcal R_\delta(W_{m,j,t})]<0,
\qquad m\in\{R,C\},\quad j\in\{L,H\}.
\end{equation*}
\begin{equation*}
\begin{aligned}
&\E_{\mathrm{hist}}[\mathcal R_\delta(W_{R,L,t+1})-\mathcal R_\delta(W_{R,L,t})]\\
&\qquad-\E_{\mathrm{hist}}[\mathcal R_\delta(W_{R,H,t+1})-\mathcal R_\delta(W_{R,H,t})]
\ge\kappa\delta^2/q.
\end{aligned}
\end{equation*}
\begin{equation*}
\begin{aligned}
&\E_{\mathrm{hist}}[\mathcal R_\delta(W_{C,H,t+1})-\mathcal R_\delta(W_{C,H,t})]\\
&\qquad-\E_{\mathrm{hist}}[\mathcal R_\delta(W_{C,L,t+1})-\mathcal R_\delta(W_{C,L,t})]
\ge\kappa\delta^2/q.
\end{aligned}
\end{equation*} At every $1\le t\le N_{\mathrm{risk}}$,
\begin{equation*}
\E_{\mathrm{hist}}\mathcal R_\delta(W_{R,L,t})-\E_{\mathrm{hist}}\mathcal R_\delta(W_{R,H,t})\ge\kappa\delta^2t/q,
\end{equation*}
\begin{equation*}
\E_{\mathrm{hist}}\mathcal R_\delta(W_{C,H,t})-\E_{\mathrm{hist}}\mathcal R_\delta(W_{C,L,t})\ge\kappa\delta^2t/q.
\end{equation*}
\end{theorem}
The risk prefix contains the full geometry window for large $r$.
These finite-budget results establish joint effective-rank, rotation, and risk properties under the same assumptions.

Section~\ref{app:foundation} assembles the proofs from three first-order propositions and the transfer estimate.
Theorems~\ref{thm:er}, \ref{thm:rotation}, and~\ref{thm:risk} respectively formalize the three items of Theorem~\ref{thm:lora-joint-main}.

\section{Complete Proofs}\label{app:proofs}
\subsection{Proof architecture}\label{app:foundation}
Section~\ref{app:reduction} reduces SGD to a response linear in the task amplitude. Sections~\ref{app:forcing-moments}--\ref{app:training-set} supply its moments, spectral margins, and one static training-set event. Sections~\ref{app:er}--\ref{app:risk} prove the first-order propositions below. Their transfer gives the three theorems at the end of this subsection. Section~\ref{app:tools} collects the standard tools.

\paragraph{Conventions.}
Matrices are uppercase, vectors and scalars lowercase, and calligraphic letters denote matrix-space maps, events, or index sets. A tilde marks Euclidean coordinates, superscript $0$ the first-order response, and a bar an average (over stored pairs for $G,\mathcal H$ and over histories for $\Gamma_t$). Superscript pop denotes a population mean. Indices $(m,j,i,\alpha,t)$ identify rule, rate, stored pair, position, and update. Run subscripts have order $(m,j,t)$. Constants $c,C$ depend only on the fixed population parameters and selected rates and may change between occurrences, and $C_x$ may also depend on $x$. Chosen window constants and margins remain fixed.

The norms $\|\cdot\|,\|\cdot\|_F,\|\cdot\|_*$ are operator, Frobenius, and nuclear norms, with $\langle A,B\rangle_F:=\tr(A^\top B)$. Map norms are induced by the Frobenius norm. The notation $O_{\op},O_F,O_*$ specifies the bounded matrix norm, whereas $O_{L^p}$ specifies the probability norm. The symbols $\sigma_{\min},\sigma_{\max}$ denote extreme singular values. Write $\|A\|_{S_p}:=(\sum_i\sigma_i(A)^p)^{1/p}$, $|A|:=(A^2)^{1/2}$ for symmetric $A$, and $\operatorname{sym}(A):=(A+A^\top)/2$. The factor $\mathcal L_r$ is a fixed power of $\Lambda_r$, whose exponent may increase in finitely many estimates. The startup exponent $J$ is chosen larger.

Unqualified $\E,\Prob$ average initialization, task, training set, and sampled history. The subscripts hist, $I$, train, fresh, and $Z$ respectively average the history, the next uniform stored-pair index, the training-set draw, an independent pair, and the task, holding the other arguments fixed. A different scope is stated locally. A tensor $A\otimes B$ acts between the compatible matrix spaces by $(A\otimes B)[D]:=A\langle B,D\rangle_F$. This convention identifies the second-moment tensors with operators on Frobenius matrix space.

\begin{center}
\begin{tabular}{p{.25\textwidth}p{.43\textwidth}p{.24\textwidth}}
\toprule
Symbol & Meaning & Definition site\\\midrule
$\Pi_m$ & Raw or centered position projection & Section~\ref{app:reduction}\\
$L_t^0,Y_t$ & Euclidean and rescaled first-order responses & Lemma~\ref{tr:linearization}, \eqref{er:operators}\\
$X_{i,\alpha},y_{i,\alpha}$ & Rescaled feature and quadratic label & \eqref{eq:feature-label}\\
$G_i,\mathcal H_i$ & Per-pair forcing and feedback & \eqref{er:operators}\\
$S,T$ & Response metric and adapter Gram metric & Section~\ref{app:reduction}\\
$\Gamma_t,\operatorname{er}_T$ & Normalized response Gram and entropy rank & Section~\ref{er:coordinates}\\
$\Gamma^{(0)},\Gamma^{(1)},\Gamma^{\mathrm{ang}}$ & Mean-Gram coefficients and angular drift & \eqref{er:gram-drift}, \eqref{rr:angular-coefficient}\\
$\Psi,p_R,p_C$ & Rotation gap and risk polynomials & Sections~\ref{app:spectrum}, \ref{app:foundation}\\
\bottomrule
\end{tabular}
\end{center}

\paragraph{Events.}
The static event is constructed without sampling histories. For every fixed confidence exponent $M>0$, the constants in its construction may depend on $M$. A sufficiently large exponent is fixed before the final union bounds.
\begin{center}
\begin{tabular}{p{.17\textwidth}p{.35\textwidth}p{.17\textwidth}p{.21\textwidth}}
\toprule
Event & Randomness or construction & Failure bound & Reference\\\midrule
$\mathcal E_{\mathrm{init}}$ & Initialization and task & $C_M(qr)^{-M}$ & Lemma~\ref{er:init}\\
$\mathcal E_\star\subset\mathcal E_{\mathrm{init}}$ & Initialization, task, training set & $C_M(qr)^{-M}$ & Lemma~\ref{er:training-set}\\
$\mathcal B_t$ & Static event and past Gram bounds & $C_M(qr)^{-M}$ & Lemma~\ref{er:state}\\
$\mathcal E_{\mathrm{good}}$ & Four runs, common geometry window & $o(\eta_L)$ & Transfer proof below\\
$\mathcal E_{\mathrm{risk}}$ & Initialization, task, training set & $o(1)$ & Proposition~\ref{rr:risk}\\
\bottomrule
\end{tabular}
\end{center}

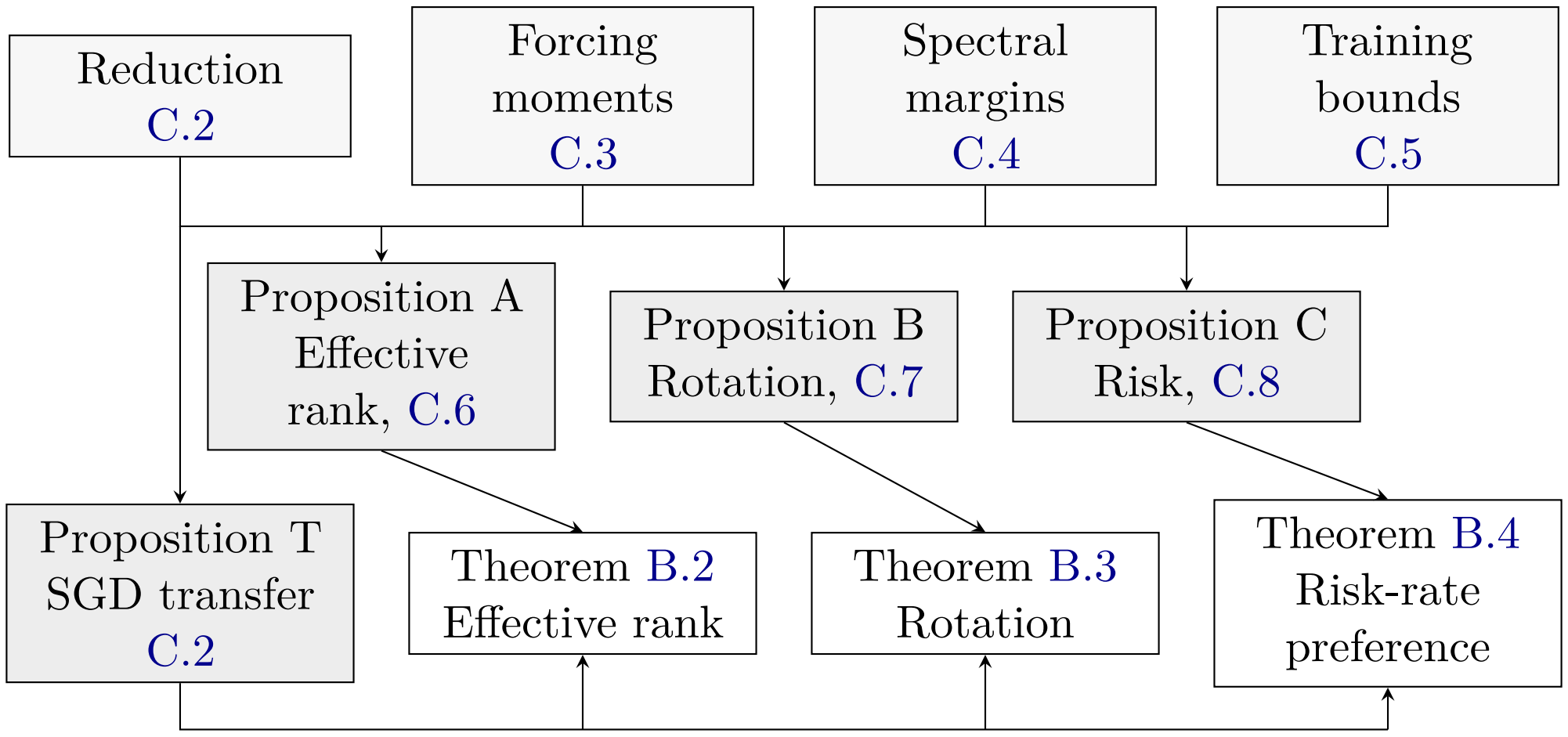
\begin{figure}[ht]
\centering
\begin{tikzpicture}[>=stealth,
  input/.style={draw,fill=black!3,align=center,text width=2.6cm,minimum height=.85cm,inner sep=4pt},
  result/.style={draw,fill=black!7,align=center,text width=2.65cm,minimum height=.95cm,inner sep=4pt},
  conclusion/.style={draw,align=center,text width=2.65cm,minimum height=.95cm,inner sep=4pt}]
\node[input] (reduction) at (-5.1,0) {Reduction\\\ref{app:reduction}};
\node[input] (moments) at (-1.7,0) {Forcing moments\\\ref{app:forcing-moments}};
\node[input] (spectrum) at (1.7,0) {Spectral margins\\\ref{app:spectrum}};
\node[input] (training) at (5.1,0) {Training bounds\\\ref{app:training-set}};
\draw (reduction.south)--(-5.1,-1.1)--(5.1,-1.1)--(training.south);
\draw (moments.south)--(-1.7,-1.1);
\draw (spectrum.south)--(1.7,-1.1);
\node[result] (a) at (-3.4,-2.2) {Proposition A\\Effective rank, \ref{app:er}};
\node[result] (b) at (0,-2.2) {Proposition B\\Rotation, \ref{app:rotation}};
\node[result] (c) at (3.4,-2.2) {Proposition C\\Risk, \ref{app:risk}};
\draw[->] (-3.4,-1.1)--(a.north);
\draw[->] (0,-1.1)--(b.north);
\draw[->] (3.4,-1.1)--(c.north);
\node[result] (transfer) at (-5.1,-4.2) {Proposition T\\SGD transfer\\\ref{app:reduction}};
\node[conclusion] (er) at (-1.7,-4.2) {Theorem~\ref{thm:er}\\Effective rank};
\node[conclusion] (rotation) at (1.7,-4.2) {Theorem~\ref{thm:rotation}\\Rotation};
\node[conclusion] (risk) at (5.1,-4.2) {Theorem~\ref{thm:risk}\\Risk-rate preference};
\draw[->] (-5.1,-1.1)--(transfer.north);
\draw[->] (a.south)--(er.north);
\draw[->] (b.south)--(rotation.north);
\draw[->] (c.south)--(risk.north);
\draw (transfer.south)--(-5.1,-5.35)--(5.1,-5.35);
\draw[->] (-1.7,-5.35)--(er.south);
\draw[->] (1.7,-5.35)--(rotation.south);
\draw[->] (5.1,-5.35)--(risk.south);
\end{tikzpicture}
\caption{Proof roadmap. The top row supplies the common response, moment, spectral, and static-event estimates. Sections~\ref{app:er}--\ref{app:risk} prove Propositions A--C; the rotation proof also reuses the trajectory estimates of Section~\ref{app:er}. Combining each first-order comparison with Proposition T proves the corresponding theorem in this subsection.}
\label{fig:proof-roadmap}
\end{figure}

The shared estimates below feed three first-order comparisons. Proposition T then transfers their margins to simultaneous-factor SGD.
Figure~\ref{fig:proof-roadmap} separates these two stages.

\subsubsection{First-order propositions and transfer}
Fix the population assumptions and a common rate pair from Proposition~\ref{prop:common-parameters}. The first-order margins determine the windows before the task radius is applied.

Proposition A is the common effective-rank increment estimate established in Section~\ref{app:er}.
\begin{proposition}[A: First-order effective-rank increments]\label{prop:first-order-er}
Under Assumptions~\ref{ass:dimensions}--\ref{ass:spectrum} and a fixed rate pair selected by Proposition~\ref{prop:common-parameters}, choose the startup and geometry-window constants as in
Section~\ref{main:theorem}.
For sufficiently large $r$, there is a fixed $\kappa_{\mathrm{ER}}>0$ such that, for every $t_0\le t<N$,
\begin{equation*}
\varsigma_{m,j}\E\bigl[\operatorname{er}_T(\Gamma_{m,j,t+1})-
\operatorname{er}_T(\Gamma_{m,j,t})\bigr]\ge\kappa_{\mathrm{ER}}\eta_j,
\end{equation*}
where $\varsigma_{R,H}:=1$ and $\varsigma_{R,L}:=\varsigma_{C,L}:=\varsigma_{C,H}:=-1$.
The expectation includes initialization, task, training set, and all sampling histories.
\end{proposition}
The sign differs only for the raw high-rate run.
Multiplying by $r$ and controlling the approximation at both endpoints gives the effective-rank theorem for SGD.

The next proposition turns the positive angular coefficient into an adjacent-update rotation gap.
\begin{proposition}[B: Left singular-subspace rotation at every update]\label{rr:rotation-step}
Suppose $\Psi(u_H;h_M)>0$ with a fixed margin and $0<u_L<u_H<2$.
For sufficiently small fixed $c_N>0$, there is a constant $\kappa_{\mathrm{rot}}>0$ such that, for sufficiently large
$r$ and every $t_0\le t<N$, the first-order response at either common rate satisfies
\begin{equation*}
\E\operatorname{rot}^{0}_{C,j}(t,t+1)-\E\operatorname{rot}^{0}_{R,j}(t,t+1)
\ge\kappa_{\mathrm{rot}}\eta_j,\qquad j=L,H.
\end{equation*}
\end{proposition}
The startup and endpoint are the common geometry window of Section~\ref{main:theorem}.
This comparison concerns the full rank-$r$ first-order left singular space at either selected rate.

The first-order prediction risk, with the response map $\mathcal P$ defined in
Section~\ref{app:reduction}, is
\begin{equation}\label{rr:first-order-risk}
\mathcal R^0(L):=\frac12\|E-\mathcal P(L)\|_F^2.
\end{equation}
For a numerical dimensionless rate $u$, define
\begin{equation}\label{rr:risk-polynomials}
\begin{aligned}
p_R(u)&:=2\gamma_d(\sigma_a^2+\sigma_o^2)\mu_1(v_H+\gamma_k)u(u-T_R),\\
p_C(u)&:=2\gamma_d(\sigma_a^2+\sigma_o^2)\mu_1\bar\vartheta_1(v_H+\gamma_k)u(u-T_C).
\end{aligned}
\end{equation}
Here $T_R,T_C$ are the thresholds of Section~\ref{main:theorem}, and a run $(m,j)$ uses $u_j$.

The following proposition identifies the first-order risk increment and its uniform persistence.
\begin{proposition}[C: Risk polynomials and their persistence]\label{rr:risk}
For fixed rates $0<u_L<u_H<2$ and a fixed horizon constant $c_{\mathrm{risk}}>0$, there is an event of probability $1-o(1)$ over the initialization, teacher, and training set on which,
uniformly for the four predetermined runs and $0\le t\le c_{\mathrm{risk}}qr$,
\begin{equation*}
\E_{\mathrm{hist}}[{\mathcal R^0} (L_{m,j,t+1}^0)-{\mathcal R^0} (L_{m,j,t}^0)]=q^{-1}\{p_m(u_j)+o(1)+O(t/(qr))\}.
\end{equation*}
If additionally $u_H<T_C$ and $T_C<u_L+u_H<T_R$ hold with fixed strict margins, there exist $T_{\max},c_*>0$ such that, for every fixed $0<c_{\mathrm{risk}}\le T_{\max}$ and all
sufficiently large $r$, all four increments on this event are at most $-c_*/q$, the raw high-rate decrement exceeds the raw low-rate decrement by at least $c_*/q$, and the centered
low-rate decrement exceeds the centered high-rate decrement by at least $c_*/q$.
On the same event, for every $1\le t\le c_{\mathrm{risk}}qr$,
\begin{equation*}
\E_{\mathrm{hist}}{\mathcal R^0} (L_{R,L,t}^0)-\E_{\mathrm{hist}}{\mathcal R^0} (L_{R,H,t}^0)\ge c_*t/q,
\end{equation*}
\begin{equation*}
\E_{\mathrm{hist}}{\mathcal R^0} (L_{C,H,t}^0)-\E_{\mathrm{hist}}{\mathcal R^0} (L_{C,L,t}^0)\ge c_*t/q.
\end{equation*}
\end{proposition}
The strict margins make all four risks decrease and separate the preferred rates over the same update budget.
The event in this proposition is denoted $\mathcal E_{\mathrm{risk}}$.

\subsubsection{Choice of constants}

Fix a rate pair from Section~\ref{main:theorem}.
Lemma~\ref{sp:moment-rates} supplies strict population margins.
The first-order estimates in Lemmas~\ref{er:mean-phase} and~\ref{er:compensation} and Propositions~\ref{rr:rotation-step} and~\ref{rr:risk} depend only on these margins and the explicit
rate/window hypotheses.
Choose $J,C_0$ large and $c_N,c_{\mathrm{risk}}$ small so their uniform remainders consume at most one quarter of the corresponding margins.
Denote the remaining positive constants by $\kappa_{\mathrm{ER}},\kappa_{\mathrm{rot}},\kappa_{\mathrm{risk}}$ and set
\begin{equation*}
\kappa:=\tfrac18\min\{\kappa_{\mathrm{ER}},\kappa_{\mathrm{rot}},\kappa_{\mathrm{risk}}\}>0.
\end{equation*}

The next proposition makes the task radius explicit after the first-order margins have been fixed.
\begin{proposition}[T: Validity of the task radius]
\label{prop:task-radius}
Fix the rate pair and windows in Section~\ref{main:theorem}.
Under the model assumptions, deterministic norm bounds, bounds on $\|y_i\|$, and a fixed $\zeta>0$ satisfying $\eta_H\|\widetilde{\mathcal H}_i\|\le2-\zeta$ for every training pair can be chosen
with failure probability $O((qr)^{-M})$ for any fixed $M>0$ such that the following holds for all sufficiently large $r$.
For any fixed $a_*,C_{\mathrm{ER}},C_{\mathrm{rot}}>0$ and positive margins
$\kappa_{\mathrm{ER}},\kappa_{\mathrm{rot}},\kappa_{\mathrm{risk}}$,
with a fixed $c_0>0$ chosen as in Lemma~\ref{tr:coupling}, every $0<\delta\le\delta_*(r,q)$ as defined in Section~\ref{main:theorem} satisfies
\begin{equation}\label{eq:task-radius}
\begin{gathered}
\max\{{\mathsf{err}_{\mathrm{fac}}} ^{\mathrm G},{\mathsf{err}_{\mathrm{fac}}} ^{\mathrm R}\}\le c_0/2,\qquad {\mathsf{err}_{\mathrm{Gram}}}
,{\mathsf{err}_{\mathrm{joint}}} \le a_*/4,\\
2C_{\mathrm{ER}}{\mathsf{err}_{\mathrm{Gram}}} \le\kappa_{\mathrm{ER}}\eta_L/4,\qquad
2C_{\mathrm{rot}}{\mathsf{err}_{\mathrm{joint}}} \le\kappa_{\mathrm{rot}}\eta_L/4,\qquad
4{\mathsf{err}_{\mathrm{risk}}} \le\kappa_{\mathrm{risk}}/(4q).
\end{gathered}
\end{equation}
Consequently, Lemma~\ref{tr:coupling} bounds the factor approximation through both $N$ and $N_{\mathrm{risk}}$.
For geometry times $t,t'\in[t_0,N]$ and a fixed $b_*>a_*$ such that $\Gamma_t,\Gamma_{t\prime}$ have spectra in $[a_*,b_*]$, Lemma~\ref{tr:observable}
bounds the effective-rank and left singular-subspace overlap errors by $C_{\mathrm{ER}}{\mathsf{err}_{\mathrm{Gram}}} $ and $C_{\mathrm{rot}}{\mathsf{err}_{\mathrm{joint}}} $, respectively.
Its risk error bound ${\mathsf{err}_{\mathrm{risk}}} $ holds through $N_{\mathrm{risk}}$ for every sampling history, without a Gram-matrix condition.
\end{proposition}

The following estimate collects the exceptional-event and independent-truncation arguments used below.
\begin{lemma}[Polynomial-tail absorption and per-pair truncation]\label{lem:tail-absorption}
Suppose $\Prob(\mathcal A_r^c)\le C_M(qr)^{-M}$ for a selectable fixed exponent $M$. If $|Z_r|\le C(qr)^b$ on an event $\mathcal E$, then
\begin{equation*}
\E[\mathbf1_{\mathcal E}|Z_r|\mathbf1_{\mathcal A_r^c}]\le C_M(qr)^{b-M}.
\end{equation*}
Alternatively, if $\|Z_r\|_{L^{2p}}\le C_p(qr)^b$ for a fixed $p>1/2$, then
\begin{equation*}
\E[|Z_r|\mathbf1_{\mathcal A_r^c}]\le C_{p,M}(qr)^{b-M(1-1/(2p))}.
\end{equation*}
Thus either bound is $O((qr)^{-K})$ for any prescribed fixed $K$ by choosing the confidence exponent sufficiently large. Taking $K=2$ gives $o(\eta_L)$. The same conclusion holds conditionally whenever the stated bounds hold uniformly in the conditioning variables. In particular, truncating each independent stored pair separately at $\|x_{i,\alpha}\|\le C_M\sqrt q$, $\|Q^\top x_{i,\alpha}\|\le C_M\sqrt r$, and $|y_{i,\alpha}|\le C_M\sqrt{r\Lambda_r}$ changes expectations of fixed-degree polynomial functionals with polynomial moment bounds by an arbitrarily small polynomial order, within the initialization event where those bounds hold.
\end{lemma}
\begin{proof}
The first inequality follows by multiplying by the indicator, and the second is H\"older's inequality with exponents $2p$ and $2p/(2p-1)$. 

Gaussian norm and quadratic-form tails, followed by a union bound over the polynomial number of pairs, give the truncation failure bound. The degree is fixed, so Gaussian moments give the required polynomial $L^{2p}$ bounds. Apply the preceding estimates to the difference of the original and truncated functionals, which vanishes when no pair is truncated. Truncating each pair separately preserves independence across pairs. Finally $\eta_L\asymp(qr)^{-1}$.
\end{proof}

\subsubsection{Transfer and the three theorem proofs}
The transfer argument uses the same deterministic task radius for all four runs.
The checklist records the first-order input and the total approximation error in each comparison.
The geometry failure bounds include all four runs and the common window. The risk event is static, with a bound uniform over histories.
\begin{center}
\begingroup
\renewcommand{\arraystretch}{1.25}
\begin{tabular}{@{}p{.15\textwidth}p{.21\textwidth}p{.31\textwidth}p{.25\textwidth}@{}}
\toprule
Observable & First-order input & Comparison error & Exceptional probability\\
\midrule
Effective rank $\ER/r$ & Proposition A (\ref{prop:first-order-er}) & $2C_{\mathrm{ER}}\mathsf{err}_{\mathrm{Gram}}$ & $o(\eta_L)$\\
Rotation & Proposition B (\ref{rr:rotation-step}) & $2C_{\mathrm{rot}}\mathsf{err}_{\mathrm{joint}}$ & $o(\eta_L)$\\
Risk-rate gap & Proposition C (\ref{rr:risk}) & $4\mathsf{err}_{\mathrm{risk}}$ after division by $\delta^2$ & $o(1)$ over initialization, task, and training set\\
\bottomrule
\end{tabular}
\endgroup
\end{center}

\begin{proof}[Proofs of Theorems~\ref{thm:er}, \ref{thm:rotation}, and~\ref{thm:risk}]
\textit{Theorem~\ref{thm:er} (Effective rank).}
Let $\mathcal E_{\mathrm{good}}$ be the intersection of the required norm and Gram eigenvalue events for the four runs and every $t_0\le t\le N$.
The trajectory estimates and a union bound give $\Prob(\mathcal E_{\mathrm{good}}^c)=o(\eta_L)$.
Proposition~\ref{prop:task-radius} and Lemma~\ref{tr:observable} bound the normalized rank error by $C_{\mathrm{ER}}\mathsf{err}_{\mathrm{Gram}}$ at each endpoint on this event.
Both normalized ranks lie in $[1/r,1]$, including exceptional histories.
Consequently Proposition~\ref{prop:first-order-er} gives
\begin{equation*}
\begin{aligned}
\frac{\varsigma_{m,j}}r\E[\ER(\Delta W_{m,j,t+1})-\ER(\Delta W_{m,j,t})]
&\stackrel{(i)}{\ge}\kappa_{\mathrm{ER}}\eta_j-2C_{\mathrm{ER}}\mathsf{err}_{\mathrm{Gram}}-2\Prob(\mathcal E_{\mathrm{good}}^c)\\
&\stackrel{(ii)}{\ge}\kappa_{\mathrm{ER}}\eta_j-\kappa_{\mathrm{ER}}\eta_L/4-o(\eta_L)
\ge\kappa\eta_j.
\end{aligned}
\end{equation*}
Here (i) applies the first-order estimate and both endpoint errors, while (ii) uses \eqref{eq:task-radius}, $\eta_L\le\eta_j$, and $\kappa_{\mathrm{ER}}\ge8\kappa$.
This proves all signs in Theorem~\ref{thm:er}.

\textit{Theorem~\ref{thm:rotation} (Left singular-subspace rotation).}
Lemma~\ref{tr:coupling} gives $\sigma_{\min}(A_t^\top)\ge c_0>0$, so $\rank A_t=r$ and $\operatorname{col}(B_tA_t)=\operatorname{col}(B_t)$.
On $\mathcal E_{\mathrm{good}}$, Lemma~\ref{tr:observable} bounds each rotation error by $C_{\mathrm{rot}}\mathsf{err}_{\mathrm{joint}}$.
Since rotations lie in $[0,1]$, Proposition~\ref{rr:rotation-step} and \eqref{eq:task-radius} imply
\begin{equation*}
\E\Rot_{C,j}(t,t+1)-\E\Rot_{R,j}(t,t+1)
\ge\kappa_{\mathrm{rot}}\eta_j-2C_{\mathrm{rot}}\mathsf{err}_{\mathrm{joint}}-o(\eta_L)
\ge\kappa\eta_j.
\end{equation*}
This proves Theorem~\ref{thm:rotation}.

\textit{Theorem~\ref{thm:risk} (Excess risk).}
Intersect $\mathcal E_{\mathrm{risk}}$ with the norm event of Proposition~\ref{prop:task-radius}. The resulting event still has probability $1-o(1)$ over initialization, task, and training set.
On each realization in this event, the transfer error bound holds for every history through $N_{\mathrm{risk}}$ and therefore remains valid after history averaging.
The error in a normalized history-averaged increment is at most $2\mathsf{err}_{\mathrm{risk}}$, and the error in comparing two such increments is at most $4\mathsf{err}_{\mathrm{risk}}$.
Proposition~\ref{rr:risk} and \eqref{eq:task-radius} therefore give, respectively,
\begin{equation*}
-\frac{\kappa_{\mathrm{risk}}}{q}+2\mathsf{err}_{\mathrm{risk}}\le-\frac\kappa q,
\qquad
\frac{\kappa_{\mathrm{risk}}}{q}-4\mathsf{err}_{\mathrm{risk}}\ge\frac\kappa q.
\end{equation*}
The first inequality proves descent for all four runs, and the second proves the raw high-rate and centered low-rate preferences.
Multiplying by $\delta^2$ and summing from the common initial weight $W_{m,j,0}=W_0$ yields the two cumulative inequalities in Theorem~\ref{thm:risk}.
\end{proof}

\paragraph{Verification of the common windows.}
The fixed choice $a_q>2$ gives
\begin{equation*}
\frac{t_0}{N}\asymp\frac{r\Lambda_r^J}{\sqrt q}\longrightarrow0,
\qquad
\frac{N}{N_{\mathrm{risk}}}\asymp\frac{c_N}{c_{\mathrm{risk}}r}\longrightarrow0.
\end{equation*}
Moreover, $r\eta_jq=u_j/(\gamma_k\mu_{1,r})\to u_j/(\gamma_k\mu_1)>0$, so $\liminf_{r\to\infty}r\eta_j(N-t_0)>0$.
Summing the effective-rank increments proves the positive absolute change after Theorem~\ref{thm:er}. Its change divided by $r$ may vanish.

\subsection{From SGD to the first-order response}\label{app:reduction}
This subsection reduces the three observables to the first-order response and proves the transfer bounds used by Proposition~\ref{prop:task-radius}.
The coupling is uniform over sampling histories. The geometry conclusions additionally require the Gram conditions stated below.
The rule and rate are fixed and suppressed, and constants follow Section~\ref{app:foundation}.

\subsubsection{First-order task response and adapter observables.}

Put $Z:=U_0Z_a+U_{\perp}Z_o$, so the output-width representation $W_0+\delta ZV_0^\top$ has the same Gram matrix as the teacher.
Define
\begin{equation*}
\begin{gathered}
E:=U_0\Sigma_0Z^\top+Z\Sigma_0U_0^\top,\qquad \Phi(x):=\frac{xx^\top-I}{\sqrt2},\\
W_0A_0^{\top}=QS^{1/2},\quad S:=A_0W_0^\top W_0A_0^\top,\quad T:=S^{1/2}(A_0 A_0^{\top})S^{1/2}.
\end{gathered}
\end{equation*}
This is the thin polar decomposition of $W_0A_0^\top$, with $Q^\top Q=I_r$ and $S^{1/2}$ positive definite almost surely.
The Gaussian initialization and $k>r$
guarantee full column rank.
For symmetric $D$, write $\phi_D(x):=x^\top Dx-\tr D=\sqrt2\langle D,\Phi(x)\rangle_F$.
The matrix $T$ is the Gram matrix of
$S^{1/2}A_0$.
For any matrix $Y$, the nonzero squared singular values of $YS^{1/2}A_0$ are the nonzero eigenvalues of $T^{1/2}Y^\top YT^{1/2}$.
It is generally
correlated with $S$.

For $\delta>0$, write $L_t^\delta:=B_t/\delta$, and denote the first-order response by $L_t^0:=\partial_\delta B_t|_{\delta=0}$.
For fixed $t$ and sampling
history, this means $B_t(\delta)=\delta L_t^0+O_F(\delta^2)$.
The order refers to the task amplitude $\delta$.
Define $\mathcal P(L):=W_0A_0^\top
L^\top+LA_0W_0^\top$.
For training pair $i$ and position $\alpha\in\{+,-\}$, define the rescaled feature and quadratic label by
\begin{equation}\label{eq:feature-label}
X_{i,\alpha}:=(x_{i,\alpha}x_{i,\alpha}^{\top}-I)Q,
\qquad y_{i,\alpha}:=\sqrt2\langle E,\Phi(x_{i,\alpha})\rangle_F.
\end{equation}
Fix a loss rule ${m} \in\{R,C\}$, where $R$ denotes the raw loss and $C$ the residual-centered loss.
Its two-position coefficient matrix is
\begin{equation*}
\Pi _R:=I_2,\qquad \Pi _C:=\tfrac12
\begin{pmatrix}1&-1\\-1&1\end{pmatrix},
\end{equation*}
Rows and columns are ordered as $(+,-)$.
Set $\Pi:=\Pi_m$ and $\Pi_{\alpha\beta}:=(\Pi_m)_{\alpha\beta}$.
These are entries of the selected loss matrix.
For $\varepsilon _{i,\alpha}(W):=f_W(x_{i,\alpha})-f_{F_\delta}(x_{i,\alpha})$ and $\varepsilon _i:=(\varepsilon _{i,+},\varepsilon _{i,-})^\top$, the losses in
\eqref{eq:losses} are $\ell_{i,{m} }=\frac14\varepsilon _i^\top \Pi _{m} \varepsilon _i$.
Define the sample Hessian $\widetilde{\mathcal H}_i$ and the additive
forcing matrix $\widetilde G_i$ by
\begin{equation*}
\begin{aligned}
\widetilde{\mathcal H}_i[L]&:=\tfrac12\sum_{\alpha,\beta\in\{+,-\}}
\Pi _{\alpha\beta}\langle L,\sqrt2\, X_{i,\beta}S^{1/2}\rangle_F\sqrt2\, X_{i,\alpha}S^{1/2},\\
\widetilde G_i&:=\tfrac12\sum_{\alpha,\beta\in\{+,-\}}
\Pi _{\alpha\beta}\langle E,\Phi(x_{i,\beta})\rangle_F\sqrt2\, X_{i,\alpha}S^{1/2}.
\end{aligned}
\end{equation*}
The loss rule is fixed in each calculation, so its index ${m} $ is suppressed in $\widetilde{\mathcal H}_i$ and $\widetilde G_i$.
When a single pair is fixed,
we also write $x_\alpha,\sqrt2\, X_\alpha S^{1/2},\varepsilon _\alpha,\ell_i$ for $x_{i,\alpha},\sqrt2\, X_{i,\alpha}S^{1/2},\varepsilon
_{i,\alpha},\ell_{i,{m} }$.
For a coefficient vector $v\in\operatorname{ran}\Pi$, define the feature maps
\begin{equation}\label{eq:feature-maps}
\mathcal X_i[v]:=\sum_{\alpha}X_{i,\alpha}v_\alpha,
\qquad \widetilde{\mathcal X}_i[v]:=\mathcal X_i[v]S^{1/2}.
\end{equation}
Their adjoints use the Frobenius inner product and the Euclidean inner product on
$\operatorname{ran}\Pi$. In particular,
$\widetilde{\mathcal X}_i^{*}[L]=\Pi(\langle X_{i,\alpha}S^{1/2},L\rangle_F)_\alpha$.
Thus $\widetilde{\mathcal H}_i=\widetilde{\mathcal X}_i\widetilde{\mathcal X}_i^{*}$
and $\widetilde G_i=\tfrac12\widetilde{\mathcal X}_i[\Pi y_i]$,
where $y_i:=(y_{i,+},y_{i,-})^\top$.
The rescaled response is $Y_t=L_t^0S^{-1/2}$.
For a nonzero positive semidefinite $\Gamma$, let $\operatorname{er}_T(\Gamma)$ be the entropy rank divided by
$r$, computed from the eigenvalues of $T^{1/2}\Gamma T^{1/2}$ normalized by their sum.
Let $\operatorname{ovlp}_L^0(t,t')$ denote the left singular-subspace
overlap of the first-order updates $L_t^0A_0,L_{t'}^0A_0$, with the completion convention from Section~\ref{main:setting}.
For full-column-rank $Y_t$, write
$P_{Y_t}:=Y_t(Y_t^\top Y_t)^{-1}Y_t^\top$ for its orthogonal projector.

The next lemma identifies the first derivative of the factor dynamics with respect to the task amplitude.
\begin{lemma}[Exact first-order response]
\label{tr:linearization}
For either loss rule, a fixed finite training set, and any fixed finite sampling history, the first-order response exists and satisfies
\begin{equation*}
L_{t+1}^0=(I-\eta \widetilde{\mathcal H}_{I_t})L_t^0+\eta\widetilde G_{I_t},\qquad L_0^0=0,
\end{equation*}
and each $\widetilde{\mathcal H}_i$ is positive semidefinite. \end{lemma}

\begin{proof}
At $B=\delta L$, the exact scaled residual is
\begin{equation*}
\frac{\varepsilon _\alpha(W_0+\delta LA)}{\delta}=\left\langle W_0A^{\top}L^\top+LA W_0^\top-E+\delta\{LA A^{\top}L^\top-ZZ^\top\},\Phi(x_\alpha)\right\rangle_F.
\end{equation*}
At $\delta=0,A=A_0$, differentiation gives
\begin{align*}
\nabla_L\{\varepsilon _\alpha(W_0+\delta LA)/\delta\}&=\sqrt2\, X_\alpha S^{1/2},\\
\nabla_L\left\{\frac14\sum_{\alpha,\beta}\Pi _{\alpha\beta}\langle\mathcal P(L)-E,\Phi(x_\alpha)\rangle_F\langle\mathcal
P(L)-E,\Phi(x_\beta)\rangle_F\right\}&=\widetilde{\mathcal H}_i[L]-\widetilde G_i,\\
\nabla_A\ell_i&=O(\delta^2),\qquad \partial_\delta A_t\big|_{\delta=0}=0.
\end{align*}

Induction in the factor updates yields the affine recursion.
Moreover,
\begin{equation*}
\langle L,\widetilde{\mathcal H}_i[L]\rangle_F=\tfrac12\sum_{\alpha,\beta}\Pi _{\alpha\beta}\langle L,\sqrt2\, X_\alpha S^{1/2}\rangle_F\langle
L,\sqrt2\, X_\beta S^{1/2}\rangle_F\ge0.
\end{equation*}
\end{proof}

The following identities turn the response matrix into the effective-rank and overlap quantities used in the theorem proofs.
\begin{lemma}[First-order adapter observables]\label{tr:first-order-observables}
Under the hypotheses of Lemma~\ref{tr:linearization}, for $\eta>0$ and every $t\ge1$ with $Y_t\ne0$,
\begin{equation*}
\frac1r\ER(L_t^0A_0)
=\operatorname{er}_T(\Gamma_t),\qquad \Gamma_t=\frac{Y_t^\top Y_t}{t\eta},\quad t\ge1,
\end{equation*}
and $\operatorname{col}(L_t^0A_0)=\operatorname{col}(Y_t)$.
Whenever $Y_t$ and $Y_{t'}$ both have full column rank,
\begin{equation*}
\operatorname{ovlp}_L^0(t,t')=\frac1r\tr(P_{Y_t}P_{Y_{t\prime}}).
\end{equation*}
\end{lemma}
\begin{proof}
Since $L_t^0=Y_tS^{1/2}$ and $T:=S^{1/2}(A_0 A_0^{\top})S^{1/2}$,
\begin{align*}
\operatorname{spec}_{>0}\{(L_t^0A_0)(L_t^0A_0)^\top\}&=\operatorname{spec}_{>0}\{T^{1/2}Y_t^\top Y_tT^{1/2}\},\\
\operatorname{er}_T(c\Gamma )&=\operatorname{er}_T(\Gamma )\quad(c>0),\qquad \operatorname{col}(L_t^0A_0)=\operatorname{col}(Y_t).
\end{align*}
These identities prove the effective-rank and column-space claims.

The orthogonal-projector formula gives
\begin{equation*}
\operatorname{ovlp}_L^0(t,t')
=\frac1r\tr\bigl((Y_t^\top Y_t)^{-1}Y_t^\top Y_{t'}
(Y_{t'}^\top Y_{t'})^{-1}Y_{t'}^\top Y_t\bigr)
=\frac1r\tr(P_{Y_t}P_{Y_{t'}}).
\end{equation*}
\end{proof}

The overlap formula is invariant under independent positive rescaling of the two responses.
It requires invertibility only of the two diagonal Gram blocks.
For positive definite $T$ and symmetric $D$, define the weighted covariance with the log-eigenvalues of $T$
\begin{equation*}
\mathcal C_T(D):=\frac{\tr\{T(\log T-{m_T}  I)D\}}{\tr T},\qquad
{m_T} :=\frac{\tr(T\log T)}{\tr T}.
\end{equation*}

\subsubsection{Coordinates and entropy derivatives.}\label{er:coordinates}

Recall the rescaled response $Y_t=L_t^0S^{-1/2}$ and the adapter Gram matrix $T$.

For $s_t=t\eta>0$, write $\Gamma_t:=Y_t^\top Y_t/s_t$.
Define
\begin{equation*}
\begin{gathered}
\operatorname{er}_T(\Gamma ):=\bar\tau\exp(-\bar\ell/\bar\tau),\qquad K:=T^{1/2}\Gamma T^{1/2},\\
\bar\tau:=\frac1r\tr K,\qquad \bar\ell:=\frac1r\tr(K\log K).
\end{gathered}
\end{equation*}
By Lemma~\ref{tr:first-order-observables}, $\operatorname{er}_T(\Gamma_t)=\ER(\Delta W_t^{0})/r$.
At zero choose $\operatorname{er}_T(0)=1/r$.
Everywhere
$1/r\le \operatorname{er}_T\le1$.

The next lemma controls the effect of Gram perturbations on normalized effective rank.
\begin{lemma}[Entropy derivatives about a positive definite reference]\label{er:calculus}
Suppose $T$ and a reference $\Gamma_*$ have eigenvalues in fixed compact subintervals of $(0,\infty)$.
There is a constant $C$, depending only on these
eigenvalue bounds and a fixed positive spectral interval, such that, for every $\Gamma$ with spectrum in that interval and symmetric directions $D,{D'} $,
\begin{equation*}
|D\operatorname{er}_T(\Gamma )[{D} ]|\le C\|{D} \|_*/r,\qquad
|D^2\operatorname{er}_T(\Gamma )[D,{D'} ]|\le C\|{D} \|_F\|{D'} \|_F/r,
\end{equation*}
and the second bound also holds with $\|{D} \|\|{D'} \|_*$ in place of the two Frobenius norms.
For every $\Gamma '\succeq0$,
\begin{equation*}
|\operatorname{er}_T(\Gamma ')-\operatorname{er}_T(\Gamma_*)-D\operatorname{er}_T(\Gamma_*)[\Gamma '-\Gamma_*]|\le C\|\Gamma '-\Gamma_*\|_F^2/r.
\end{equation*}
In addition $D\operatorname{er}_T(\Gamma )[\Gamma ]=0$ for every positive definite $\Gamma$, and, for every $\lambda>0$ and symmetric ${D} $,
\begin{equation*}
D\operatorname{er}_T(\lambda I)[{D} ]=-\frac{\operatorname{er}_T(I)}{\lambda}\mathcal C_T(D).
\end{equation*}
\end{lemma}
\begin{proof}
\textit{Local derivatives.}
The normalized spectral traces $\bar\tau,\bar\ell$ are distinct from rescaled time $s_t$.
For $h(x)=x$ or $x\log x$, choose a fixed contour ${\mathcal C} $ around the positive spectral interval.
Resolvent differentiation gives
\begin{align*}
Dh(K)[{D} ]&=\frac1{2\pi i}\int_{\mathcal C}  h(z)(z-K)^{-1}D(z-K)^{-1}\,dz,\\
D\tr h(K)[{D} ]&=\tr(h'(K)D),\\
|D^2\tr h(K)[D,{D'} ]|&\le C\min\{\|{D} \|_F\|{D'} \|_F,\|{D} \|\|{D'} \|_*\}.
\end{align*}
The last inequality uses bounded resolvents and H\"older's inequality for Schatten norms.
In the definition of $\operatorname{er}_T$, $\bar\tau$ stays bounded below.
The chain rule for $\bar\tau\exp(-\bar\ell/\bar\tau)$ and normalized traces proves the derivative bounds.

\textit{Global quadratic remainder.}
For $h(x):=x\log x$, with $h(0):=0$, Taylor's theorem near a fixed positive $y$, continuity at zero, and $h(x)=o(x^2)$ at infinity give $|h(x)-h(y)-h'(y)(x-y)|\le C|x-y|^2$ for every $x\ge0$. Diagonalizing $K'$ and $K_*$ and using the weights $|u_i^\top v_j|^2$ therefore yields
\begin{equation*}
|\tr\{h(K')-h(K_*)-h'(K_*)(K'-K_*)\}|
\stackrel{(i)}{\le}C\sum_{i,j}|u_i^\top v_j|^2(x_i-y_j)^2
\stackrel{(ii)}{=}C\|K'-K_*\|_F^2,
\end{equation*}
where (i) uses the scalar remainder and (ii) uses orthonormality. If $\|K'-K_*\|_F/\sqrt r$ is small, the normalized trace stays bounded below, and the two-variable Taylor formula for $(\bar\tau,\bar\ell)\mapsto\bar\tau e^{-\bar\ell/\bar\tau}$ gives the claimed remainder. Otherwise $1/r\le\operatorname{er}_T\le1$ and the reference derivative bound give $1+C\|K'-K_*\|_F/\sqrt r\le C\|K'-K_*\|_F^2/r$. Finally $\|K'-K_*\|_F\le\|T\|\|\Gamma'-\Gamma_*\|_F$.

\textit{Scale invariance and scalar reference.}
Positive scaling gives $\operatorname{er}_T(c\Gamma )=\operatorname{er}_T(\Gamma )$ for $c>0$, hence $D\operatorname{er}_T(\Gamma )[\Gamma ]=0$.
At $\Gamma =\lambda I$, the chain rule yields
\begin{equation*}
D\operatorname{er}_T(\lambda I)[{D} ]=-\frac{\operatorname{er}_T(I)}{\lambda}
\frac{\tr\{T(\log T-m_T I){D} \}}{\tr T},
\end{equation*}
which is the scalar derivative formula in the statement.
\end{proof}

\subsubsection{Coupling to simultaneous-factor SGD.}
The next estimate is uniform over histories on the static event $\mathcal E_\star$ of Lemma~\ref{er:training-set}. On that event, choose a fixed $0<c_0\le1$ so that $\sigma_{\min}(A_0^\top)\ge2c_0$, and use
\begin{equation*}
\|W_0\|+\|A_0\|\le C,\quad \|Z\|_F\le C\sqrt r,\quad
\max_{i,\alpha}\|\Phi(x_{i,\alpha})\|_F\le Cq,\quad
\max_i\|\Pi y_i\|\le C\sqrt{r\Lambda_r},\quad
0\preceq\eta_H\widetilde{\mathcal H}_i\preceq(2-\zeta)I.
\end{equation*}
Fix $N_*\le N_{\mathrm{risk}}$ and put $T_*:=\eta_HN_*$. All constants below can depend on the fixed norm and spectral bounds and on $\zeta$, but not on the history. Define the factor error by
\begin{equation}\label{eq:factor-error}
\mathsf{err}_{\mathrm{fac}}:=C\delta T_*q^2r(1+T_*\Lambda_r)(1+T_*r\Lambda_r).
\end{equation}
The horizon dependence is retained because the geometry and risk transfers use different time scales.

\begin{lemma}[Uniform approximation of the SGD factors]\label{tr:coupling}
On $\mathcal E_\star$, for either selected rate $0<\eta\le\eta_H$, the first-order response satisfies $\max_{t\le N_*}\|L_t^0\|_F\le C\sqrt{T_*r\Lambda_r}$. If $0<\delta\le1$ and the right-hand side of \eqref{eq:factor-error}, with $C$ chosen sufficiently large, is at most $c_0/2$, then for every history
\begin{equation*}
\max_{t\le N_*}\{\|B_t/\delta-L_t^0\|_F,\|A_t-A_0\|_F\}\le\mathsf{err}_{\mathrm{fac}},
\qquad \sigma_{\min}(A_t^\top)\ge c_0.
\end{equation*}
\end{lemma}
\begin{proof}
\textit{Response bound and stopping time.}
The identities after \eqref{eq:feature-maps} and completion of the square give
\begin{equation*}
\|L_{t+1}^0\|_F^2-\|L_t^0\|_F^2
\stackrel{(i)}{\le}\frac\eta\zeta\|\Pi y_{I_t}/2\|^2
\stackrel{(ii)}{\le}C\eta r\Lambda_r,
\end{equation*}
where (i) uses $0\preceq\eta\widetilde{\mathcal H}_{I_t}\preceq(2-\zeta)I$ and (ii) uses the displayed target bound. Summation proves the response estimate. Stop at the first exit from $\|B_t/\delta\|_F\le1+C\sqrt{T_*r\Lambda_r}$ or $\|A_t-A_0\|_F\le c_0$.

\textit{Gradient perturbation.}
For $B=\delta L$, the exact residual in the proof of Lemma~\ref{tr:linearization} and
\begin{equation*}
\nabla_B f=2\Phi(x)(W_0+\delta LA)A^\top,\qquad
\nabla_A f=2\delta L^\top\Phi(x)(W_0+\delta LA)
\end{equation*}
imply, before exit,
\begin{align*}
\|\delta^{-1}\nabla_B\ell_i-(\widetilde{\mathcal H}_i[L]-\widetilde G_i)\|_F
&\stackrel{(i)}{\le} Cq^2r(1+T_*\Lambda_r)(1+T_*r\Lambda_r)(\delta+\|A-A_0\|_F),\\
\|\nabla_A\ell_i\|_F
&\stackrel{(i)}{\le} C\delta^2q^2r(1+T_*\Lambda_r)(1+T_*r\Lambda_r).
\end{align*}
Here (i) expands the residual into its linear and quadratic parts and uses $\frac12\sum_{\alpha,\beta}|\Pi_{\alpha\beta}|\le1$. Their norms are at most $C(\sqrt r+\|L\|_F)$ and $C(r+\|L\|_F^2)$. Every product in the two gradients is bounded by $Cq^2(r+\|L\|_F^2)(1+\|L\|_F^2)$, giving the displayed order.

\textit{Summation and closure.}
Since $\|I-\eta\widetilde{\mathcal H}_i\|\le1$, summing the right-factor increments and then the left-factor error increments, up to and including the first exit, yields
\begin{align*}
\max_t\|A_t-A_0\|_F&\le C\delta^2T_*q^2r(1+T_*\Lambda_r)(1+T_*r\Lambda_r),\\
\max_t\|B_t/\delta-L_t^0\|_F
&\le C\delta T_*q^2r(1+T_*\Lambda_r)(1+T_*r\Lambda_r)
\{1+C\delta T_*q^2r(1+T_*\Lambda_r)(1+T_*r\Lambda_r)\}.
\end{align*}
Choose the constant in \eqref{eq:factor-error} larger than the constants in these estimates. Its assumed smallness bounds both errors by $\mathsf{err}_{\mathrm{fac}}\le c_0/2$, strictly inside the stopping region. An exit is therefore impossible. Weyl's singular-value bound gives the last conclusion.
\end{proof}

\subsubsection{Observable errors.}
Set $s_0:=t_0\eta_L$. With sufficiently large fixed constants, define the remaining transfer errors by
\begin{equation}\label{eq:observable-errors}
\begin{aligned}
\mathsf{err}_{\mathrm{Gram}}&:=C(1+T_*r\Lambda_r)\mathsf{err}_{\mathrm{fac}}/s_0,\\
\mathsf{err}_{\mathrm{joint}}&:=C(1+\sqrt{T_*r\Lambda_r})\mathsf{err}_{\mathrm{fac}}/s_0,\\
\mathsf{err}_{\mathrm{risk}}&:=C\sqrt{r\Lambda_r}\{(1+\sqrt{T_*r\Lambda_r})\mathsf{err}_{\mathrm{fac}}+\delta r\Lambda_r\}\\
&\quad+C\{(1+\sqrt{T_*r\Lambda_r})\mathsf{err}_{\mathrm{fac}}+\delta r\Lambda_r\}^2.
\end{aligned}
\end{equation}
Only the first two bounds require a positive lower bound on the normalized response Gram matrix.
Because $S$ and $A_0A_0^\top$ have fixed positive spectral bounds, the conditions below on $\Gamma_t$ are equivalent to the earlier conditions on the response and adapter Gram matrices after adjusting the fixed interval endpoints.

\begin{lemma}[Effective-rank, left singular-subspace overlap, and risk approximation]\label{tr:observable}
Under Lemma~\ref{tr:coupling}, let $\eta\ge\eta_L$. For a history and $t_0\le t,t'\le N_*$ with $a_*I\preceq\Gamma_t,\Gamma_{t'}\preceq b_*I$, where $0<a_*<b_*$ are fixed, assume $\mathsf{err}_{\mathrm{Gram}},\mathsf{err}_{\mathrm{joint}}\le a_*/4$. Then
\begin{equation*}
\left|r^{-1}\ER(B_tA_t)-r^{-1}\ER(L_t^0A_0)\right|\le C_{\mathrm{ER}}\mathsf{err}_{\mathrm{Gram}},\qquad
\left|\operatorname{ovlp}_L(t,t')-\operatorname{ovlp}_L^0(t,t')\right|\le C_{\mathrm{rot}}\mathsf{err}_{\mathrm{joint}}.
\end{equation*}
The constants depend only on the fixed spectral bounds. Without either Gram condition, for every history and $0\le t\le N_*$,
\begin{equation*}
\left|\delta^{-2}\mathcal R_\delta(W_t)-\tfrac12\|\mathcal P(L_t^0)-E\|_F^2\right|\le\mathsf{err}_{\mathrm{risk}}.
\end{equation*}
\end{lemma}
\begin{proof}
\textit{Effective rank.}
The square-root map is Lipschitz on a fixed positive spectral interval: the Sylvester identity and its integral solution give
\begin{equation*}
\|(A_tA_t^\top)^{1/2}-(A_0A_0^\top)^{1/2}\|_F
\stackrel{(i)}{\le}C\|A_tA_t^\top-A_0A_0^\top\|_F
\stackrel{(ii)}{\le}C\mathsf{err}_{\mathrm{fac}},
\end{equation*}
where (i) uses the common lower singular-value bound and (ii) uses Lemma~\ref{tr:coupling}. Expanding the two adapter Gram matrices consequently bounds their Frobenius difference, after division by $s_t=t\eta\ge s_0$, by $C(1+T_*r\Lambda_r)\mathsf{err}_{\mathrm{fac}}/s_0$. The first-order adapter Gram is a fixed well-conditioned congruence of $\Gamma_t$. Enlarge the constant in \eqref{eq:observable-errors} to account for this congruence. Its assumed smallness then preserves the positive spectral interval along the comparison segment. Lemma~\ref{er:calculus} and $\|D\|_*/r\le\|D\|_F/\sqrt r$ prove the effective-rank bound.

\textit{Subspace overlap.}
Apply $(D-D')^\top D+D'^\top(D-D')=D^\top D-D'^\top D'$ to the two-column-block matrices $[L_t^\delta/\sqrt{s_t},L_{t'}^\delta/\sqrt{s_{t'}}]$ and their first-order versions. Their joint Gram difference is at most $C(1+\sqrt{T_*r\Lambda_r})\mathsf{err}_{\mathrm{fac}}/s_0$. The diagonal blocks stay positive definite. The derivative identity $D(M^{-1})[H]=-M^{-1}HM^{-1}$, applied to the projector formula in Lemma~\ref{tr:first-order-observables}, bounds the overlap change by $C_{\mathrm{rot}}\mathsf{err}_{\mathrm{joint}}$. Full row rank of $A_t$ identifies the left column spaces of $B_tA_t$ and $B_t$.

\textit{Excess risk.}
Expand $W_t=W_0+\delta L_t^\delta A_t$ and the teacher Gram matrix. Lemma~\ref{tr:coupling} gives
\begin{align*}
\|\delta^{-1}(W_tW_t^\top-F_\delta F_\delta^\top)-\{\mathcal P(L_t^0)-E\}\|_F
&\le C\{(1+\sqrt{T_*r\Lambda_r})\mathsf{err}_{\mathrm{fac}}+\delta r\Lambda_r\},\\
\|\mathcal P(L_t^0)-E\|_F&\le C\sqrt{r\Lambda_r}.
\end{align*}
The identity $|\|U+D\|_F^2-\|U\|_F^2|\le2\|U\|_F\|D\|_F+\|D\|_F^2$ and Fact~\ref{fact:gaussian-isometry} prove the risk bound.
\end{proof}

\subsubsection{An explicit task radius.}
We verify the same deterministic task radius as in Appendix B. Superscripts $\mathrm G,\mathrm R$ evaluate the factor error at $N_*=N$ and $N_*=N_{\mathrm{risk}}$, respectively. Gram and overlap errors use the former, and risk uses the latter.
\begin{proof}[Proof of Proposition~\ref{prop:task-radius}]
Lemma~\ref{er:training-set} supplies the static norm and step-size event. The windows satisfy $T_*^{\mathrm G}=O(r^{-1})$, $T_*^{\mathrm R}=O(1)$ and $s_0^{-1}=O(\sqrt q/\Lambda_r^J)$. Substitution in \eqref{eq:factor-error}--\eqref{eq:observable-errors} gives
\begin{align*}
\mathsf{err}_{\mathrm{fac}}^{\mathrm G}&=O(\delta q^2\Lambda_r),&
\mathsf{err}_{\mathrm{fac}}^{\mathrm R}&=O(\delta q^2r^2\Lambda_r^2)=o(1),\\
\mathsf{err}_{\mathrm{Gram}}&=O(\delta q^{5/2}\Lambda_r^{2-J}),&
\mathsf{err}_{\mathrm{joint}}&=O(\delta q^{5/2}\Lambda_r^{3/2-J}),\\
\mathsf{err}_{\mathrm{risk}}&=O(\delta q^2r^3\Lambda_r^3).
\end{align*}

The two terms defining $\delta_*$ therefore imply
\begin{equation*}
\mathsf{err}_{\mathrm{Gram}}+\mathsf{err}_{\mathrm{joint}}
\stackrel{(i)}{=}O((qr\Lambda_r)^{-1})=o(\eta_L),\qquad
\mathsf{err}_{\mathrm{risk}}\stackrel{(ii)}{=}O((q\Lambda_r)^{-1})=o(q^{-1}),
\end{equation*}
where (i) uses $\delta\le\Lambda_r^{J-3}/(q^{7/2}r)$ and (ii) uses $\delta\le(q^3r^3\Lambda_r^4)^{-1}$. The fixed margins in \eqref{eq:task-radius} then dominate all four errors for sufficiently large $r$.
\end{proof}

\subsection{Rescaled coordinates and forcing moments}\label{app:forcing-moments}\label{app:moments}
We put the first-order dynamics into regression form and compute their forcing moments.
Throughout, the rule is fixed and suppressed, and unqualified expectations average fresh pairs conditional on initialization and task.
The constants have the dependence specified in Section~\ref{app:foundation}.

\subsubsection{Rescaled response coordinates and initialization.}
For the selected rule $m$, use the features and quadratic labels in \eqref{eq:feature-label} and the projection $\Pi=\Pi_m$ from Section~\ref{app:reduction}:
\begin{equation}\label{er:operators}
\begin{aligned}
Y_{t+1}&=Y_t+\eta\bigl(G_{I_t}-\mathcal H_{I_t}[Y_t]\bigr),\qquad Y_0=0,\\
G_i&:=\frac12\sum_{\alpha,\beta\in\{+,-\}}\Pi_{\alpha\beta}y_{i,\beta}X_{i,\alpha},\\
\mathcal H_i[Y]&:=\sum_{\alpha,\beta\in\{+,-\}}\Pi_{\alpha\beta}X_{i,\alpha}\langle X_{i,\beta},YS\rangle_F.
\end{aligned}
\end{equation}
Here $G_i\in\R^{d_o\times r}$ is the additive forcing and $\mathcal H_i$ is a linear operator on response matrices.
The coordinate relations are
\begin{equation*}
\widetilde G_i=G_iS^{1/2},\qquad
\mathcal H_i[Y]=\widetilde{\mathcal H}_i[YS^{1/2}]S^{-1/2},\qquad
\Delta W_t^0=\delta Y_tS^{1/2}A_0.
\end{equation*}
Here $I_t$ is uniform over the stored pairs. For one fresh pair omit $i$. Write $G_{m,i}$ when the rule is shown and $\mathcal H_iY:=\mathcal H_i[Y]$.

For row blocks, define
\begin{equation*}
P_{\mathrm{sig}}:=U_0U_0^\top+U_{\perp}U_{\perp}^\top,\qquad P_{\mathrm{nuis}}:=I-P_{\mathrm{sig}},
\qquad X_{\alpha,\mathrm{sig}}:=P_{\mathrm{sig}}X_\alpha,\quad X_{\alpha,\mathrm{nuis}}:=P_{\mathrm{nuis}}X_\alpha.
\end{equation*}
These row blocks have dimensions $k+p$ and $q$.
The same convention applies to forcing and response matrices. For Gram matrices and Hessians it denotes the
contributions of the projected features.
Put
\begin{equation*}
a_{i,\alpha}:=Q^\top x_{i,\alpha},\qquad
R_a:=\begin{pmatrix}1&\rho_a\\\rho_a&1\end{pmatrix},\qquad
R_{\times}:=\rho_aU_0U_0^\top+\rho_oU_{\perp}U_{\perp}^\top.
\end{equation*}
Let $z_{i,\alpha}\in\R^q$ be the nuisance coordinates in a fixed orthonormal basis.
Then $\E[a_{i,\alpha}a_{i,\beta}^\top]=(R_a)_{\alpha\beta}I_r$, and $R_{\times}=\Cov(x_{i,+},x_{i,-})$.
The two coordinate norms are equivalent on the initialization event below.

The next lemma supplies the uniformly conditioned coordinates used in all subsequent estimates.
\begin{lemma}[Initialization and teacher bounds]\label{er:init}
Under the dimension and bounded-spectrum conditions of Assumptions~\ref{ass:dimensions} and \ref{ass:spectrum}, for every fixed $M>0$ and all sufficiently large
$r$, with probability at least $1-C_M(qr)^{-M}$ over initialization and teacher, the spectra of $S$, $A_0 A_0^{\top}$, and $T$ lie in a fixed interval
$[a_*,b_*]\subset(0,\infty)$, $\|E\|\le C$, and $c_*r\le \|E\|_F^2\le Cr$.
The constants $a_*,b_*,c_*,C>0$ depend only on the fixed model bounds.
The same event
can be chosen to satisfy
\begin{equation*}
\left|\frac{\tr S}{r\mu_{1,r}}-\frac{k}{r}\right|\le r^{-1/2}.
\end{equation*}
Moreover $\tr S/r\to {\gamma_k} \mu_1$ in every fixed moment, where ${\gamma_k} =\lim k/r$ and $\mu_1=\lim\mu_{1,r}$.
\end{lemma}
\begin{proof}
Fact~\ref{fact:gaussian-extremes}, applied to standard Gaussian matrices with $k$ and $d_i$ rows and $r$ columns,
gives singular values between $\sqrt j-\sqrt r-\varepsilon\sqrt r$ and $\sqrt j+\sqrt r+\varepsilon\sqrt r$
for $j=k,d_i$, except with probability $Ce^{-cr}$.
Together with the bounded spectrum of $\Sigma_0$ and the definitions of $S,T$, this gives
\begin{equation*}
\lambda_{\min}(\Sigma_0)^2A_0V_0V_0^\top A_0^\top
\preceq S\preceq\lambda_{\max}(\Sigma_0)^2A_0V_0V_0^\top A_0^\top,
\qquad a_*I\preceq S,A_0A_0^\top,T\preceq b_*I.
\end{equation*}
Sub-Gaussian matrix norm bounds give $\|Z_a\|+\|Z_o\|\le C$ with the same exponential confidence.
Since $\E\|Z_o\Sigma_0\|_F^2=\sigma_o^2\tr\Sigma_0^2\asymp r$, Hanson--Wright gives
\begin{equation*}
c_*r\stackrel{(i)}{\le}2\|Z_o\Sigma_0\|_F^2
\le\|E\|_F^2\le2(k+p)\|E\|^2
\stackrel{(ii)}{\le}Cr.
\end{equation*}
Here (i) uses Fact~\ref{fact:hanson-wright}, while (ii) uses
$\|E\|\le2\|\Sigma_0\|(\|Z_a\|+\|Z_o\|)\le C$.

For independent standard Gaussian entries $\mathsf G_{ij}$,
\begin{equation*}
\frac{\tr S}{r}=\frac1{r^2}\sum_{i=1}^k\sum_{j=1}^r\lambda_i(M_a)\mathsf G_{ij}^2,
\qquad
\Prob\left(\left|\frac{\tr S}{r}-\frac{k}{r}\mu_{1,r}\right|>x\right)
\le2e^{-cr^2\min\{x^2,x\}}.
\end{equation*}
Taking $x=\mu_{1,r}r^{-1/2}$ gives the stated trace event with exponential confidence.
Gaussian quadratic-form moments also give, for every fixed $p$,
\begin{equation*}
\left\|\frac{\tr S}{r}-\gamma_k\mu_1\right\|_{L^p}
\le\frac{C_p}{r}+\left|\frac{k}{r}\mu_{1,r}-\gamma_k\mu_1\right|\longrightarrow0.
\end{equation*}
Denote the intersection of these initialization, singular-value and teacher events by $\mathcal E_{\mathrm{init}}$.
Its failure probability is $Ce^{-cr}=O_M((qr)^{-M})$ for every fixed $M$.
\end{proof}
\subsubsection{Forcing moments}

Put $Q_a:=U_0^\top Q$.
The teacher perturbation and polar frame from Section~\ref{app:reduction} have the following block forms.
In the pretrained, new, and nuisance coordinates,
\begin{equation*}
E=\begin{pmatrix}E_{aa}&E_{oa}^\top&0\\E_{oa}&0&0\\0&0&0\end{pmatrix},\qquad
E_{aa}:=\Sigma_0Z_a^\top+Z_a\Sigma_0,\quad E_{oa}:=Z_o\Sigma_0,\qquad Q=\begin{pmatrix}Q_a\\0\\0\end{pmatrix}.
\end{equation*}
Define the label energies and their compressed matrices:
\begin{equation*}
\nu _a:=2\|E_{aa}\|_F^2,\quad \nu _o:=4\|E_{oa}\|_F^2,\qquad
\Upsilon_a:=Q_a^\top E_{aa}^2Q_a,\quad \Upsilon_o:=Q_a^\top E_{oa}^\top E_{oa}Q_a.
\end{equation*}
The retained fractions are defined in Section~\ref{main:spectrum}.
For rule ${m} \in\{R,C\}$, let $G_{m} $ be the rescaled forcing term in \eqref{er:operators}, $\widetilde G_m=G_mS^{1/2}$ its Euclidean version, and $\Omega
_{m} $ the covariance of any one nuisance row of $G_{m} $.
Define
\begin{equation*}
\nu_R:=(\nu _a+\nu _o)/2,\qquad \nu_C:=(\vartheta_a\nu _a+\vartheta_o\nu _o)/2.
\end{equation*}

Here $\E G_m$ is the coherent fitting drift, while $\Omega_m$ describes initialized stochastic forcing in each of the $q$ nuisance rows. These rows have zero mean, so their covariance equals their second moment.
Relative to the raw rule, centering multiplies the pretrained and orthogonal signal components of the mean by $\vartheta_a<1/2<\vartheta_o$, respectively.
\begin{lemma}[Initialized drift and prediction variance]\label{rr:source}
Conditional on the teacher and initialization, expectation over a fresh Gaussian pair gives
\begin{align*}
\E G_R&=2\begin{pmatrix}E_{aa}Q_a\\E_{oa}Q_a\\0\end{pmatrix},\\
\E G_C&=2\begin{pmatrix}\vartheta_aE_{aa}Q_a\\\vartheta_oE_{oa}Q_a\\0\end{pmatrix},\\
\langle2E W_0A_0^{\top},\E\widetilde G_R\rangle&=4(\tr(S\Upsilon_a)+\tr(S\Upsilon_o)),\\
\langle2E W_0A_0^{\top},\E\widetilde G_C\rangle&=4(\vartheta_a\tr(S\Upsilon_a)+\vartheta_o\tr(S\Upsilon_o)).
\end{align*}
The nuisance-row covariances satisfy
\begin{equation}\label{rr:source-covariances}
\begin{aligned}
\Omega _R&=\frac{\nu _a+\nu _o}{2}I+4(\Upsilon_a+\Upsilon_o),\\
\Omega _C&=\frac{\vartheta_a\nu _a+\vartheta_o\nu _o}{2}I+2\vartheta_a\Upsilon_a+(1+\rho_a^2-2\rho_a\rho_o)\Upsilon_o.
\end{aligned}
\end{equation}
Consequently, on the bounded initialization and teacher event of Lemma~\ref{er:init},
\begin{equation*}
\frac12\E\|\mathcal P(\widetilde G_m)\|_F^2=q\tr(S^2)\nu_{m} +O(qr+r^3),\qquad {m} \in\{R,C\}.
\end{equation*}
The constants are uniform over the fixed model-parameter bounds.
\end{lemma}
\begin{proof}
\textit{Mean forcing.}
Gaussian integration by parts and regression across the pair give
\begin{equation*}
\E[y_+X_+]=2EQ,\qquad \E[y_-X_+]=2R_\times ER_\times Q,
\end{equation*}
so \eqref{er:operators} gives the two stated means.
The inner-product identities follow from $W_0A_0^\top=QS^{1/2}$ and
$Q^\top E^2Q=\Upsilon_a+\Upsilon_o$.
For symmetric $D,D'$, two integrations by parts give the fourth-moment identity
\begin{equation*}
\E[(x^\top Dx-\tr D)(x^\top D'x-\tr D')xx^\top]
=2\tr(DD')I+4(DD'+D'D).
\end{equation*}

\textit{Nuisance-row covariance.}
For one nuisance row, its coordinates $z_+,z_-$ are independent standard Gaussians, independent of the signal coordinates. Thus
\begin{equation*}
\begin{aligned}
G_{R,b}&=\tfrac12(y_+z_+a_+^\top+y_-z_-a_-^\top),&
\Omega_R&=\tfrac12\E[y_+^2a_+a_+^\top],\\
G_{C,b}&=\tfrac14(y_+-y_-)(z_+a_+^\top-z_-a_-^\top),&
\Omega_C&=\tfrac18\E[(y_+-y_-)^2a_+a_+^\top].
\end{aligned}
\end{equation*}
The fourth-moment identity immediately gives $\Omega_R=\tr(E^2)I+4Q^\top E^2Q$.
For $\Omega_C$, Gaussian regression retains the conditional variance as well as the conditional mean:
\begin{equation*}
\E[y_-\mid x_+]=x_+^\top R_\times ER_\times x_+-\tr(R_\times ER_\times),
\qquad
\E[y_-^2x_+x_+^\top]=2\tr(E^2)I+8R_\times E^2R_\times.
\end{equation*}
Substitution and the bilinear fourth-moment identity yield
\begin{equation*}
\begin{aligned}
\Omega_C
&\stackrel{(i)}{=}\frac{\tr(E^2)-\tr(ER_\times ER_\times)}2I
+Q^\top(E^2+R_\times E^2R_\times-ER_\times ER_\times-R_\times ER_\times E)Q\\
&\stackrel{(ii)}{=}\frac{\vartheta_a\nu_a+\vartheta_o\nu_o}{2}I
+2\vartheta_a\Upsilon_a+(1+\rho_a^2-2\rho_a\rho_o)\Upsilon_o.
\end{aligned}
\end{equation*}
Here (i) expands $(y_+-y_-)^2$, while (ii) uses the displayed blocks of $E$, $R_\times Q=\rho_aQ$, and
$2\tr(ER_\times ER_\times)=\rho_a^2\nu_a+\rho_a\rho_o\nu_o$.

\textit{Prediction variance.}
The nuisance block contributes
$\E\|\mathcal P(\widetilde G_m)_{\mathrm{nuis},a}\|_F^2
=\E\|G_{m,\mathrm{nuis}}SQ_a^\top\|_F^2=q\tr(S^2\Omega_m)$.
On $\mathcal E_{\mathrm{init}}$, $\|\Omega_m-\nu_mI\|\le C$, $\tr S^2\le Cr$ and $\|\mathcal P\|\le C$.
Gaussian moments give $\E|y_\alpha|^4\le Cr^2$ and $\E\|X_{\alpha,\mathrm{sig}}\|_F^4\le Cr^4$, hence
\begin{equation*}
\begin{aligned}
\E\|\widetilde G_{m,\mathrm{sig}}\|_F^2
&\stackrel{(i)}{\le}C(\E|y_\alpha|^4)^{1/2}(\E\|X_{\alpha,\mathrm{sig}}\|_F^4)^{1/2}\le Cr^3,\\
\tfrac12\E\|\mathcal P(\widetilde G_m)\|_F^2
&\stackrel{(ii)}{=}q\tr(S^2\Omega_m)+O(r^3)
=q\tr(S^2)\nu_m+O(qr+r^3).
\end{aligned}
\end{equation*}
Here (i) is Cauchy--Schwarz and (ii) includes both symmetric nuisance blocks.
\end{proof}

\subsection{Adapter spectrum and coefficient margins}\label{app:spectrum}
The matrices $S$ and $T$ are correlated, so the entropy response requires their joint spectral limit.
We first obtain that limit and its population response formulas, then average over the teacher and derive strict coefficient margins.
In the free-probability calculations, $t,s,z$ are resolvent or subordination variables, $\tau$ is normalized trace, and $\delta_x$ denotes a Dirac mass.
The auxiliary traces $m,\ell$ and the scalar transforms below are local to these spectral calculations.
Constants follow Section~\ref{app:foundation}.

\subsubsection{Population notation.}
For the positive compact law $H$ in Assumption~\ref{ass:spectrum}, let $\lambda\sim H$, put $\varrho:=\lambda/\mu_1$, and define the probability measure $\pi(dx):=x\,\operatorname{Law}(\varrho )(dx)$.
Its total mass is $\E
\varrho=1$.
Expectations and covariances with subscript $\pi$ use this measure.
Define
\begin{equation*}
\begin{gathered}
\chi_H(w):=\E_\pi(1+{w} \varrho )^{-1},\qquad
\varphi _H(w):=\frac {\gamma_k} {\gamma_d} \,{w} \chi_H(w),\qquad
j_H(w):=\Cov_\pi(\varrho ,\log(1+{w} \varrho )),\\
{f_{\mathrm{MP}}} (z):=\frac{(1-\gamma_d z)({\gamma_d} +1-2\gamma_d z)^2}{{\gamma_d} +1-\gamma_d z},\qquad
{h_{\mathrm{MP}}} (z):=\frac{{\gamma_d} ({\gamma_d} +2-2\gamma_d z)}{{\gamma_d} +1-\gamma_d z},\\
{w_*}  := \inf\{{w} >0:\varphi _H(w)\ge1/{\gamma_d} \}.
\end{gathered}
\end{equation*}
Endpoint values are interpreted by continuity.

Let $M_Q:=Q_a^\top M_aQ_a$ be the compressed population covariance.
The population spectral moments and $v_H$ are defined in Appendix B.
Define its normalized leverage limit by
\begin{equation*}
h_M:=\gamma_k\{1-\chi_H(w_*)\}.
\end{equation*}
\subsubsection{Spectrum of the first-order adapter.}
For the population covariance $M_a$ from Appendix B, let ${\mathsf G} \in\R^{d_i\times r}$ be standard Gaussian, and let
${\mathsf G} _a$ contain its
first $k$ rows.
Orthogonal invariance and the polar decomposition give
\begin{equation*}
\begin{gathered}
S={\mathsf G}_a^\top M_a{\mathsf G}_a/r,\quad Q_a=M_a^{1/2}{\mathsf G} _aS^{-1/2}/\sqrt r,\\
T=S^{1/2}({\mathsf G} ^\top {\mathsf G} /{d_i})S^{1/2}.
\end{gathered}
\end{equation*}
Here $Q_a^\top Q_a=I_r$.
For the shared spectral calculations define
\begin{equation*}
\begin{gathered}
M:=\diag(M_a,0_{d_i-k}),\qquad
{Q_{\mathrm{ext}}} :=\binom{Q_a}{0_{(d_i-k)\times r}},\qquad
W_{\mathrm{cov}}:={\mathsf G}{\mathsf G}^\top/r,\\
S_1:=M^{1/2}W_{\mathrm{cov}}M^{1/2},\qquad S_2:=M^{1/2}W_{\mathrm{cov}}^2M^{1/2}.
\end{gathered}
\end{equation*}
The weighted log covariance is defined in Section~\ref{app:reduction}.

The joint spectral limit controls the two entropy covariances and normalized effective rank.
\begin{lemma}[Limits and concentration of adapter spectral statistics]\label{sp:physical-limit}
Suppose $k/r\to {\gamma_k} $, ${d_i}/r\to {\gamma_d} $, $1<{\gamma_k} <{\gamma_d} $, and the eigenvalues of $M_a$ lie in a fixed
$[\lambda_-,\lambda_+]\subset(0,\infty)$.
Along any subsequence with limiting empirical law $H$, there are constants $0<m_*<M_*<\infty$ and $c_*>0$ such that,
for all sufficiently large $r$,
\begin{equation*}
\Prob\{m_*I\preceq S,T\preceq M_*I\}\ge1-\exp(-c_*r).
\end{equation*}
There are deterministic limits ${\omega_1} ,{\omega_2} ,{\mathrm{er}_H} $ such that
\begin{equation*}
\left(\mathcal C_T({M_Q} ),\mathcal C_T(S),\frac{\ER(\sqrt T)}r\right)
\longrightarrow({\omega_1} ,{\omega_2} ,{\mathrm{er}_H} )
\end{equation*}
jointly in probability and in $L^1$.
Moreover,
\begin{equation*}
\begin{gathered}
\E\left[\mathcal C_T({M_Q} )\frac{\ER(\sqrt T)}r\right]\to {\omega_1} {\mathrm{er}_H} ,\\
\E\left[\mathcal C_T(S)\frac{\ER(\sqrt T)}r\right]\to {\omega_2} {\mathrm{er}_H} .
\end{gathered}
\end{equation*}
\end{lemma}
\begin{proof}

The Gaussian singular-value bounds included in $\mathcal E_{\mathrm{init}}$ (Lemma~\ref{er:init}) give the stated common positive spectral interval with failure $Ce^{-cr}$.
For continuous $f$ with $f(0)=0$, the polar decomposition gives
\begin{equation*}
\begin{gathered}
S_1=Q_{\mathrm{ext}}SQ_{\mathrm{ext}}^\top,\qquad
S_2=Q_{\mathrm{ext}}((d_i/r)T)Q_{\mathrm{ext}}^\top,\\
\tr\{f((d_i/r)T)M_Q\}=\tr\{f(S_2)M\},\qquad
\tr\{f((d_i/r)T)S\}=\tr\{f(S_2)S_1\}.
\end{gathered}
\end{equation*}
\begin{fact}[Joint covariance--Wishart limit, {\citep[Theorem 54]{MingoPopa2013}}]
For every mixed polynomial $P$, the bounded deterministic covariance $M$ and independent orthogonally invariant real Wishart matrix $W_{\mathrm{cov}}$ satisfy
\begin{equation*}
\frac1{d_i}\tr P(M,W_{\mathrm{cov}})\xrightarrow{\Prob}\tau P(M,W_{\mathrm{cov}}),\qquad
M\sim(1-\gamma_k/\gamma_d)\delta_0+(\gamma_k/\gamma_d)H,\quad W_{\mathrm{cov}}\sim\mathrm{MP}_{\gamma_d},
\end{equation*}
where the limit variables are freely independent and $\tau$ is normalized trace.
\end{fact}
Here $\mathrm{MP}_{\gamma_d}$ has mean one, zero mass $1-1/\gamma_d$, and positive support
$[(\sqrt{\gamma_d}-1)^2,(\sqrt{\gamma_d}+1)^2]$. The same Wishart variable enters $S_1,S_2$.
On the compact spectral interval, polynomial approximation applies to $M^{1/2}$ and to $f(x)=x,x\log x$, since
\begin{equation*}
\left|d_i^{-1}\tr\{[f(S_2)-p(S_2)]D\}\right|\le\|f-p\|_\infty\|D\|,
\qquad D=M,S_1.
\end{equation*}
Consequently, with $m:=\tau S_2>0$ and $\ell:=\tau(S_2\log S_2)/m$,
\begin{equation*}
\begin{aligned}
\mathcal C_T(M_Q)&\xrightarrow{\Prob}\omega_1:=\tau\{S_2(\log S_2-\ell)M\}/m,\\
\mathcal C_T(S)&\xrightarrow{\Prob}\omega_2:=\tau\{S_2(\log S_2-\ell)S_1\}/m,\\
\ER(\sqrt T)/r
&=\frac{\tr S_2}{r}\exp\{-\tr(S_2\log S_2)/\tr S_2\}
\xrightarrow{\Prob}\mathrm{er}_H:=\gamma_dm e^{-\ell}.
\end{aligned}
\end{equation*}

The entropy bound $\sum_i p_i|\log p_i-\sum_jp_j\log p_j|\le2\log r$ gives
$|\mathcal C_T(D)|\le2\|D\|\log r$, whereas $|\mathcal C_T(D)|\le C$ on $\mathcal E_{\mathrm{init}}$.
Both $M_Q$ and $S$ have bounded fixed norm moments, so the global moment form of Lemma~\ref{lem:tail-absorption}
makes the initialization-complement contribution $o(1)$.
This proves $L^1$ convergence. Since $0\le\ER(\sqrt T)/r\le1$, the two product limits follow from
$\E|XZ-xz|\le\E|X-x|+|x|\E|Z-z|$.
\end{proof}

\subsubsection{Population integral formulas and ordered entropy responses.}
For the free limiting variables in Lemma~\ref{sp:physical-limit}, let $\tau$ be their normalized trace and put $H_0:=(1-{\gamma_k} /{\gamma_d}
)\delta_0+({\gamma_k} /{\gamma_d} )H$, where $\delta_x$ is the probability measure concentrated at $x$.
For $t,s>0$ and $D=M,S_1$, define
\begin{equation*}
\begin{gathered}
\varphi (t):=\tau\{S_2(S_2+t)^{-1}\},\qquad
\mathcal U(s):=\int\frac{\lambda}{1+s\lambda}\,dH_0(\lambda),\\
R_D(t):=\frac{\tau\{S_2(S_2+t)^{-1}D\}}{\varphi (t)},\qquad
R_D(\infty):=\frac{\tau(S_2D)}{\tau S_2}.
\end{gathered}
\end{equation*}
The next result derives both responses in one subordination coordinate, making their strict ordering visible in the integrands.
\begin{proposition}[Population integral formulas and ordering of entropy responses]
\label{sp:positive-covariances}\label{sp:population-formulas}
Under the assumptions of Lemma~\ref{sp:physical-limit}, suppose the limiting pretrained law $H$ is nonconstant.
Then $w_*$ is the unique finite root of $\varphi_H(w)=1/\gamma_d$, and the following integrals are finite and positive.
In these integrals, $\varphi:=\varphi_H(w)$ and unmarked $\chi_H$ means $\chi_H(w)$. Primes denote ordinary derivatives.
\begin{equation}\label{sp:population-responses}
\begin{aligned}
\frac{{\omega_1} }{\mu_1}
&=\frac1{{\gamma_d} +1}\int_0^{{w_*} }
\{{f_{\mathrm{MP}}} (\varphi )-w{f_{\mathrm{MP}}} '(\varphi )\varphi '\}j_H'(w)\,dw,\\
\frac{\omega_2-\omega_1}{\mu_1}
&=\frac{{\gamma_k} }{{\gamma_d} ({\gamma_d} +1)}\int_0^{{w_*} }
\chi_H\left\{\frac{{f_{\mathrm{MP}}} (\varphi )}{w} -{f_{\mathrm{MP}}} '(\varphi )\varphi '\right\}
\{{h_{\mathrm{MP}}} (0)-\chi_H{h_{\mathrm{MP}}} (\varphi )\}\,dw.
\end{aligned}
\end{equation}
Consequently $0<\omega_1<\omega_2$, so $\omega=\omega_1/\omega_2\in(0,1)$.
The leverage limit and its bounds are those in Lemma~\ref{sp:leverage-bounds}.
\end{proposition}
\begin{proof}
\textit{Multiplicative subordination.}
\begin{fact}[Multiplicative subordination, {\citep[Lemma 3.4 and Corollary 3.6]{LehnerSzpojankowski2020}}]
For the freely independent positive variables $M,W_{\mathrm{cov}}^2$, at resolvent argument $-1/t$ the positive subordination parameters $s,z$ satisfy
\begin{equation*}
\varphi=s\mathcal U(s)=\int\frac{z\lambda^2}{1+z\lambda^2}\,d\mathrm{MP}_{\gamma_d}(\lambda),
\qquad sz=\frac{\varphi}{t(1-\varphi)}.
\end{equation*}
The identities extend to the zero atoms by analytic continuation for $t>0$.
\end{fact}
Here $0<\varphi(t)<1/\gamma_d$ and
$\varphi'(t)=-\tau\{S_2(S_2+t)^{-2}\}<0$.
Both integrals in the subordination identity increase in their positive parameters, so $s'(t),z'(t)<0$.
Conditioning onto the algebra generated by $M$ gives
\begin{equation*}
\tau\{(S_2+t)^{-1}M\}=\mathcal U(s)/t,\qquad
R_M(t)=\frac{\int\lambda^2/(1+s\lambda)\,dH_0(\lambda)}{\mathcal U(s)}.
\end{equation*}
Use $w:=\mu_1s$ and the energy-weighted law $\pi$.
Since $\E\varrho=\E_\pi\varrho^{-1}=1$ and $\E_\pi\varrho=v_H$,
\begin{equation*}
\mathcal U(s)=\frac{\gamma_k}{\gamma_d}\mu_1\chi_H(w),\qquad
\varphi(t)=\varphi_H(w),\qquad
R_M(t)=\mu_1\frac{\E_\pi[\varrho/(1+w\varrho)]}{\chi_H(w)}.
\end{equation*}
Moreover $\varphi_H'(w)=(\gamma_k/\gamma_d)\E_\pi(1+w\varrho)^{-2}>0$,
$\varphi_H(0)=0$, and $\varphi_H(\infty)=\gamma_k/\gamma_d>1/\gamma_d$.
Thus the root $w_*$ exists and is unique, and $t:\infty\to0$ corresponds to $w:0\to w_*$.
The endpoint at $t=0$ uses the positive support projection of $S_2$, whose trace is $1/\gamma_d$ by Lemma~\ref{sp:physical-limit}.
For independent $\varrho,\varrho'\sim\pi$, the covariance identity gives
\begin{equation*}
\begin{aligned}
j_H'(w)&=\tfrac12\E\frac{(\varrho-\varrho')^2}{(1+w\varrho)(1+w\varrho')}>0,\\
v_H\chi_H(w)-\E_\pi\frac{\varrho}{1+w\varrho}&=wj_H'(w),\\
R_M(\infty)-R_M(t)&=\frac{\mu_1w j_H'(w)}{\chi_H(w)}.
\end{aligned}
\end{equation*}
The last equality uses $R_M(\infty)=\mu_1v_H$. Nonconstancy is needed for the strict first inequality.

\textit{Wishart resolvent difference.}
Put $\upsilon(z):=\int\lambda/(1+z\lambda^2)\,d\mathrm{MP}_{\gamma_d}(\lambda)$.
The resolvent identity for $W_{\mathrm{cov}}M^{1/2}$ and its adjoint, followed by subordination onto the Wishart algebra, gives
\begin{equation*}
\begin{aligned}
\tau\{(S_2+t)^{-1}S_1\}
&=\tau\{W_{\mathrm{cov}}^\dagger(W_{\mathrm{cov}}MW_{\mathrm{cov}})
(W_{\mathrm{cov}}MW_{\mathrm{cov}}+t)^{-1}\}=z\upsilon(z),\\
R_{S_1}(t)-R_M(t)
&=\frac{\mathcal U(s)-tz\upsilon(z)}{\varphi}
=\frac1s\left(1-\frac{\upsilon(z)}{1-\varphi}\right).
\end{aligned}
\end{equation*}
The pseudoinverse is bounded because the positive Marchenko--Pastur support is bounded away from zero.
For $L(\zeta):=\int(1+\zeta\lambda)^{-1}\,d\mathrm{MP}_{\gamma_d}(\lambda)$,
the quadratic equation at $\zeta=i\sqrt z$ reads
\begin{equation*}
\begin{aligned}
(1-L)(1+\gamma_d\zeta L)&=\zeta L,&
L(i\sqrt z)&=1-\varphi-i\sqrt z\upsilon(z),\\
\varphi&=z\upsilon(z)(\gamma_d+1-2\gamma_d\varphi),&
\upsilon(z)(1+\gamma_dz\upsilon(z))&=(1-\varphi)(1-\gamma_d\varphi).
\end{aligned}
\end{equation*}
Eliminating $z\upsilon(z)$ and using $sz=\varphi/[t(1-\varphi)]$ yields
\begin{equation*}
R_{S_1}(t)-R_M(t)=\mathcal U(s)h_{\mathrm{MP}}(\varphi),\qquad
t=\frac{f_{\mathrm{MP}}(\varphi)}s=\frac{\mu_1f_{\mathrm{MP}}(\varphi)}w.
\end{equation*}
Consequently
\begin{equation*}
\begin{aligned}
&[R_{S_1}(\infty)-R_M(\infty)]-[R_{S_1}(t)-R_M(t)]\\
&\qquad=\frac{\gamma_k}{\gamma_d}\mu_1
\{h_{\mathrm{MP}}(0)-\chi_H(w)h_{\mathrm{MP}}(\varphi)\}>0.
\end{aligned}
\end{equation*}
Indeed $0<\chi_H(w)<1$, $h_{\mathrm{MP}}>0$, and
$h_{\mathrm{MP}}'(\varphi)=-\gamma_d^3/(\gamma_d+1-\gamma_d\varphi)^2<0$ on the relevant interval.

\textit{Change of variables and positivity.}
The spectral identity
$x\log x=\int_0^\infty\{x/(1+t)-x/(x+t)\}\,dt$, with $0\log0:=0$,
and $\tau\{S_2[D-R_D(\infty)]\}=0$ give
\begin{equation*}
\begin{aligned}
\omega_1&=\frac1m\int_0^\infty\varphi(t)[R_M(\infty)-R_M(t)]\,dt,\\
\omega_2-\omega_1&=\frac1m\int_0^\infty\varphi(t)
\{[R_{S_1}(\infty)-R_M(\infty)]-[R_{S_1}(t)-R_M(t)]\}\,dt,
\end{aligned}
\end{equation*}
where $m=\tau S_2=(\gamma_k/\gamma_d)\mu_1(\gamma_d+1)$.
The integrands are bounded at zero: $0\le\varphi\le1/\gamma_d$ and $|R_D(t)|\le\|D\|$.
At infinity, $\varphi(t)=m/t+O(t^{-2})$ and $R_D(t)=R_D(\infty)+O(t^{-1})$, so the integrands are $O(t^{-2})$.
These bounds justify the improper integrals and the change of variables.

Differentiating this expression for $t$ gives
\begin{equation*}
-\frac{dt}{dw}=\frac{\mu_1}{w^2}
\{f_{\mathrm{MP}}(\varphi)-wf_{\mathrm{MP}}'(\varphi)\varphi_H'(w)\}>0.
\end{equation*}
Here $f_{\mathrm{MP}}(0)=\gamma_d+1$, $f_{\mathrm{MP}}(1/\gamma_d)=0$, and for $0<u<1/\gamma_d$,
\begin{equation*}
\frac{f_{\mathrm{MP}}'(u)}{f_{\mathrm{MP}}(u)}
=-\frac{\gamma_d}{1-\gamma_du}-\frac{4\gamma_d}{\gamma_d+1-2\gamma_du}
+\frac{\gamma_d}{\gamma_d+1-\gamma_du}<0.
\end{equation*}
The population and Wishart identities reduce the normalized resolvent integrands to
\begin{equation*}
\begin{aligned}
\frac{\varphi}{m}[R_M(\infty)-R_M(t)]
&=\frac{w^2j_H'(w)}{\gamma_d+1},\\
\frac{\varphi}{m}\{[R_{S_1}(\infty)-R_M(\infty)]-[R_{S_1}(t)-R_M(t)]\}
&=\frac{\gamma_kw\chi_H(w)}{\gamma_d(\gamma_d+1)}
\{h_{\mathrm{MP}}(0)-\chi_H(w)h_{\mathrm{MP}}(\varphi)\}.
\end{aligned}
\end{equation*}
Multiplying by $-dt/dw$ proves both formulas in \eqref{sp:population-responses}.
Their integrands are strictly positive because $j_H'>0$, $f_{\mathrm{MP}}>0$, $f_{\mathrm{MP}}'<0$, and $\varphi_H'>0$,
together with the strict Wishart difference derived above. Thus $0<\omega_1<\omega_2$.
The subordination identities, unique root and finite integral representations also hold for constant $H$:
then $j_H'\equiv0$, so $\omega_1=0$, while the Wishart difference remains positive.
The leverage lemma below uses these identities for both constant and nonconstant laws.
\end{proof}

The same subordination parameter determines the compressed covariance trace.
\begin{lemma}[Trace of the compressed population covariance]\label{sp:leverage-bounds}
Under the assumptions of Lemma~\ref{sp:physical-limit}, along each subsequence with limiting law $H$, the variable $\tr{M_Q} /(r\mu_1)$ converges in probability
and in $L^1$ to the deterministic $h_M={\gamma_k} \{1-\chi_H({w_*} )\}$, which satisfies
\begin{equation*}
1\le h_M\le v_H,\qquad h_M=\gamma_k-w_*^{-1}<\gamma_k.
\end{equation*}
\end{lemma}
\begin{proof}

Let $P:=1_{(0,\infty)}(S_1)=1_{(0,\infty)}(S_2)$. The equality follows from
$\ker(W_{\mathrm{cov}}^{1/2}M^{1/2})=\ker(W_{\mathrm{cov}}M^{1/2})$.
The positive spectral gap in Lemma~\ref{sp:physical-limit} permits continuous approximation of $P$, and hence
\begin{equation*}
\frac{\tr M_Q}{r\mu_1}=\frac{\tr(MP)}{r\mu_1}
\xrightarrow{\Prob,L^1}\frac{\gamma_d}{\mu_1}\tau(MP),
\qquad \lambda_-\le\tr M_Q/r\le\lambda_+.
\end{equation*}
Under the subordination parametrization $w=\mu_1s$,
$\varphi(t)=\varphi_H(w)\to1/\gamma_d$ as $t\downarrow0$ and
$\mathcal U(s)=(\gamma_k/\gamma_d)\mu_1\chi_H(w)$.
Since $\gamma_kw\chi_H(w)$ increases strictly from $0$ to $\gamma_k>1$, $w\to w_*$ and
\begin{equation*}
\frac{\gamma_d}{\mu_1}\tau(MP)
\stackrel{(i)}{=}\frac{\gamma_d}{\mu_1}\lim_{t\downarrow0}\{\tau M-\mathcal U(s(t))\}
=\gamma_k\{1-\chi_H(w_*)\}=\gamma_k-w_*^{-1}=h_M.
\end{equation*}
Here (i) uses the conditional resolvent identity from Proposition~\ref{sp:positive-covariances}.

Jensen's inequality, respectively under $\operatorname{Law}(\varrho)$ and $\pi$, gives
\begin{equation*}
1=\gamma_k\E\frac{w_*\varrho}{1+w_*\varrho}\le\frac{\gamma_kw_*}{1+w_*},
\qquad
1=\gamma_kw_*\E_\pi\frac1{1+w_*\varrho}\ge\frac{\gamma_kw_*}{1+v_Hw_*}.
\end{equation*}
Thus $\gamma_k-v_H\le w_*^{-1}\le\gamma_k-1$, proving the bounds.
\end{proof}

\subsubsection{From spectral moments to the entropy response.}
For a positive compact population law supported in $[\lambda_-,\lambda_+]$, write
\begin{equation*}
D_H:=\frac{\mu_1\mu_3-\mu_2^2}{\lambda_+\mu_1^3}.
\end{equation*}
The next lemma turns ordinary spectral dispersion into a lower bound for the entropy-response ratio.

\begin{lemma}[Entropy response from ordinary spectral moments]
\label{sp:moment-dispersion}
For nonconstant $H$ supported in $[\lambda_-,\lambda_+]\subset(0,\infty)$ and $1<\gamma_k<\gamma_d$, the entropy-response ratio $\omega=\omega_1/\omega_2$ satisfies
\begin{equation*}
0<D_H\le v_H,\qquad
\omega\ge\left(1+\frac{C(\gamma_k,\gamma_d)}{D_H}\right)^{-1}.
\end{equation*}
Here $C(\gamma_k,\gamma_d)>0$ depends only on $\gamma_k,\gamma_d$, independently of $H$ and its spectral bounds.
\end{lemma}
\begin{proof}

Since $\Var_\pi\varrho=(\lambda_+/\mu_1)D_H\le(\lambda_+/\mu_1)v_H$, we have $0<D_H\le v_H$.
The independent-copy covariance identity and the derivative of $\log(1+w\varrho)$ give
\begin{equation*}
j_H(1/(2\gamma_k))
\stackrel{(i)}{\ge}\frac{\Var_\pi\varrho}{2\gamma_k+\lambda_+/\mu_1}
\stackrel{(ii)}{\ge}\frac{D_H}{2\gamma_k+1}.
\end{equation*}
Here (i) uses $\varrho\le\lambda_+/\mu_1$ and (ii) uses $\lambda_+/\mu_1\ge1$.
Because $\varphi_H(1/(2\gamma_k))\le1/(2\gamma_d)$ and $f_{\mathrm{MP}}'\le0$, the first integral in \eqref{sp:population-responses} gives
\begin{equation*}
\omega_1/\mu_1\ge
\frac{f_{\mathrm{MP}}(1/(2\gamma_d))}{\gamma_d+1}\,j_H(1/(2\gamma_k)).
\end{equation*}

Write $\chi:=\chi_H(w)$.
Jensen's inequality gives $\chi\ge(1+v_Hw)^{-1}$, while direct differentiation gives
\begin{equation*}
j_H'(w)=\frac{(1+v_Hw)\{\chi-(1+v_Hw)^{-1}\}}{w^2},
\qquad
\chi(1-\chi)\le\frac{v_Hw}{(1+v_Hw)^2}+\frac{w^2j_H'(w)}{1+v_Hw}.
\end{equation*}
Using $\chi\le\min\{1,(\gamma_kw)^{-1}\}$ and splitting the first integral at $1/(2\gamma_k)$ gives
\begin{equation*}
\begin{aligned}
\int_0^{w_*}\frac{\chi(1-\chi)}w\,dw
&\stackrel{(i)}{\le}1+\frac{j_H(1/(2\gamma_k))}{v_H}
+\int_{1/(2\gamma_k)}^\infty\frac{dw}{\gamma_kw^2}
=3+\frac{j_H(1/(2\gamma_k))}{v_H},\\
\int_0^{w_*}\chi^3\,dw
&\le\int_0^\infty\min\{1,(\gamma_kw)^{-3}\}\,dw=\frac3{2\gamma_k}.
\end{aligned}
\end{equation*}
In (i), integrate $v_H/(1+v_Hw)^2$ over the positive axis and use $w/(1+v_Hw)\le1/v_H$.
Moreover,
\begin{equation*}
0\le h_{\mathrm{MP}}(0)-h_{\mathrm{MP}}(z)\le\frac{\gamma_d^2z}{\gamma_d+1},
\quad f_{\mathrm{MP}}(z)\le\gamma_d+1,
\quad\int_0^{1/\gamma_d}-f_{\mathrm{MP}}'(z)\,dz=\gamma_d+1.
\end{equation*}
Substitute these bounds and $\varphi_H(w)=(\gamma_k/\gamma_d)w\chi$ into the second response integral:
\begin{equation*}
\begin{aligned}
\frac{\omega_2-\omega_1}{\mu_1}
&\le C(\gamma_k,\gamma_d)\left\{\int_0^{w_*}\frac{\chi(1-\chi)}w\,dw+\int_0^{w_*}\chi^3\,dw+1\right\}\\
&\le C(\gamma_k,\gamma_d)\left\{1+\frac{j_H(1/(2\gamma_k))}{v_H}\right\},\\
\frac{\omega_2-\omega_1}{\omega_1}
&\le C(\gamma_k,\gamma_d)\left\{\frac1{j_H(1/(2\gamma_k))}+\frac1{v_H}\right\}
\le\frac{C(\gamma_k,\gamma_d)}{D_H}.
\end{aligned}
\end{equation*}
The first line bounds the term containing $-f_{\mathrm{MP}}'$ by its total variation. The last uses the lower bound for the first response.
Rearranging proves the claimed ratio bound.
\end{proof}

\subsubsection{Teacher averages.}

The following identities separate teacher fluctuations from initialization fluctuations.
\begin{lemma}[Teacher-averaged trace identities and concentration]\label{rr:teacher}
For the independent-entry teacher and Gaussian initialization in the stated model, the exact finite-dimensional identities are
\begin{align*}
\E E_{aa}^2&=\sigma_a^2(1+2/k)\Sigma_0^2+\frac{\sigma_a^2}{k}\tr(\Sigma_0^2)I,\\
 \E E_{oa}^\top E_{oa}&=\sigma_o^2\Sigma_0^2,\\
\E \nu_a&=4\sigma_a^2(1+1/k)\tr\Sigma_0^2,\\
 \E \nu_o&=4\sigma_o^2\tr\Sigma_0^2,\\
\E \tr(S\Upsilon_a)&=\sigma_a^2({d_i}/r)k\{(1+2/k)\mu_{2,r}+\mu_{1,r}^2\},\\
 \E \tr(S\Upsilon_o)&=\sigma_o^2({d_i}/r)k\mu_{2,r},\\
\E\tr S^2&=k(1+1/r)\mu_{2,r}+(k^2/r)\mu_{1,r}^2.\end{align*}
Under the dimension and bounded-spectrum conditions of Assumptions~\ref{ass:dimensions} and \ref{ass:spectrum}, the following normalized traces and energies
converge in $L^2$, with uniformly bounded moments of every fixed order:
\begin{align*}
\nu _a/r&\longrightarrow4{\gamma_d} {\gamma_k} (\sigma_a^2+\sigma_o^2)f\mu_1,\\
 \nu _o/r&\longrightarrow4{\gamma_d} {\gamma_k} (\sigma_a^2+\sigma_o^2)(1-f)\mu_1,\\
\tr(S\Upsilon_a)/r&\longrightarrow {\gamma_d} {\gamma_k} (\sigma_a^2+\sigma_o^2)f\mu_1^2({v_H} +1),\\
 \tr(S\Upsilon_o)/r&\longrightarrow {\gamma_d} {\gamma_k} (\sigma_a^2+\sigma_o^2)(1-f)\mu_1^2{v_H} ,\\
\tr\Upsilon_a/r&\longrightarrow {\gamma_d} (\sigma_a^2+\sigma_o^2)f\mu_1(h_M+1),\\
\tr\Upsilon_o/r&\longrightarrow {\gamma_d} (\sigma_a^2+\sigma_o^2)(1-f)\mu_1h_M,\\
\tr S/r&\longrightarrow {\gamma_k} \mu_1,\\
\tr S^2/r&\longrightarrow {\gamma_k} \mu_1^2({v_H} +{\gamma_k} ).
\end{align*}
\end{lemma}
\begin{proof}
\textit{Conditional means.}
Expanding $(\Sigma_0Z_a^\top+Z_a\Sigma_0)^2$ gives four expectations:
$\sigma_a^2\Sigma_0^2/k$, $\sigma_a^2\Sigma_0^2$,
$\sigma_a^2\tr(\Sigma_0^2)I/k$, and $\sigma_a^2\Sigma_0^2/k$.
Also $\E_ZE_{oa}^\top E_{oa}=\sigma_o^2\Sigma_0^2$.
These give the stated matrix and energy identities and
\begin{equation*}
\begin{aligned}
\E_Z\tr(S\Upsilon_a)
&=\sigma_a^2(1+2/k)(d_i/r)\tr(SM_Q)+(\sigma_a^2/k)\tr(\Sigma_0^2)\tr S,\\
\E_Z\tr(S\Upsilon_o)&=\sigma_o^2(d_i/r)\tr(SM_Q),\\
\E_Z\tr\Upsilon_a
&=\sigma_a^2(1+2/k)(d_i/r)\tr M_Q+(\sigma_a^2r/k)\tr(\Sigma_0^2),\\
\E_Z\tr\Upsilon_o&=\sigma_o^2(d_i/r)\tr M_Q.
\end{aligned}
\end{equation*}
Since $S=\mathsf G_a^\top M_a\mathsf G_a/r$ and
$\tr(SM_Q)=r^{-1}\tr(\mathsf G_a^\top M_a^2\mathsf G_a)$, Gaussian pairings give
\begin{equation*}
\E\tr S=k\mu_{1,r},\qquad \E\tr(SM_Q)=k\mu_{2,r},\qquad
\E\tr S^2=(1+1/r)\tr M_a^2+r^{-1}(\tr M_a)^2.
\end{equation*}
Substituting these means proves the remaining exact identities.

\textit{Concentration and higher moments.}
For independent centered unit-variance sub-Gaussian entries $\xi_i$ and symmetric $D$,
\begin{equation*}
\Var(\xi^\top D\xi)=2\sum_{i\ne j}D_{ij}^2+\sum_i(\E\xi_i^4-1)D_{ii}^2\le C\|D\|_F^2.
\end{equation*}
If $\|F\|\le C/r$ and $\|F\|_F^2\le C/r$, the coefficient matrices of
$\tr(FE_{aa}^2)$ and $\tr(FE_{oa}^\top E_{oa})$, as quadratic forms in teacher entries, have Frobenius norm at most
\begin{equation*}
\frac{\sigma_a^2}{k}(2\|\Sigma_0\|)^2\|I_k\otimes F\|_F\le C/r,
\qquad
\left\|\frac{\sigma_o^2}{p}I_p\otimes(\Sigma_0F\Sigma_0)\right\|_F\le C/r.
\end{equation*}
The first uses the Frobenius operator norm $2\|\Sigma_0\|$ of $Z\mapsto\Sigma_0Z^\top+Z\Sigma_0$.
Thus each teacher-centered trace has $L^2(Z)$ norm $O(r^{-1})$. Hanson--Wright gives the same order in every fixed $L^p(Z)$.
For initialization, Gaussian quadratic-form variance and Poincar\'e give
\begin{equation*}
\begin{aligned}
\Var(\tr S)&=2r^{-1}\tr M_a^2\le C,\qquad
\Var(\tr(SM_Q))=2r^{-1}\tr M_a^4\le C,\\
\Var(\tr S^2)&\stackrel{(i)}{\le}\frac{16}{r^4}\E\|M_a\mathsf G_a(\mathsf G_a^\top M_a\mathsf G_a)\|_F^2\le C,\\
\left\|\frac{\tr M_Q}{r\mu_1}-h_M\right\|_{L^2}^2
&\stackrel{(ii)}{\le}C\left\|\frac{\tr M_Q}{r\mu_1}-h_M\right\|_{L^1}\longrightarrow0.
\end{aligned}
\end{equation*}
Here (i) is Gaussian Poincar\'e, while (ii) uses Lemma~\ref{sp:leverage-bounds} and $\lambda_-I\preceq M_Q\preceq\lambda_+I$.
Apply the teacher estimate with $F=I_k/r,Q_aQ_a^\top/r,Q_aSQ_a^\top/r$ on $\mathcal E_{\mathrm{init}}$,
then combine the conditional means with these initialization estimates.
This proves every stated $L^2$ limit.
Sub-Gaussian matrix norm moments bound all normalized traces in every fixed $L^p$. The Gaussian trace polynomials also obey Fact~\ref{fact:hypercontractivity}.
The global moment form of Lemma~\ref{lem:tail-absorption} removes $\mathcal E_{\mathrm{init}}^c$ without changing these limits.
\end{proof}

\subsubsection{Common coefficient margins.}
Define the rotation sign function, for $u>0$ and $h\ge1$, by
\begin{equation*}
\Psi(u;h):=-\gamma_k+2\{(1-g)v_H+f(1-\vartheta_a^2/\bar\vartheta_1)\}
+\frac2u\{(1-g)h+f(1-\vartheta_a^2/\bar\vartheta_1)\}
-\frac{v_H+\gamma_k}{2}u.
\end{equation*}
For the constants and rate intervals from Section~\ref{main:theorem}, set
\begin{equation*}
\begin{aligned}
C_*:=1+\max\biggl\{&1,\frac{C(\gamma_k,\gamma_d)\theta}{1-\theta},
\frac{\gamma_ku_{\max}-2f\vartheta_a/\bar\vartheta_1}{2-u_{\max}},\\
&\frac{\gamma_k(1+u_{\max}/2)-2f(1-\vartheta_a^2/\bar\vartheta_1)
-2\{1-g+f(1-\vartheta_a^2/\bar\vartheta_1)\}/u_{\max}}
{2(1-g)-u_{\max}/2}\biggr\}.
\end{aligned}
\end{equation*}
Here $C(\gamma_k,\gamma_d)$ is the fixed constant from Lemma~\ref{sp:moment-dispersion}.
Also $f(1-\vartheta_a^2/\bar\vartheta_1)>0$ and $2(1-g)-u_{\max}/2>0$.
The next lemma verifies that this sufficient spectral threshold gives all common margins.

\begin{lemma}[Common coefficient margins from the moment assumption]
\label{sp:moment-rates}
Under Assumptions~\ref{ass:dimensions}--\ref{ass:spectrum}, with $C_{\mathrm{sp}}\ge C_*$, every rate pair chosen in Section~\ref{main:theorem} satisfies
\begin{equation*}
\begin{gathered}
0<u_L<\theta<\omega<1<u_H<u_{\max}<2g\omega<2,\\
u_H<T_C<u_L+u_H<T_R,\qquad
\Psi(u_L;h_M)>\Psi(u_H;h_M)>0.
\end{gathered}
\end{equation*}
All strict margins are independent of rank and of the random draws.
\end{lemma}
\begin{proof}

The moment assumption and Lemma~\ref{sp:moment-dispersion} give
$v_H\ge D_H\ge C_*$ and $\omega\ge(1+C(\gamma_k,\gamma_d)/D_H)^{-1}>\theta$.
The last term in the definition of $C_*$ makes
\begin{equation*}
\Psi(u_{\max};1)
=\{2(1-g)-u_{\max}/2\}v_H-\gamma_k(1+u_{\max}/2)
+2f(1-\vartheta_a^2/\bar\vartheta_1)
+\frac{2\{1-g+f(1-\vartheta_a^2/\bar\vartheta_1)\}}{u_{\max}}>0.
\end{equation*}
Proposition~\ref{prop:common-parameters} gives $u_H<T_C<u_L+u_H<T_R$ and
$0<u_L<\theta<\omega<1<u_H<u_{\max}<2g\theta<2g\omega<2$.

Direct differentiation gives
\begin{equation*}
\partial_u\Psi=-\frac{2\{(1-g)h+f(1-\vartheta_a^2/\bar\vartheta_1)\}}{u^2}-\frac{v_H+\gamma_k}{2}<0,
\qquad \partial_h\Psi=\frac{2(1-g)}u>0.
\end{equation*}
Together with $h_M\ge1$, these signs give
$\Psi(u_L;h_M)>\Psi(u_H;h_M)\ge\Psi(u_H;1)>\Psi(u_{\max};1)>0$.
\end{proof}

\subsubsection{A nonempty class for the moment assumption.}
For a fixed $1<{b_{\circ}} <2$, the two-point laws
\begin{equation*}
H_\epsilon:=(1-\epsilon)\delta_{\epsilon^{b_{\circ}} }+\epsilon\delta_1,
\qquad 0<\epsilon<1,
\end{equation*}
have $\mu_1\sim\epsilon$ and
\begin{equation*}
\mu_1\mu_3-\mu_2^2
=\epsilon(1-\epsilon)\epsilon^{b_{\circ}} (1-\epsilon^{b_{\circ}} )^2,
\qquad D_{H_\epsilon}\sim\epsilon^{{b_{\circ}} -2}\longrightarrow\infty
\end{equation*}
with $\lambda_+=1$.
Selecting one sufficiently small fixed $\epsilon$ therefore satisfies Assumption~\ref{ass:spectrum} with $C_{\mathrm{sp}}\ge C_*$.
Its
smallest support point $\epsilon^{b_{\circ}} >0$ is fixed before rank grows.

\subsection{Training-set bounds and moment recursion}\label{app:training-set}
Fix a loss rule and a selected rate unless stated otherwise. Conditional on initialization, teacher, and the realized training set, define
\begin{equation*}
\bar G:=\E_I G_I,\qquad \overline{\mathcal H}:=\E_I\mathcal H_I,\qquad
\Gamma^{(0)}:=\eta\E_I G_I^\top G_I,\qquad \nu_m^\eta:=q\eta\nu_m.
\end{equation*}
Here $\E_I$ averages a uniform next index. $\E_{\mathrm{fresh}}$ averages an independent pair with initialization and teacher fixed.
Tensor products and operator norms use the Frobenius matrix-space convention in Section~\ref{app:foundation}.
The next event supplies the stability, forcing concentration, and mixed-map estimates used throughout the three observable proofs.

\begin{lemma}[A common static event]\label{er:training-set}\label{er:cross-map}
For every fixed $M>0$, there is an event $\mathcal E_\star\subseteq\mathcal E_{\mathrm{init}}$, measurable before the index histories are drawn, such that
$\Prob(\mathcal E_\star^c)\le C_M(qr)^{-M}$.
On this event the following bounds hold simultaneously for both losses and both selected rates, with fixed $\zeta,C>0$. The norm-bound constants may depend on $M$:
\begin{equation*}
\begin{gathered}
\max_{i,\alpha}\|x_{i,\alpha}\|\le C\sqrt q,\qquad
\max_{i,\alpha}\|\Phi(x_{i,\alpha})\|_F\le Cq,\qquad
\max_i\|\Pi y_i\|\le C\sqrt{r\Lambda_r},\\
0\preceq\eta\widetilde{\mathcal H}_i\preceq(2-\zeta)I,\qquad
\|\E_I\widetilde{\mathcal H}_I\|\le C,\qquad
\eta\E_I\|G_I\|_F^2\le Cr,\qquad \|\bar G\|_F\le C\sqrt r,\\
\max_i\|\mathcal H_i\|\le Cqr,\qquad
\E_I\|\mathcal H_IY\|_F^2\le Cqr\mathcal L_r\|Y\|_F^2,\qquad
\|\E_I G_IG_I^\top\|\le Cr^2\mathcal L_r,\\
\Gamma^{(0)}=\nu_m^\eta I+o_{\mathrm{op}}(1),\qquad c\le\nu_m^\eta\le C,\qquad
\bigl\|Y\mapsto\E_I G_I^\top\mathcal H_IY\bigr\|_{F\to F}\le Cqr\sqrt{\Lambda_r}.
\end{gathered}
\end{equation*}
The bounds involving $Y$ hold for every matrix of the rescaled coordinate dimensions.
For every fixed $s_{\max}$, there is a fixed $b$ such that $\|Y_t\|_F\le C(qr)^b$ for every history with $t\eta\le s_{\max}$.
\end{lemma}
\begin{proof}
\textit{Per-pair truncation.}
All calculations first condition on $\mathcal E_{\mathrm{init}}$. Independent-sum inequalities use independent truncation of each stored pair.
Gaussian tails, Hanson--Wright, and a union bound give
\begin{equation*}
\max_{i,\alpha}\|x_{i,\alpha}\|\le C_M\sqrt q,\quad
\max_{i,\alpha}\|a_{i,\alpha}\|\le C_M\sqrt r,\quad
\max_{i,\alpha}|y_{i,\alpha}|\le C_M\sqrt{r\Lambda_r}
\end{equation*}
outside probability $C_M(qr)^{-M}$: the label threshold is initially $C_M(\sqrt{r\Lambda_r}+\Lambda_r)$, and $\Lambda_r=o(r)$.
Since $\|\Pi\|\le1$ and $\Phi(x)=(xx^\top-I)/\sqrt2$, this also proves the first line of the statement.
Every polynomial used below has fixed degree and all fixed Gaussian moments bounded by a fixed power of $qr$, by Fact~\ref{fact:hypercontractivity}. Lemma~\ref{lem:tail-absorption}, conditional on initialization and teacher, therefore makes each truncation bias smaller than any prescribed negative power by increasing the tail exponent.

\textit{Hessian bounds.}
For each pair, Gaussian norm and inner-product concentration gives
\begin{equation*}
\left\|\left(\frac{\langle X_{i,\alpha} S^{1/2},X_{i,\beta} S^{1/2}\rangle_F}{q\tr S}\right)_{\alpha,\beta\in\{+,-\}}-I_2\right\|
\le C\left(\sqrt{\mathcal L_r/r}+\sqrt{\mathcal L_r/q}+r/q\right)=o(1).
\end{equation*}
Since $\eta_jq\tr S=u_j(1+o(1))$, the nonzero spectrum of $\eta_j\widetilde{\mathcal H}_i$ lies in $[\zeta,2-\zeta]$ for $\zeta:=\frac12\min\{u_L,2-u_H\}>0$ and sufficiently large $r$. This also gives the lower bound on $\operatorname{ran}\Pi$ used in the kernel representation.
For the empirical mean, $0\preceq\Pi\preceq I_2$ bounds each paired feature operator by the sum of its two position operators.
Treat the two positions separately: within each position the $n$ columns are independent across stored pairs.
In the nuisance basis of Section~\ref{app:moments}, the feature is $z a^\top S^{1/2}$, where $z\in\R^q$ denotes the nuisance coordinates and $a=Q^\top x$.
Every unit Frobenius projection has bounded $\psi_1$ norm by conditioning on these independent Gaussian factors (or by the Gaussian product moment bound).
Its norm is at most $C\sqrt{qr}$ on the truncation event.
Partition the sample indices into finitely many deterministic subsets, each of size at most a sufficiently small fixed multiple of $qr$.
Fact~\ref{fact:independent-columns}, including the diagonal Gram entries, bounds each subset Gram norm by $Cqr$. Summing and dividing by $n\asymp qr$ gives a bounded nuisance covariance.
For the signal feature, the ambient dimension is $O(r^2)$, its covariance is bounded by the Gaussian quadratic-form isometry, and its truncated squared norm is at most $Cr^2\mathcal L_r$.
Matrix Bernstein gives
\begin{equation*}
\|\E_I\widetilde{\mathcal H}_{I,\mathrm{sig}}\|
\le C+C\sqrt{r^2\mathcal L_r/n}+Cr^2\mathcal L_r/n\le C.
\end{equation*}
The inequality $\|v+w\|^2\le2\|v\|^2+2\|w\|^2$ combines the signal and nuisance blocks.
Finally, $\eta\widetilde{\mathcal H}_i^2\preceq2\widetilde{\mathcal H}_i$ and the bounded coordinate change by $S^{1/2}$ imply the stated bounds for $\mathcal H_i$.

\textit{Forcing moments and $\Gamma^{(0)}$.}
The forcing definition, Gaussian fourth moments, and scalar Bernstein after truncation give $\E_I\|G_I\|_F^2\le Cqr^2$.
Cauchy--Schwarz in the feature map gives $\|\bar G\|_F^2\le C\E_I\|\Pi y_I\|^2\le Cr$.
For a unit output vector $w$, use
\begin{equation*}
\|X_\alpha^\top w\|^2\le2(x_\alpha^\top w)^2\|a_\alpha\|^2+2,
\qquad \E_{\mathrm{fresh}}\|G_I^\top w\|^2\le Cr^2.
\end{equation*}
The truncated forcing outer products have norm at most $Cqr^2\mathcal L_r$. Matrix Bernstein yields $\|\E_I G_IG_I^\top\|\le Cr^2\mathcal L_r$.
Lemma~\ref{rr:source} gives $\E_{\mathrm{fresh}}G_{I,\mathrm{nuis}}^\top G_{I,\mathrm{nuis}}=q\nu_m I+O_{\mathrm{op}}(q)$.
For every unit $w\in\R^r$, $\|X_{\mathrm{sig}}w\|_{L^4(F)}\le C\sqrt r$ and $\|y\|_{L^4}\le C\sqrt r$, so the signal block contributes $O_{\mathrm{op}}(r^2)$.
Consequently
\begin{equation*}
\eta\E_{\mathrm{fresh}}G_I^\top G_I=q\eta\nu_m I+O_{\mathrm{op}}(r^{-1}+r/q),\qquad
\|\Gamma^{(0)}-\eta\E_{\mathrm{fresh}}G_I^\top G_I\|
\le C\sqrt{r\mathcal L_r/n}+Cr\mathcal L_r/n=o(1).
\end{equation*}
For the second bound, the truncated positive summands $\eta G_I^\top G_I$ have norm $Cr\mathcal L_r$ and bounded expectation.
The retained label covariance and $\|E\|_F^2\asymp r$ on $\mathcal E_{\mathrm{init}}$ give $\nu_m\asymp r$, hence $\nu_m^\eta\asymp1$.

\textit{Forcing-Hessian cross map.}
An elementary position term is the rank-one operator
\begin{equation*}
(y_{\alpha'}X_{\beta'}^\top X_\alpha)\otimes(X_\beta S),\qquad
\alpha',\beta',\alpha,\beta\in\{+,-\}.
\end{equation*}
Write its two factors locally as $A,B$. For every fixed $p$ and deterministic $K,Y$,
\begin{equation*}
\begin{aligned}
\|\langle A,K\rangle_F\|_{L^p}&\le C_pqr\|K\|_F,&
\|A\|_{L^p(F)}&\le C_pqr^{3/2},\\
\|\langle B,Y\rangle_F\|_{L^p}&\le C_p\|Y\|_F,&
\|B\|_{L^p(F)}&\le C_p\sqrt{qr}.
\end{aligned}
\end{equation*}
Indeed,
\begin{equation*}
X_{\beta'}^\top X_\alpha
=a_{\beta'}(x_{\beta'}^\top x_\alpha)a_\alpha^\top-a_{\beta'}a_{\beta'}^\top-a_\alpha a_\alpha^\top+I,
\end{equation*}
whose inner product with $K$ has $L^p$ norm at most $C_pq\sqrt r\|K\|_F$, including the trace term. Multiply by $\|y\|_{L^{2p}}\le C_p\sqrt r$.
The projection of $B$ is a centered quadratic form with coefficient Frobenius norm at most $C\|Y\|_F$.
H\"older's inequality now gives
\begin{equation*}
\|\E(A\otimes B)\|\le Cqr,\quad
\|\E[\|B\|_F^2(A\otimes A)]\|\le Cq^3r^3,\quad
\|\E[\|A\|_F^2(B\otimes B)]\|\le Cq^2r^3.
\end{equation*}
Using the weaker cutoff $|y|\le C_M\sqrt r\Lambda_r$, each truncated operator has norm at most $Cq^{3/2}r^2\Lambda_r$.
Apply Fact~\ref{fact:matrix-bernstein} to its self-adjoint dilation in dimension $O(qr)$:
\begin{equation*}
\begin{aligned}
\left\|\E_I(A_I\otimes B_I)-\E_{\mathrm{fresh}}(A\otimes B)\right\|
&\le C\sqrt{q^3r^3\Lambda_r/n}+Cq^{3/2}r^2\Lambda_r^2/n\\
&\stackrel{(i)}{\le}C(qr\sqrt{\Lambda_r}+\sqrt q\,r\Lambda_r^2)\stackrel{(ii)}{\le}Cqr\sqrt{\Lambda_r}.
\end{aligned}
\end{equation*}
Here (i) uses $n\asymp qr$ and (ii) uses $\Lambda_r^{3/2}/\sqrt q=o(1)$. The displayed fixed moments control the truncation bias.
Sum the finitely many position terms. Intersect all these events with $\mathcal E_{\mathrm{init}}$ to obtain $\mathcal E_\star$.
Finally, $\|I-\eta\widetilde{\mathcal H}_i\|\le1$ gives $\|L_t^0\|_F\le t\eta\max_i\|\widetilde G_i\|_F$. The bounded coordinate change proves the polynomial bound for every history.
\end{proof}

The conditional moments are the matrix $m_t:=\E_{\mathrm{hist}}L_t^0$ and the positive operator $V_t:=\E_{\mathrm{hist}}(L_t^0\otimes L_t^0)$.
Thus $\|V_t\|_*=\E_{\mathrm{hist}}\|L_t^0\|_F^2$. The nuclear norm here is that of an operator on Frobenius matrix space.
Use the weighted norm $\|(m,V)\|:=\|m\|_F/\sqrt r+\|V\|_*/r$.

\begin{lemma}[Euclidean moment recursion]\label{er:moment-recursion}
On $\mathcal E_\star$ there is a linear operator $\mathcal L_\eta$ on the direct sum of response matrices and self-adjoint operators on their Frobenius space, bounded in this weighted norm, such that
\begin{equation*}
\begin{gathered}
(m_{t+1},V_{t+1})=(I+\eta\mathcal L_\eta)(m_t,V_t)
+\eta\bigl(\bar G S^{1/2},\eta\E_I(\widetilde G_I\otimes\widetilde G_I)\bigr),\qquad (m_0,V_0)=0,\\
\|\mathcal L_\eta\|+\bigl\|\bigl(\bar G S^{1/2},\eta\E_I(\widetilde G_I\otimes\widetilde G_I)\bigr)\bigr\|\le C,\qquad
\|(m_t,V_t)\|\le Cs_te^{Cs_t}.
\end{gathered}
\end{equation*}
Moreover,
\begin{equation*}
(m_t,V_t)=\left(s_t I+\frac{s_t(s_t-\eta)}2\mathcal L_\eta\right)
\bigl(\bar G S^{1/2},\eta\E_I(\widetilde G_I\otimes\widetilde G_I)\bigr)+O(s_t^3e^{Cs_t}),
\end{equation*}
with increment remainder $O(\eta s_t^2e^{Cs_t})$ in the same norm.
\end{lemma}
\begin{proof}
The exact update $L^+=(I-\eta\widetilde{\mathcal H}_I)L+\eta\widetilde G_I$ and independence of the next index from the past give
\begin{equation*}
\begin{aligned}
m^+-m&=\eta(\bar G S^{1/2}-\E_I\widetilde{\mathcal H}_I[m]),\\
V^+-V&=\eta\{\bar G S^{1/2}\otimes m+m\otimes\bar G S^{1/2}
-\E_I(\widetilde{\mathcal H}_IV+V\widetilde{\mathcal H}_I)\}\\
&\quad+\eta^2\E_I\{\widetilde{\mathcal H}_IV\widetilde{\mathcal H}_I
-\widetilde{\mathcal H}_I[m]\otimes\widetilde G_I-\widetilde G_I\otimes\widetilde{\mathcal H}_I[m]
+\widetilde G_I\otimes\widetilde G_I\}.
\end{aligned}
\end{equation*}
These identities define $\mathcal L_\eta$ by collecting the terms linear in $(m,V)$.

For every self-adjoint $V$, signed spectral decomposition and $\eta\widetilde{\mathcal H}_I^2\preceq2\widetilde{\mathcal H}_I$ give
\begin{equation*}
\begin{aligned}
\eta\|\E_I(\widetilde{\mathcal H}_IV\widetilde{\mathcal H}_I)\|_*
&\stackrel{(i)}{\le}\eta\tr(|V|\E_I\widetilde{\mathcal H}_I^2)
\stackrel{(ii)}{\le}2\|\E_I\widetilde{\mathcal H}_I\|\|V\|_*,\\
\eta\E_I\|\widetilde{\mathcal H}_I[m]\|_F\|\widetilde G_I\|_F
&\le\{\eta\E_I\|\widetilde{\mathcal H}_I[m]\|_F^2\}^{1/2}
\{\eta\E_I\|\widetilde G_I\|_F^2\}^{1/2}\le C\sqrt r\|m\|_F.
\end{aligned}
\end{equation*}
Here (i) uses signed spectral decomposition and (ii) uses positivity and the trace/operator duality bound.
The other coefficients and the affine term are bounded by Lemma~\ref{er:training-set} in the weighted norm.

Iterating the affine recursion gives the geometric bound $C\eta\sum_{v<t}(1+C\eta)^v\le Cs_te^{Cs_t}$.
In that same geometric sum, the constant and linear terms of the binomial expansion are $s_t I$ and $s_t(s_t-\eta)\mathcal L_\eta/2$. Its remaining terms are $O(s_t^3e^{Cs_t})$.
Taking one increment of the sum gives the stated $O(\eta s_t^2e^{Cs_t})$ remainder.
\end{proof}

\subsection{Proof of Proposition~\ref{prop:first-order-er}}\label{app:er}
We compute the mean drift, control sampled trajectories, and compare the expected increment with the derivative at the conditional mean.
Throughout this subsection, the rule $m$ and rate $j$ are fixed and suppressed, and constants follow Section~\ref{app:foundation}.
The static event $\mathcal E_\star$ is established in Lemma~\ref{er:training-set} before histories are sampled.
Use $\Gamma^{(0)}$ and $\nu_m^\eta$ from Lemma~\ref{er:training-set}.
Conditional on initialization, task, and the realized training set, write $\bar\Gamma_t:=\E_{\mathrm{hist}}\Gamma_t$ and
\begin{equation}\label{er:gram-drift}
\Gamma^{(1)}:=\bar G^\top\bar G-\tfrac\eta2(M_1+M_1^\top+M_2+M_2^\top)+\tfrac{\eta^2}2M_3,
\end{equation}
where independent uniform stored-pair indices $i,i'$ define
\begin{equation*}
M_1:=\E_{i,i'}G_i^\top\mathcal H_{i'}[G_i],\quad
M_2:=\E_{i,i'}G_{i'}^\top\mathcal H_{i'}[G_i],\quad
M_3:=\E_{i,i'}(\mathcal H_{i'}[G_i])^\top\mathcal H_{i'}[G_i].
\end{equation*}
The energies and compressed task matrices are defined in Section~\ref{app:moments}.

\subsubsection{Shared drift moments and the conditional mean}
In the next calculation $\E_{\mathrm{train}}$ averages the training set conditional on initialization and task.
Unqualified expectations average fresh pairs with the same conditioning.
Let $G^{\mathrm{pop}}:=\E G$. The channel coefficients used also in Section~\ref{app:rotation} are
\begin{equation*}
\begin{array}{c|ccc}
\vartheta_{m,\bullet}&a&o&\mathrm{nuis}\\\hline
R&1&1&1\\
C&\vartheta_a&\vartheta_o&1/2
\end{array}
\qquad \hat u:=\eta q\tr S.
\end{equation*}
The four expectations below determine both effective-rank and angular drift.
\begin{lemma}[Shared drift moments under repeated sampling]\label{er:phase}
Conditional on initialization and task satisfying Lemma~\ref{er:init},
\begin{equation}\label{er:drift-blocks}
\begin{aligned}
\E_{\mathrm{train}}\bar G^\top\bar G
&=(1-n^{-1})4\sum_{\bullet\in\{a,o\}}\vartheta_{m,\bullet}^2\Upsilon_\bullet
  +\frac{q\nu_m}{n}I+o_{\mathrm{op}}(1),\\
\tfrac\eta2\E_{\mathrm{train}}(M_1+M_1^\top)
&=(1-n^{-1})2\vartheta_{m,\mathrm{nuis}}\nu_m^\eta S
  +\frac{q\nu_m}{n}\hat u I+o_{\mathrm{op}}(1),\\
\tfrac\eta2\E_{\mathrm{train}}(M_2+M_2^\top)
&=(1-n^{-1})4q\eta\sum_{\bullet\in\{a,o\}}\vartheta_{m,\bullet}^2\tr(S\Upsilon_\bullet)I
  +\frac{q\nu_m}{n}\hat u I+o_{\mathrm{op}}(1),\\
\tfrac{\eta^2}2\E_{\mathrm{train}}M_3
&=(1-n^{-1})\vartheta_{m,\mathrm{nuis}}(q\eta)^2\nu_m\tr(S^2)I
  +\frac{q\nu_m}{2n}\hat u^2I+o_{\mathrm{op}}(1).
\end{aligned}
\end{equation}
Every remainder has norm at most $C(r^{-1/2}+r^2/q)$. The first has the sharper bound $C(r^{-1}+r/q)$.
Consequently
\begin{equation}\label{er:phase-drift}
\begin{aligned}
\E_{\mathrm{train}}\Gamma_R^{(1)}&=4(\Upsilon_a+\Upsilon_o)-2\nu_R^\eta S+O(1)I+o_{\mathrm{op}}(1),\\
\E_{\mathrm{train}}\Gamma_C^{(1)}&=4(\vartheta_a^2\Upsilon_a+\vartheta_o^2\Upsilon_o)-\nu_C^\eta S+O(1)I+o_{\mathrm{op}}(1),
\end{aligned}
\end{equation}
where each $O(1)I$ is a scalar multiple of the identity and the same remainder bound holds.
\end{lemma}
In \eqref{er:phase-drift}, $4(\vartheta_{m,a}^2\Upsilon_a+\vartheta_{m,o}^2\Upsilon_o)$ records coherent fitting along task-aligned directions, while
$-2\vartheta_{m,\mathrm{nuis}}\nu_m^\eta S$ comes from Hessian feedback $-\eta(M_1+M_1^\top)/2$ on nuisance gradient noise.
Their competition determines the leading effective-rank response at the scalar Gram reference. Scalar multiples of $I$ make no contribution because $D\operatorname{er}_T(\lambda I)[I]=0$.

\begin{proof}
\textit{Distinct pairs ($i\ne i'$).}
Each empirical double average splits with weights $1-n^{-1}$ and $n^{-1}$ according as $i\ne i'$ or $i=i'$.
Lemma~\ref{rr:source} and Gaussian pairing give
\begin{equation*}
\begin{aligned}
G_R^{\mathrm{pop}}&=2EQ,&G_C^{\mathrm{pop}}&=(E-R_\times ER_\times)Q,\\
\E\mathcal H_R[Y]&=2(YS+QSY^\top Q),&
\E\mathcal H_C[Y]&=(I-\rho_aR_\times)YS+(1-\rho_a^2)QSY^\top Q,\\
(G^{\mathrm{pop}})^\top G^{\mathrm{pop}}&=4\sum_\bullet\vartheta_{m,\bullet}^2\Upsilon_\bullet,&
\E G^\top G&=q\nu_mI+O_{\mathrm{op}}(q+r^2).
\end{aligned}
\end{equation*}
Thus the first row of \eqref{er:drift-blocks} is exact up to $O((q+r^2)/n)$.
For independent copies $(G,\mathcal H),(G',\mathcal H')$, the nuisance action and the signal fourth moments yield
\begin{equation*}
\begin{aligned}
\E G^\top\mathcal H'[G]
&=2\vartheta_{m,\mathrm{nuis}}q\Omega_mS+O_{\mathrm{op}}(r^3),\\
\sum_{\alpha,\beta}(\Pi^2)_{\alpha\beta}
\E[y_\alpha\langle X_\beta,G^{\mathrm{pop}}S\rangle_F]
&=8\sum_\bullet\vartheta_{m,\bullet}^2\tr(S\Upsilon_\bullet),\\
\E G^\top\mathcal H[G^{\mathrm{pop}}]
&=4q\sum_\bullet\vartheta_{m,\bullet}^2\tr(S\Upsilon_\bullet)I+O_{\mathrm{op}}(q+r^3).
\end{aligned}
\end{equation*}
Here $\Omega_m=\nu_mI+O_{\mathrm{op}}(1)$ by \eqref{rr:source-covariances}.
For the middle identity, $\langle X_\beta,G^{\mathrm{pop}}S\rangle_F$ is the centered quadratic form with coefficient
$\operatorname{sym}(G^{\mathrm{pop}}SQ^\top)$. Applying Fact~\ref{fact:isserlis} gives its stated two-position contraction.
The free-index contractions have norm $O(q)$, and signal output contributes $O(r^3)$, giving the last remainder.

For the squared-Hessian term, the two required contractions are
\begin{equation*}
\begin{aligned}
\E[(z_\alpha^\top z_{\alpha'})(z_\beta^\top z_{\beta'})]
&=q^2\mathbf1_{\{\alpha=\alpha'\}}\mathbf1_{\{\beta=\beta'\}}
+q\bigl(\mathbf1_{\{\alpha=\beta\}}\mathbf1_{\{\alpha'=\beta'\}}
+\mathbf1_{\{\alpha=\beta'\}}\mathbf1_{\{\alpha'=\beta\}}\bigr),\\
\E[a_\alpha a_\alpha^\top(a_\beta^\top S\Omega_mSa_\beta)]
&=\tr(S\Omega_mS)I+2(R_a)_{\alpha\beta}^2S\Omega_mS.
\end{aligned}
\end{equation*}
Since $\|S\Omega_mS\|\le Cr$ and $\tr(S\Omega_mS)\le Cr^2$, the nuisance part of
$\frac{\eta^2}{2}\E(\mathcal H'[G])^\top\mathcal H'[G]$ is
$\frac{(\eta q)^2}{2}\|\Pi\|_F^2\tr(S\Omega_mS)I+O_{\mathrm{op}}(r^{-1}+q^{-1})$.
A signal output or inner product contributes at most $C\eta^2(qr^4+r^5)\le Cr^2/q$:
the forcing covariance, signal feature energy, and full feature energy are bounded by $Cr,Cr^2,Cqr$, respectively.
Using $\|\Pi\|_F^2=2\vartheta_{m,\mathrm{nuis}}$ now proves all distinct-pair terms.

\textit{Repeated pairs ($i=i'$).}
For every unit vector $w$ and fixed $p$, Gaussian norm and quadratic-form moments give
\begin{equation*}
\begin{aligned}
\left\|\frac{(\langle X_\alpha,X_\beta S\rangle_F)_{\alpha\beta}}{q\tr S}-I_2\right\|_{L^p}
&\le C_p(r^{-1/2}+q^{-1/2}+r/q),\\
\left\|\frac{(\langle X_\alpha w,X_\beta w\rangle)_{\alpha\beta}}q
-\diag((a_+^\top w)^2,(a_-^\top w)^2)\right\|_{L^p}
&\le C_p(q^{-1/2}+r/q),\qquad \|y\|_{L^p}\le C_p\sqrt r.
\end{aligned}
\end{equation*}
Indeed $G=\mathcal X[\Pi y]/2$ and
$\eta\mathcal H[G]=\mathcal X[\Pi\eta(\langle X_\alpha,X_\beta S\rangle_F)_{\alpha\beta}\Pi y]/2$.
Substitution into the quadratic forms of $G^\top G$, $\frac\eta2\{G^\top\mathcal H[G]+\mathcal H[G]^\top G\}$,
and $\frac{\eta^2}2\mathcal H[G]^\top\mathcal H[G]$ gives, after division by $n$,
\begin{equation*}
\frac{q\nu_m}{n}I,\qquad
\frac{q\nu_m}{n}\hat u I,\qquad
\frac{q\nu_m}{2n}\hat u^2I,
\end{equation*}
each with operator error $C(r^{-1/2}+r^2/q)$.
H\"older controls the substitutions uniformly in $w$. Pairing the remaining signal coordinates leaves a scalar part $\nu_m$
and a bounded matrix remainder, whose factor $q/n=O(r^{-1})$ vanishes.
Both $M_1$ and $M_2$ have this same repeated-pair contraction.

Combining the four rows with \eqref{er:gram-drift} gives the drift expansion. All terms except the displayed task matrices and $S$ are bounded scalar multiples of $I$.
\end{proof}

The matrix moment recursion controls the conditional mean Gram matrix.
\begin{lemma}[Bounded affine moment evolution]\label{er:affine}
On $\mathcal E_\star$, for every integer $t\ge1$ with $s_t=t\eta\le1$,
\begin{equation*}
\|\Gamma^{(1)}\|_*/r\le C,\qquad
\|\bar\Gamma_t-\Gamma^{(0)}\|_*\le Crs_t,\qquad
\|\bar\Gamma_{t+1}-\bar\Gamma_t-\eta\Gamma^{(1)}\|_*/r\le C\eta s_t.
\end{equation*}
\end{lemma}
\begin{proof}

Compress the covariance expansion of Lemma~\ref{er:moment-recursion} to $Y_t^\top Y_t/s_t$.
Its bounded generator and increment remainder give
\begin{equation*}
\|\bar\Gamma_t-\Gamma^{(0)}-(s_t-\eta)\Gamma^{(1)}\|_*\le Crs_t^2,
\qquad
\|\bar\Gamma_{t+1}-\bar\Gamma_t-\eta\Gamma^{(1)}\|_*\le Cr\eta s_t.
\end{equation*}

Since $Y_2=\eta G_i+\eta G_{i'}-\eta^2\mathcal H_{i'}[G_i]$, expanding
$\E_{i,i'}Y_2^\top Y_2/(2\eta)=\Gamma^{(0)}+\eta\Gamma^{(1)}$
gives \eqref{er:gram-drift}, its $Cr$ coefficient bound, and the discrete offset.
\end{proof}

The strict rate margins turn the population expansion into a persistent signed entropy derivative.
\begin{lemma}[Sign and persistence of the effective-rank drift at the mean Gram matrix]\label{er:mean-phase}
Suppose $0<u_L<\omega<u_H<2g\omega$ with fixed strict margins, where $\omega=\omega_1 /\omega_2 $.
There exist constants $\kappa,c_N>0$ such that, for all sufficiently large $r$, the
expected reference drift $\E[\mathbf1_{\mathcal E_\star }D\operatorname{er}_T(\Gamma^{(0)})[\Gamma^{(1)}]]$ is at least $\kappa$ for raw high rate
and at most $-\kappa$ for raw low rate and both centered rates.
For every integer $t\ge1$ with $0<s_t\le c_N/r$, the expected increment
\begin{equation*}
\E[\mathbf1_{\mathcal E_\star }D\operatorname{er}_T(\bar \Gamma_t)[\bar \Gamma_{t+1}-\bar \Gamma_t]]
\end{equation*}
is at least $\kappa\eta$ for raw high rate and at most $-\kappa\eta$ for raw low rate and both centered rates.
\end{lemma}
\begin{proof}

Lemma~\ref{rr:teacher} gives
\begin{equation*}
\E_Z\Upsilon_a=\sigma_a^2(1+2/k)(d_i/r)M_Q+(\sigma_a^2/k)\tr\Sigma_0^2I,
\qquad \E_Z\Upsilon_o=\sigma_o^2(d_i/r)M_Q,
\end{equation*}
and $\nu_a=4\sigma_a^2(1+1/k)\tr\Sigma_0^2+o_{L^2}(r)$,
$\nu_o=4\sigma_o^2\tr\Sigma_0^2+o_{L^2}(r)$.
The scalar reference in Lemma~\ref{er:training-set}, the affine bound, and Lemma~\ref{er:calculus} imply
\begin{equation*}
\left|D\operatorname{er}_T(\Gamma^{(0)})[\Gamma^{(1)}]
-D\operatorname{er}_T(\nu_m^\eta I)[\Gamma^{(1)}]\right|
\le C\|\Gamma^{(0)}-\nu_m^\eta I\|\,\|\Gamma^{(1)}\|_*/r=o(1).
\end{equation*}
Inserting $\mathbf1_{\mathcal E_\star}$ into the conditional training-set polynomial averages in
\eqref{er:phase-drift} costs $o(1)$ by Lemma~\ref{lem:tail-absorption}. Initialization and teacher remain restricted as above.
Apply the scalar derivative formula to this expansion.
Since $\mathcal C_T(I)=0$, the raw and centered factors, apart from the common positive factor
$\operatorname{er}_T(I)\omega_2$, are respectively $2(1-\omega/u)$ and $1-2g\omega/u$,
where $\eta=u/(\gamma_k\mu_{1,r}qr)$.
Lemma~\ref{sp:physical-limit} gives joint convergence with $\operatorname{er}_T(I)$, and
Proposition~\ref{sp:positive-covariances} gives $\omega_2>\omega_1>0$.
The strict rate margins therefore determine the reference signs.

For persistence, Lemmas~\ref{er:training-set}, \ref{er:affine}, and~\ref{er:calculus} give
$\|\bar\Gamma_t-\Gamma^{(0)}\|\le Crs_t$, $\|\bar\Gamma_{t+1}-\bar\Gamma_t\|_*\le Cr\eta$, and
\begin{equation*}
\left|D\operatorname{er}_T(\bar\Gamma_t)[\bar\Gamma_{t+1}-\bar\Gamma_t]
-\eta D\operatorname{er}_T(\Gamma^{(0)})[\Gamma^{(1)}]\right|
\le C\eta(rs_t+s_t+r\eta).
\end{equation*}
Choose $c_N$ small so $rs_t\le c_N$ preserves the spectral bounds and strict margin, then take $r$ large so $s_t+r\eta=o(1)$.
\end{proof}

\subsubsection{Control of sampled trajectories}

Condition on the bounded initialization and teacher in Lemma~\ref{er:init} and evaluate Gaussian moments before event restriction.
For a fixed history with bounded counts, $Y_t^{[0]},Y_t^{[1]}$ are fixed-degree Gaussian polynomials.
Independent pair truncations preserve the independent-sum inequalities. Lemma~\ref{er:polynomial} transfers their moment bounds to the localized events.

Bounded visit counts keep repeated updates from increasing the Gaussian polynomial degree with the horizon.
\begin{lemma}[Visit counts]\label{er:visit-counts}
Let $N_i(t):=\sum_{v=0}^{t-1}\mathbf1_{\{I_v=i \}}$ count the visits to pair $i $ and let $N=\lfloor c_N q\rfloor$.
For every fixed $M>0$, there exist constants
$N_{\max},C_{N_{\max}},C_M$ such that, with probability at least $1-C_M(qr)^{-M}$,
\begin{equation*}
\max_iN_i (t)\le N_{\max},\qquad
\#\{i :N_i (t)\ge2\}\le C_{N_{\max}}\{t^2/n+\log(qr)\}
\end{equation*}
simultaneously for every $t\le N+1$.
\end{lemma}
\begin{proof}

Choose $N_{\max}$ after the confidence exponent.
A union bound and the binomial factorial-moment bound give
\begin{equation*}
\Prob(\max_iN_i (N+1)\ge N_{\max})
\le n\left(\frac{e(N+1)}{N_{\max}n}\right)^{N_{\max}}\le C_M(qr)^{-M},
\end{equation*}
using $N/n=O(r^{-1})$.

At each of the first $t$ draws, the chance of a repeat is at most $t/n$.
A binomial domination and Chernoff bound imply
\begin{equation*}
\Prob\left(\#\{i :N_i (t)\ge2\}>C\{t^2/n+M\log(qr)\}\right)
\le C(qr)^{-M-1}.
\end{equation*}
Union over the polynomial number of prefixes yields the claimed simultaneous bounds.
\end{proof}

Use the rescaled feature maps $\mathcal X_i$ and their Euclidean versions
$\widetilde{\mathcal X}_i$ from Section~\ref{app:reduction}.
The coefficient vector for pair $i$ is $\tfrac12\Pi y_i$.
For the sampled indices $i_v=I_{v-1}$, the two strictly lower block operators are
\begin{equation*}
\begin{aligned}
(\mathcal T_t^{\mathrm{cross}})_{v,u}&=\widetilde{\mathcal{X}}_{i_v}^*\widetilde{\mathcal{X}}_{i_u}\mathbf1_{\{u<v,\ i_u\ne i_v\}},\\
(\mathcal T_t^{\mathrm{same}})_{v,u}&=\widetilde{\mathcal{X}}_{i_v}^*\widetilde{\mathcal{X}}_{i_u}\mathbf1_{\{u<v,\ i_u=i_v\}}.
\end{aligned}
\end{equation*}
Write $\mathcal T_t^{\mathrm{rep}}:=(I+\eta \mathcal T_t^{\mathrm{same}})^{-1}$ and $y_{[t]}:=(\tfrac12\Pi y_{i_v})_{v\le t}$.
For each integer $h \ge0$, the term with exactly $h $
interactions between distinct training pairs is
\begin{equation*}
Y_t^{[h ]}=\eta\left\{\sum_{v\le t}\widetilde{\mathcal{X}}_{i_v}
[(-\eta \mathcal T_t^{\mathrm{rep}}\mathcal T_t^{\mathrm{cross}})^h\mathcal T_t^{\mathrm{rep}}y_{[t]}]_v\right\}S^{-1/2},\qquad h \ge0,
\end{equation*}
Define
\begin{equation}\label{er:interaction-decomposition}
Y_t=Y_t^{[0]}+Y_t^{[1]}+Y_t^{[\ge2]},\qquad
Y_t^{[\ge2]}=\sum_{h \ge2}Y_t^{[h ]}.
\end{equation}
The repeated-update inverse includes every same-pair update, and $Y_t^{[h]}$ has exactly $h$ distinct-pair interactions.
Strict triangularity makes the expansion finite.

On the geometry window $t\in[t_0,N]$, reaching $t=\Theta(q)$ and $s_t=O(1/r)$, cumulative feedback is small but still affects the leading observable increments.
The interaction expansion separates $Y_t^{[0]}$ and $Y_t^{[1]}$, which are bounded-degree Gaussian polynomials on bounded-visit histories, from the contractive tail $Y_t^{[\ge2]}$.
The next lemma bounds the interaction operators and the tail containing two or more interactions.
\begin{lemma}[Interaction operators for earlier and later updates]\label{er:kernel}
For every fixed $M>0$, with probability at least $1-C_M(qr)^{-M}$, the bounds
\begin{equation*}
\|\eta \mathcal T_t^{\mathrm{cross}}\|\le C_{N_{\max}}\log(2t)\sqrt{\frac t{qr}}\log\!\left(e\sqrt{\frac{qr}{t}}\right)+C_{N_{\max}}s_t\log(2t)\le \mathcal L_r\sqrt{s_t}
\end{equation*}
and
\begin{equation*}
\|\mathcal T_t^{\mathrm{rep}}\|\le C,\qquad
\|Y_t^{[\ge2]}\|_F\le \mathcal L_r\sqrt r\,s_t^{3/2}
\end{equation*}
hold simultaneously for every $t_0\le t\le N+1$.
\end{lemma}
\begin{proof}

\textit{Resolvent expansion.}
Recursive substitution of the Euclidean update gives
\begin{equation*}
L_t^0=\eta\sum_{v\le t}\widetilde{\mathcal X}_{i_v}
[(I+\eta\mathcal T_t^{\mathrm{same}}+\eta\mathcal T_t^{\mathrm{cross}})^{-1}y_{[t]}]_v.
\end{equation*}
Factor the inverse as $(I+\eta\mathcal T_t^{\mathrm{rep}}\mathcal T_t^{\mathrm{cross}})^{-1}\mathcal T_t^{\mathrm{rep}}$.
Its finite triangular expansion is exactly \eqref{er:interaction-decomposition}.

\textit{Cross-pair operator.}
For a block Gram matrix, dyadic rectangles partition the strict lower triangle into at most $\lceil\log_2(2t)\rceil$ levels,
with disjoint input/output supports within each level. Hence triangular restriction multiplies its norm by at most this factor.
To apply Fact~\ref{fact:independent-columns}, independently select one position of each distinct visited pair with probability $1/2$.
The paired off-diagonal Gram equals four times the selection average of the selected Gram embedded in the full coefficient space, since each ordered distinct-pair position choice survives with probability $1/4$.
Selected columns are independent across pairs and satisfy the fact's moment hypotheses.
For a fixed admissible history there are at least $t/N_{\max}$ distinct pairs.
Fact~\ref{fact:independent-columns} bounds the joint probability of failure on the per-pair norm event by $Ce^{-c\sqrt t}$.
Markov's inequality makes the conditional fraction of bad selections at most $Ce^{-c\sqrt t/2}$ outside probability $Ce^{-c\sqrt t/2}$.
The Gram entries are degree-four Gaussian polynomials. On the common per-pair norm event their norms have a deterministic polynomial bound.
The contribution of bad selections is therefore smaller than every fixed negative power, and averaging preserves the claimed bound.
The bounded projection $\Pi$ and visit multiplicities change only constants.
Dyadic restriction now gives the first term of the cross-operator estimate. For signal features, Matrix Bernstein gives
\begin{equation*}
\bigl\|(\widetilde{\mathcal X}_{i_v}^*P_{\mathrm{sig}}\widetilde{\mathcal X}_{i_u})_{v,u\le t}\bigr\|
\stackrel{(i)}{\le}C\{t+\sqrt{tr^2\mathcal L_r}+r^2\mathcal L_r\}
\stackrel{(ii)}{\le}Ct.
\end{equation*}
Here (i) uses covariance norm $C$ and independently truncated squared feature norms $Cr^2\mathcal L_r$,
while (ii) uses $t\ge t_0\gg r^2\mathcal L_r$. Triangular restriction gives the additional $Cs_t\log(2t)$ term.

\textit{Same-pair inverse and tail.}
For one stored pair, write $K:=\widetilde{\mathcal X}_i^*\widetilde{\mathcal X}_i$ on $\operatorname{ran}\Pi$.
The per-pair Gram bounds in Lemma~\ref{er:training-set} give $\zeta I\preceq\eta K\preceq(2-\zeta)I$.
The repeated-update inverse has diagonal $I$ and $h$th subdiagonal $-\eta K(I-\eta K)^{h-1}$, so
\begin{equation*}
\|\mathcal T_t^{\mathrm{rep}}\|
\stackrel{(i)}{\le}1+\sum_{h\ge1}\|\eta K(I-\eta K)^{h-1}\|
\le1+2\sum_{h\ge0}(1-\zeta)^h=1+2/\zeta.
\end{equation*}
Here (i) applies the block Schur test (Fact~\ref{fact:block-schur}).
The feature-column map has norm $C\sqrt{qr}$ and $\|y_{[t]}\|\le C\sqrt{tr\mathcal L_r}$.
Because $\|\eta\mathcal T_t^{\mathrm{rep}}\mathcal T_t^{\mathrm{cross}}\|\le C\mathcal L_r\sqrt{s_t}=o(1)$,
\begin{equation*}
\|Y_t^{[\ge2]}\|_F
\stackrel{(i)}{\le}C\eta\sqrt{qr}\sqrt{tr\mathcal L_r}
\sum_{h\ge2}(C\mathcal L_r\sqrt{s_t})^h
\stackrel{(ii)}{\le}\mathcal L_r\sqrt r\,s_t^{3/2}.
\end{equation*}
Here (i) is the preceding inverse expansion and (ii) uses $\eta qr\asymp1$, absorbing fixed logarithmic powers into $\mathcal L_r$.
Independent truncation and the union over polynomially many prefixes preserve the stated confidence by Lemma~\ref{lem:tail-absorption}.
\end{proof}

The following moment inequalities control independent matrix sums and restrictions to large-probability events.
The symbol $P$ denotes a polynomial local to the next lemma.
\begin{lemma}[Fourth moments and restriction to high-probability events]\label{er:polynomial}
For independent centered rectangular matrices $Z_i$ of the same dimensions and with finite fourth moments,
\begin{equation*}
\E\tr\!\left[\left\{\left(\sum_i Z_i\right)^\top\left(\sum_i Z_i\right)\right\}^2\right]
\le C\left\{\tr\left(\sum_i\E Z_i^\top Z_i\right)^2+
\tr\left(\sum_i\E Z_i Z_i^\top\right)^2+
\sum_i\E\|Z_i\|_F^4\right\}.
\end{equation*}
If $P$ is a nonnegative Gaussian polynomial of fixed degree, then $\|P\|_{L^p}\le C_p\E P$ for every fixed $p\ge1$, where $C_p$ depends only on $p$ and the degree bound.
There is also a
constant $\epsilon>0$, depending only on the degree bound, such that, for every event $\mathcal E_{\mathrm{loc}}$ with $\Prob(\mathcal E_{\mathrm{loc}}^c)\le\epsilon$,
\begin{equation*}
\E P\le2\E(P\mathbf1_{\mathcal E_{\mathrm{loc}}}).
\end{equation*}
\end{lemma}
\begin{proof}

Expand $\sum_{i,j,k,l}\E\tr(Z_i^\top Z_jZ_k^\top Z_l)$.
Independence and centering kill every singleton index. The two paired patterns
$i=j,k=l$ and $i=l,j=k$ give the two variance traces. The remaining pattern satisfies
$|\tr(Z_i^\top Z_jZ_i^\top Z_j)|\le\tr(Z_i^\top Z_jZ_j^\top Z_i)$ by Frobenius Cauchy--Schwarz.
All-equal terms are bounded by $\sum_i\E\|Z_i\|_F^4$.

Hypercontractivity gives $\|P\|_4\le C\|P\|_2$, while interpolation gives
$\|P\|_2\le\|P\|_1^{1/3}\|P\|_4^{2/3}\le C\|P\|_1^{1/3}\|P\|_2^{2/3}$.
Hence $\|P\|_2\le C\E P$ and then $\|P\|_p\le C_p\E P$.
Consequently $\E(P\mathbf1_{\mathcal E_{\mathrm{loc}}^c})\le
\|P\|_2\sqrt{\Prob(\mathcal E_{\mathrm{loc}}^c)}\le\frac12\E P$ for sufficiently small failure probability.
\end{proof}

These fourth-moment bounds compare the response with its accumulated forcing and control the Taylor remainder.
\begin{lemma}[Fourth moments of the difference between the response and accumulated forcing]\label{er:feedback}
Let $G_{[t]}:=\eta\sum_{v<t}G_{I_v}$ be the accumulated additive forcing, and let $\Delta_t:=Y_t-G_{[t]}$ be the correction due to the state-dependent terms.
For every $t_0\le t\le N+1$,
with $t_0$ as in Section~\ref{main:theorem},
\begin{equation*}
\E[\mathbf1_{\mathcal E_\star }\|\Delta_t \|^4]\le \mathcal L_rrs_t^4,\qquad
\E[\mathbf1_{\mathcal E_\star }\|\Delta_t \|_F^4]\le \mathcal L_rr^2s_t^4.
\end{equation*}
\end{lemma}
\begin{proof}

\textit{Repeated-visit correction.}
Fix a history with the visit-count bounds of Lemma~\ref{er:visit-counts}.
For each pair set $K_i:=\widetilde{\mathcal X}_i^*\widetilde{\mathcal X}_i$ and
\begin{equation*}
R_i:=\left\{\widetilde{\mathcal X}_i
\left[\sum_{h=0}^{N_i(t)-1}(I-\eta K_i)^h-N_i(t)I\right]\tfrac12\Pi y_i\right\}S^{-1/2}.
\end{equation*}
Thus $Y_t^{[0]}-G_{[t]}=\eta\sum_{N_i(t)\ge2}R_i$.
On independent per-pair stability truncations, the coefficient matrices in this formula are bounded.
Gaussian H\"older bounds, uniformly for deterministic $w,K$, are
\begin{equation*}
\E[y_\beta^2\|X_\alpha w\|^2]\le Cqr\|w\|^2,\quad
\E[y_\beta^2\|X_\alpha^\top w\|^2]\le Cr^2\|w\|^2,\quad
\|y_\beta\|_2\|\langle X_\alpha,K\rangle_F\|_2\le C\sqrt r\|K\|_F.
\end{equation*}
They imply $\|\E R_i\|_F\le C\sqrt r$, $\E R_i^\top R_i\preceq CqrI$,
$\E R_iR_i^\top\preceq Cr^2I$, and $\E\|R_i\|_F^4\le Cq^2r^4$.
There are at most $Cqrs_t^2$ repeated pairs on the window.
The two variance traces, individual fourth moments, and mean in Lemma~\ref{er:polynomial} therefore give
\begin{equation*}
\begin{aligned}
\E\|Y_t^{[0]}-G_{[t]}\|_{S_4}^4
&\stackrel{(i)}{\le}C\eta^4\{(qr)^2r(qrs_t^2)^2+r^4q(qrs_t^2)^2+q^2r^4(qrs_t^2)\}
+C\{\eta\sqrt r(qrs_t^2)\}^4\\
&\stackrel{(ii)}{\le}C(rs_t^4+rs_t^2/q+r^2s_t^8),
\end{aligned}
\end{equation*}
where $\|A\|_{S_4}^4:=\tr((A^\top A)^2)$ and (ii) uses $\eta qr\asymp1$.

\textit{Single distinct-pair interaction.}
For one distinct-pair interaction, let $Y_{v-1}^{[0],\ne i_v}$ omit contributions from pair $i_v$.
Keeping every repeated update exactly gives
\begin{equation*}
Y_t^{[1]}=-\eta\sum_v\mathcal X_{i_v}(I-\eta K_{i_v})^{N_{i_v}(t)-N_{i_v}(v)}
\mathcal X_{i_v}^*(Y_{v-1}^{[0],\ne i_v}S).
\end{equation*}
Randomly split the pairs into two sets, each with probability $1/2$.
Retain feedback pairs in the first set and forcing pairs in the second.
Every ordered distinct pair survives with probability $1/4$, so Jensen costs at most $4^4$ in its fourth Schatten moment.
Conditional on the second set, aggregate at most $N_{\max}$ visits for each feedback pair. These aggregates are independent.
For a deterministic $Y$, the matrix $\mathcal X_i(I-\eta K_i)^h\mathcal X_i^*(YS)$ has mean norm at most $C\|Y\|_F$,
right covariance at most $Cq\|Y\|_F^2I_r$, left covariance at most $Cr\|Y\|_F^2I_{k+p+q}$,
and fourth Frobenius moment at most $Cq^2r^2\|Y\|_F^4$.
These follow by H\"older from
\begin{equation*}
\|\langle X_\beta,YS\rangle_F\|_{L^p}\le C_p\|Y\|_F,\quad
\|X_\alpha w\|^2\le2\|x_\alpha\|^2(a_\alpha^\top w)^2+2\|w\|^2,\quad
\|X_\alpha^\top w\|^2\le2(x_\alpha^\top w)^2\|a_\alpha\|^2+2\|w\|^2.
\end{equation*}
For the response restricted to the forcing set, the independent-sum fourth moment is bounded by
$C\{t^2\eta^4q^2r^4+t\eta^4q^2r^4+r^2s_t^4\}\le Cr^2s_t^2$.
Applying Lemma~\ref{er:polynomial} conditionally and then averaging this set bounds the right/left variance traces,
individual fourth moments, and conditional mean by
\begin{equation*}
Crs_t^4,\qquad C(r^2/q)s_t^4,\qquad C(r/q)s_t^3,\qquad Cr^2s_t^6,
\end{equation*}
respectively. Hence
\begin{equation*}
\E\|Y_t^{[1]}\|_{S_4}^4\le C(rs_t^4+r^2s_t^6+rs_t^3/q).
\end{equation*}
Lemma~\ref{er:polynomial} transfers these fixed-degree Gaussian moment estimates from the pairwise stability truncations.

\textit{Combining the corrections.}
Finally $\Delta_t=(Y_t^{[0]}-G_{[t]})+Y_t^{[1]}+Y_t^{[\ge2]}$.
Since $qs_t^2\ge cC_0^2\Lambda_r^{2J}$, $s_t\le C/r$, and Lemma~\ref{er:kernel} bounds the last summand,
\begin{equation*}
\begin{aligned}
\E[\mathbf1_{\mathcal E_\star}\|\Delta_t\|_{S_4}^4]
&\stackrel{(i)}{\le}C\E[\mathbf1_{\mathcal E_\star}
\{\|Y_t^{[0]}-G_{[t]}\|_{S_4}^4+\|Y_t^{[1]}\|_{S_4}^4+\|Y_t^{[\ge2]}\|_F^4\}]
\le\mathcal L_r rs_t^4,\\
\E[\mathbf1_{\mathcal E_\star}\|\Delta_t\|^4]&\le\mathcal L_r rs_t^4,\qquad
\E[\mathbf1_{\mathcal E_\star}\|\Delta_t\|_F^4]
\stackrel{(ii)}{\le}r\E[\mathbf1_{\mathcal E_\star}\|\Delta_t\|_{S_4}^4]\le\mathcal L_r r^2s_t^4.
\end{aligned}
\end{equation*}
Here (i) is the fourth-power triangle inequality and (ii) is Cauchy--Schwarz on the at most $r$ singular values.
On $\mathcal E_\star$ the finite formulas for the first two interaction terms and the all-history bound for $Y_t$ grow polynomially.
Lemma~\ref{lem:tail-absorption} therefore absorbs count/kernel failures at any required polynomial order.
\end{proof}

The zero-interaction Gram is the reference term for the trajectory variance.
\begin{lemma}[Gram of the zero-interaction response]\label{er:zero-gram}
Under the standing hypotheses and common window of Lemma~\ref{er:state}, the localized moment estimate is, for each $t_0\le t\le N+1$,
\begin{equation*}
\frac{\E[\mathbf1_{\mathcal E_\star}\|Y_t^{[0]\top}Y_t^{[0]}-s_t\Gamma^{(0)}\|_F^2]}{rs_t^2}
\le C\mathcal L_r(s_t^2+r/t+r/q).
\end{equation*}
The expectation averages the Gaussian training sample and sampling history under the initialization convention above.
Moreover, $\|Y_t^{[0]}\|\le C\sqrt{s_t}$ with probability at least $1-C_M(qr)^{-M}$ for every prescribed fixed $M>0$.
\end{lemma}
\begin{proof}

\textit{Nuisance Gram replacement.}
Fix an admissible index history. Replace each same-pair nuisance Gram by $qI_2$ in the repeated-update products,
retaining the exact signal Gram, and denote the resulting response by $\widehat Y_t^{[0]}$.
Nuisance concentration, the bounded repeated-update inverse, and Gaussian fourth moments give
\begin{equation*}
\E(\|Y_t^{[0]}\|_F+\|\widehat Y_t^{[0]}\|_F)^4\le C(rs_t)^2,\qquad
\E\|Y_t^{[0]}-\widehat Y_t^{[0]}\|_F^4\le\mathcal L_r(rs_t/q)^2.
\end{equation*}
Thus Cauchy--Schwarz applied to
$\|Y^\top Y-\widehat Y^\top\widehat Y\|_F\le(\|Y\|_F+\|\widehat Y\|_F)\|Y-\widehat Y\|_F$
bounds the normalized squared Gram replacement error by $\mathcal L_r r/q$.

\textit{Conditional variances.}
Conditional on signal coordinates and counts, the modified nuisance rows are independent Gaussian vectors.
Write $C_b:=q\operatorname{Cov}(\text{one nuisance row}\mid\text{signals, counts})/s_t$, with $\|C_b\|\le C$. Then Gaussian pairing gives the normalized variance
$\{(\tr C_b)^2+\tr(C_b^2)\}/(qr)\le Cr/q$.
Each stored pair contributes a covariance with mean norm $Cr$ and squared Frobenius moment $Cr^4$.
Independence across pairs gives $Cr/t$ after normalization. Moreover,
\begin{equation*}
\E\#\{N_i\ge2\}\le Ct^2/n,\qquad
\E\{\#\{N_i\ge2\}\}^2\le C(t^2/n+t^4/n^2).
\end{equation*}
The repeated covariance correction therefore has mean norm $Ct/n=O(s_t)$ and normalized variance $Cr/n$.
Using the per-pair corrections from Lemma~\ref{er:feedback},
\begin{equation*}
\|\E Y_t^{[0]}\|
\stackrel{(i)}{\le}s_t\|G^{\mathrm{pop}}\|+\eta\sum_{N_i(t)\ge2}\|\E R_i\|
\le Cs_t+C\sqrt r\,s_t^2\stackrel{(ii)}{\le}Cs_t.
\end{equation*}
Here (i) uses $\E Y_t^{[0]}=s_tG^{\mathrm{pop}}+\eta\sum_{N_i(t)\ge2}\E R_i$ and (ii) uses $s_t\le C/r$.
The independent-pair fourth-trace bound also gives
$\E\|Y_{t,\mathrm{sig}}^{[0]}-\E Y_{t,\mathrm{sig}}^{[0]}\|_F^4\le Cr^4s_t^2/q^2$.
Expanding the signal Gram into mean, cross, and centered-square terms contributes
$C(s_t^2+rs_t/q+r^3/q^2)$ to the normalized squared moment. Consequently
\begin{equation*}
\frac{\E[\mathbf1_{\mathcal E_\star}\|Y_t^{[0]\top}Y_t^{[0]}-s_t\Gamma^{(0)}\|_F^2]}{rs_t^2}
\le\mathcal L_r(s_t^2+r/t+r/q+rs_t/q+r^3/q^2)
\le C\mathcal L_r(s_t^2+r/t+r/q).
\end{equation*}

\textit{Operator-norm bound.}
Matrix Chernoff for the independent covariance contributions (means bounded by $CrI$, truncated norms $Cr^2\mathcal L_r$)
and the Gaussian row bound give $\|Y_t^{[0]}\|\le C\sqrt{s_t}$ because $t\gg r^2\mathcal L_r$.
Lemma~\ref{er:polynomial} justifies the localized Gaussian moments, and Lemma~\ref{lem:tail-absorption} removes exceptional histories.
\end{proof}

The first interaction requires a sharper estimate than a product of separate norm bounds.
The letters $k,i,j,p,s,v$ in the following proof are local summation indices.
\begin{lemma}[Cross Gram by Gaussian graph counting]\label{er:cross-gram}
Under the conditions of Lemma~\ref{er:zero-gram}, for every $t_0\le t\le N+1$,
\begin{equation*}
\frac{\E[\mathbf1_{\mathcal E_\star}\|Y_t^{[0]\top}Y_t^{[1]}\|_F^2]}{rs_t^2}
\le\mathcal L_r(s_t^2+r/q).
\end{equation*}
\end{lemma}
\begin{proof}

\textit{Surviving partitions.}
In each term of the finite position sum write $a_i:=Q^\top x_{i,\alpha}$, $z_i:=z_{i,\alpha}$ and $y_i:=y_{i,\alpha}$,
retaining each occurrence's position in its covariance.
The elementary forcing and feedback terms $\eta z_ka_k^\top y_k$ and
$-\eta^2z_ia_i^\top(z_i^\top z_j)(a_i^\top Sa_j)y_j$, $i\ne j$, give squared cross-Gram monomials
\begin{equation*}
\eta^6(z_k^\top z_i)(z_p^\top z_s)(z_i^\top z_j)(z_s^\top z_v)
(a_i^\top Sa_j)(a_s^\top Sa_v)(a_i^\top a_s)(a_p^\top a_k)y_ky_jy_py_v.
\end{equation*}
For a partition $\mathfrak p$ of these six pair labels, the nuisance and signal quotient multigraphs have edge multisets
$\{ki,ps,ij,sv\}$ and $\{ij,sv,is,pk\}$. Write $c_{\mathrm{nuis}},c_{\mathrm{sig}}$ for their component counts.
Gaussian pairing leaves precisely those partitions with $i\not\sim j$, $s\not\sim v$, and even nuisance degrees.
The four rows list all surviving partitions, with bars separating blocks. The final column bounds each class's contribution after division by $rs_t^2$:
\begin{equation*}
\text{normalized contribution}\ \lesssim
\frac{\eta^6t^{|\mathfrak p|}q^{v_{\mathrm{nuis}}}r^{v_{\mathrm{sig}}}}{rs_t^2}.
\end{equation*}
\begingroup
\renewcommand{\arraystretch}{1.25}
\setlength{\tabcolsep}{3pt}
\begin{center}
\begin{tabular}{@{}c c p{0.36\linewidth} c c c c@{}}
\toprule
\shortstack{Class\\$(c_{\mathrm{nuis}},|\mathfrak p|)$}&Count&Surviving partitions $\mathfrak p$
&$c_{\mathrm{sig}}\le$&$v_{\mathrm{nuis}}\le$&$v_{\mathrm{sig}}\le$&Bound\\
\midrule
$(2,4)$&1&$\{kj\mid i\mid pv\mid s\}$&1&2&3&$s_t^2$\\[2pt]
$(1,4)$&2&$\{kp\mid i\mid jv\mid s\}$, $\{kv\mid i\mid jp\mid s\}$&2&1&4&$(r/q)s_t^2$\\[2pt]
$(1,3)$&10&$\{kip\mid jv\mid s\}$, $\{kiv\mid jp\mid s\}$,\newline
$\{kj\mid ipv\mid s\}$, $\{kj\mid is\mid pv\}$,\newline
$\{kjpv\mid i\mid s\}$, $\{kjs\mid i\mid pv\}$,\newline
$\{kp\mid is\mid jv\}$, $\{kps\mid i\mid jv\}$,\newline
$\{kv\mid i\mid jps\}$, $\{kv\mid is\mid jp\}$
&2&2&5&$(r^2/t)s_t^2$\\[2pt]
$(1,2)$&4&$\{kips\mid jv\}$, $\{kiv\mid jps\}$,\newline
$\{kjpv\mid is\}$, $\{kjs\mid ipv\}$&2&3&6&$r/q$\\
\bottomrule
\end{tabular}
\end{center}
\endgroup
The cycle columns give the conservative upper bounds derived below.

\textit{Cycle-space power counting.}
There are $17$ partitions, and $c_{\mathrm{sig}}+c_{\mathrm{nuis}}\le3$ in every row.
Let $v_{\mathrm{sig}},v_{\mathrm{nuis}}$ count coordinate cycles of a fixed pairing.
Centering a quadratic label excludes pairing its two legs together, but permits cycles with one label and ordinary feature edges.
Erase label edges from each cycle and contract the resulting gaps within pairs.
Every cycle containing ordinary edges becomes a nonempty closed trail in the quotient graph.
Its nonzero edge-indicator vector lies in the cycle space over $\mathbb F_2$.
The vectors of different cycles have disjoint supports and hence are linearly independent.
The cycle-space dimension is $4-|\mathfrak p|+c_{\mathrm{sig}}$.
Label-only cycles use at least two of the four labels, so
\begin{equation*}
v_{\mathrm{sig}}\le4-|\mathfrak p|+c_{\mathrm{sig}}+2\le9-|\mathfrak p|-c_{\mathrm{nuis}},
\qquad v_{\mathrm{nuis}}\le4-|\mathfrak p|+c_{\mathrm{nuis}}.
\end{equation*}
Each nuisance loop costs $Cq$, and each signal loop costs $Cr$ since $k+p=O(r)$.
The four classes $(c_{\mathrm{nuis}},|\mathfrak p|)=(2,4),(1,4),(1,3),(1,2)$ have multiplicities $1,2,10,4$.
Their contributions therefore satisfy
\begin{equation*}
\begin{aligned}
\frac{\E\|Y_{t,\mathrm{nuis}}^{[0]\top}Y_{t,\mathrm{nuis}}^{[1]}\|_F^2}{rs_t^2}
&\le \frac{C\eta^6}{rs_t^2}
\{t^4q^2r^3+t^4qr^4+t^3q^2r^5+t^2q^3r^6\}\\
&\le C\{s_t^2+(r/q)s_t^2+(r^2/t)s_t^2+r/q\}
\le C(s_t^2+r/q).
\end{aligned}
\end{equation*}
The last line uses $\eta qr\asymp1$ and $t\gg r^2$.
Correlated two-position Gaussians are deterministic linear transforms of independent Gaussians.
Expanding the finite position sums and dropping the equal-index constraints in Isserlis pairings changes the absolute bound only by a fixed factor.

\textit{Repeated updates and signal blocks.}
Repeated-update powers insert at most $C_{N_{\max}}$ loops.
Each nuisance Gram insertion costs $C\eta qr=O(1)$ and adds one loop in each graph,
while each signal Gram insertion costs $C\eta r^2=O(r/q)$ and adds at most two signal loops.
Thus these powers multiply the bound by at most $C_{N_{\max}}$.
Constant terms from centered features delete edges and cannot increase cycle-space dimension. The four labels are unchanged.
The signal part of $\eta\mathcal T_t^{\mathrm{cross}}$ has norm $Cs_t\log(2t)$, giving normalized squared cross-Gram moment at most $\mathcal L_rs_t^2$.
Terms with signal output cost at most $\mathcal L_r(s_t^2+r/q)$ by the centered signal fourth moment used above.
Finally Lemmas~\ref{er:polynomial} and~\ref{lem:tail-absorption} justify event restriction after the Gaussian calculation.
\end{proof}

The interaction bounds control both the Gram variance and its spectrum on the common window.
\begin{lemma}[Gram variance and uniform eigenvalue bounds]\label{er:state}\label{er:eigenvalue-bounds}
For every $t_0\le t\le N+1$,
\begin{equation*}
\frac1r\E[\mathbf1_{\mathcal E_\star}\|\Gamma_t-\bar\Gamma_t\|_F^2]
\le\mathcal L_r(s_t^2+r/t+r/q).
\end{equation*}
For any prescribed fixed $M$, with probability at least $1-C_M(qr)^{-M}$,
\begin{equation*}
cI\preceq Y_t^\top Y_t/s_t\preceq CI,\qquad
cI\preceq G_{[t]}^\top G_{[t]}/s_t\preceq CI
\end{equation*}
simultaneously for $t_0\le t\le N+1$.
\end{lemma}
\begin{proof}
\textit{Gram variance.}
Lemmas~\ref{er:zero-gram} and~\ref{er:cross-gram} bound the zero-interaction Gram and its first cross term.
The remaining products satisfy
\begin{equation*}
\begin{aligned}
\frac{\E\|Y_t^{[1]\top}Y_t^{[1]}\|_F^2}{rs_t^2}
&\stackrel{(i)}{\le}C(s_t^2+rs_t^4+s_t/q)\le C\mathcal L_rs_t^2,\\
\frac{\|Y_t^{[0]\top}Y_t^{[\ge2]}\|_F^2}{rs_t^2}
&\stackrel{(ii)}{\le}\frac{Cs_t\mathcal L_rrs_t^3}{rs_t^2}=C\mathcal L_rs_t^2.
\end{aligned}
\end{equation*}
Here (i) uses Lemma~\ref{er:feedback}, while (ii) uses the zero-interaction operator norm and Lemma~\ref{er:kernel}.
Every other product has at least three interactions and contributes at most
$\mathcal L_rrs_t^3\le C\mathcal L_rs_t^2$ by the same kernel expansion.
On each fixed realization in $\mathcal E_\star$,
\begin{equation*}
\E_{\mathrm{hist}}\|\Gamma_t-\Gamma^{(0)}\|_F^2
=\E_{\mathrm{hist}}\|\Gamma_t-\bar\Gamma_t\|_F^2
+\|\bar\Gamma_t-\Gamma^{(0)}\|_F^2.
\end{equation*}
Average this identity after multiplying by $\mathbf1_{\mathcal E_\star}$.
Lemma~\ref{lem:tail-absorption} and the polynomial state bounds remove count/kernel failures, proving the variance claim.

\textit{Uniform eigenvalue bounds.}
Conditional on signal coordinates and counts, the nuisance rows of $G_{[t]}$ are independent Gaussian vectors of covariance
\begin{equation*}
\frac{\eta^2}{4}\sum_iN_i(t)^2\sum_{\alpha\in\{+,-\}}(\Pi y_i)_\alpha^2a_{i,\alpha}a_{i,\alpha}^\top.
\end{equation*}
The independent per-pair contributions have means between $crI$ and $CrI$ and truncated norms $Cr^2\mathcal L_r$.
Counts are bounded and have $o(t)$ repeats. Matrix Chernoff and Gaussian singular-value bounds therefore put this covariance between
$cs_tI/q$ and $Cs_tI/q$, and put $G_{[t],\mathrm{nuis}}^\top G_{[t],\mathrm{nuis}}/s_t$ between fixed positive multiples of $I$.
The signal mean has norm at most $Cs_t$, and hypercontractivity gives
\begin{equation*}
\Prob\{\|G_{[t],\mathrm{sig}}-\E G_{[t],\mathrm{sig}}\|>\epsilon\sqrt{s_t}\}
\le C_{p,\epsilon}(r^2/q)^p.
\end{equation*}
For each fixed admissible history, Lemmas~\ref{er:polynomial} and~\ref{er:feedback} also give
\begin{equation*}
\begin{aligned}
\E\left[\left(\frac{\|Y_t^{[0]}-G_{[t]}\|_{S_4}^4+\|Y_t^{[1]}\|_{S_4}^4}{s_t^2}\right)^p\right]
&\stackrel{(i)}{\le}C_p\left[\frac{\E\{\|Y_t^{[0]}-G_{[t]}\|_{S_4}^4+\|Y_t^{[1]}\|_{S_4}^4\}}{s_t^2}\right]^p\\
&\le C_p\mathcal L_r^pr^{-p}.
\end{aligned}
\end{equation*}
In (i), the bracket is a nonnegative Gaussian polynomial of fixed degree.
Markov makes both correction operator norms at most $r^{-1/8}\sqrt{s_t}$ with failure $C_p\mathcal L_r^pr^{-p/2}$.
The remaining tail is at most $\mathcal L_r\sqrt r\,s_t\sqrt{s_t}=o(\sqrt{s_t})$.
Singular-value perturbation transfers the two bounds for $G_{[t]}$ to $Y_t$.
Finally choose $p$ and the count/kernel confidence exponents large enough to take a union over the polynomially many prefixes and all four runs.
\end{proof}

\subsubsection{From mean drift to the expected effective-rank increment}

Excluding the currently sampled pair separates its forcing from the independent past contribution.
\begin{lemma}[Forcing moments after excluding the currently sampled pair]\label{er:leave-one}
For $t\le n$, $s=t\eta\le1$, the accumulated forcing $G_{[t]} =\eta\sum_{v<t}G_{I_v}$ under repeated sampling from the training set satisfies
\begin{equation*}
\begin{aligned}
\E\E_I\|\mathcal{H}_IG_{[t]} \|_F^4&\le Cq^2r^4(s^2+s^4+t/n),\\
\E\E_I\|G_{[t]} ^\top \mathcal{H}_IG_{[t]} \|_F^2&\le C(r^3s^2+r^2s).
\end{aligned}
\end{equation*}
\end{lemma}
\begin{proof}

\textit{Leave-one-out decomposition.}
Fix the selected pair $i$ and condition on its visit counts.
The decomposition
\begin{equation}\label{er:forcing-decomposition}
\begin{aligned}
G_{[t]}&=\mathring G_{[t]}^{\setminus i}+s_t^{\setminus i}G^{\mathrm{pop}}+\eta N_iG_i,\\
\mathring G_{[t]}^{\setminus i}&:=\eta\sum_{i'\ne i}N_{i'}(G_{i'}-G^{\mathrm{pop}}),
\qquad s_t^{\setminus i}:=\eta(t-N_i)
\end{aligned}
\end{equation}
separates a centered sum independent of pair $i$.
Gaussian quadratic-form moments give $\E|\langle K,G-G^{\mathrm{pop}}\rangle_F|^{2p}\le C_pr^p\|K\|_F^{2p}$ for $p=1,2$.
The independent fourth-moment expansion yields
\begin{equation*}
\begin{aligned}
\E|\langle X_\beta,\mathring G_{[t]}^{\setminus i}S\rangle_F|^4
&\le Cr^2\eta^4\left(\sum_{i'\ne i}N_{i'}^2\right)^2\|X_\beta\|_F^4,\\
\E\|(\mathring G_{[t]}^{\setminus i})^\top X_\alpha\|_F^4
&\le Cr^4\eta^4\left(\sum_{i'\ne i}N_{i'}^2\right)^2\|X_\alpha\|_F^4.
\end{aligned}
\end{equation*}
For the second estimate, the $r^2$ scalar coefficient matrices have total squared Frobenius norm $r\|X_\alpha\|_F^2$.

The six within-pair estimates used when substituting \eqref{er:forcing-decomposition} are
\begin{equation*}
\begin{aligned}
\E\|X_\alpha\|_F^8&\le Cq^4r^4,&
\E|\langle X_\beta,GS\rangle_F|^4+\E\|G^\top X_\alpha\|_F^4&\le Cq^4r^6,\\
\E[\|X_\alpha\|_F^2|\langle X_\beta,GS\rangle_F|^2]
+\E[\|G^\top X_\alpha\|_F^2\|X_\beta\|_F^2]&\le Cq^3r^4,&
\E[\|G^\top X_\alpha\|_F^2|\langle X_\beta,GS\rangle_F|^2]&\le Cq^4r^6,\\
\E\|(G^{\mathrm{pop}})^\top X_\alpha\|_F^4&\le Cr^4,&
\E|\langle X_\beta,G^{\mathrm{pop}}S\rangle_F|^4&\le Cr^2.
\end{aligned}
\end{equation*}
Indeed, substitute $G=\frac12\sum\Pi_{\alpha\beta}y_\beta X_\alpha$ and apply H\"older with
\begin{equation*}
\|\|X_\alpha\|_F\|_{L^p}\le C_p\sqrt{qr},\quad
\|y_\beta\|_{L^p}\le C_p\sqrt r,\quad
\|\|X_\alpha^\top X_\beta\|_F\|_{L^p}
+\|\langle X_\alpha,X_\beta S\rangle_F\|_{L^p}\le C_pqr.
\end{equation*}
The last two estimates use the signal support and bounded operator norm of $G^{\mathrm{pop}}$.

\textit{Visit-count averaging.}
For falling factorials, multinomial sampling gives
\begin{equation*}
\begin{gathered}
\E\prod_{i'}(N_{i'})_{h(i')}=\frac{(t)_{\sum h(i')}}{n^{\sum h(i')}},\qquad
\E\left(\sum_{i'\ne i}N_{i'}^2\right)^2\le Ct^2,\\
\E\left[N_i^2\sum_{i'\ne i}N_{i'}^2\right]\le C(t^4/n^3+t^3/n^2+t^2/n)\le Ct^2/n,\qquad
\E N_i^4\le Ct/n.
\end{gathered}
\end{equation*}
Thus the three summands in \eqref{er:forcing-decomposition}, after applying $\mathcal H_i$, contribute
$Cq^2r^4s^2$, $Cq^2r^4s^4$, and $Cq^2r^4t/n$ to the fourth moment.
Expanding both factors of $G_{[t]}^\top\mathcal H_iG_{[t]}$ bounds its centered quadratic, mixed,
and repeated-pair quadratic terms by $Cr^3s^2$, $C(r^3+r^2)s^2(1+t/n)$, and $Cr^2t/n$.
Since $t/n\asymp s\le1$, these give both claims.
\end{proof}

For a fixed time $t$, define the event
\begin{equation*}
\mathcal B_t:=\mathcal E_\star \cap\{cI\preceq\Gamma_t\preceq CI,\quad cI\preceq G_{[t]} ^\top G_{[t]} /s_t\preceq CI\},
\end{equation*}
with the constants from Lemma~\ref{er:state}.
This event is measurable before drawing the next index.

The update-direction fourth moment makes the entropy Taylor remainder smaller than the signed first-order increment.
\begin{lemma}[Fourth moment of the update direction and effective-rank Taylor remainder]\label{er:one-step}
Let $D_t:=G_I-\mathcal{H}_IY_t$.
For every $t_0\le t<N$,
\begin{equation*}
\E[\mathbf1_{\mathcal E_\star }\E_I\|D_t \|_F^4]\le \mathcal L_rq^2r^4(1+qr s_t^4)
\end{equation*}
and
\begin{equation*}
\frac1{r\eta}\E[\mathbf1_{\mathcal B_t}\E_I\|\Gamma_{t+1}-\Gamma_t\|_F^2]
\le \mathcal L_r\left\{\frac{qr^2}{t^2}+rs_t^2+\frac{r^2}t+\frac rq+\frac rt+\sqrt r\,s_t+s_t\right\}=o(1).
\end{equation*}
The $o(1)$ is uniform on this interval whenever the fixed exponent of ${\Lambda_r^J}$ is sufficiently large.
\end{lemma}
\begin{proof}

The update direction decomposes as
\begin{equation}\label{er:update-decomposition}
D_t=G_I-\mathcal H_IG_{[t]}-\mathcal H_I\Delta_t.
\end{equation}
Gaussian moments and Lemma~\ref{er:leave-one} bound the first two fourth moments. Lemmas~\ref{er:training-set} and~\ref{er:feedback} control the third through
\begin{equation*}
\E_I\|\mathcal H_I\Delta_t\|_F^4
\le\max_I\|\mathcal H_I\|^2\|\Delta_t\|_F^2\E_I\|\mathcal H_I\Delta_t\|_F^2
\le C(qr)^3\mathcal L_r\|\Delta_t\|_F^4.
\end{equation*}
The fourth-power triangle inequality proves the stated bound for $D_t$.

On $\mathcal B_t$, write $Y:=Y_t$. Both $Y,G_{[t]}$ have operator norm $C\sqrt{s_t}$ and Frobenius norm $C\sqrt{rs_t}$.
Then
\begin{equation}\label{er:hessian-decomposition}
\begin{aligned}
Y^\top\mathcal H_IY
&=G_{[t]}^\top\mathcal H_IG_{[t]}+\Delta_t^\top\mathcal H_IG_{[t]}+Y^\top\mathcal H_I\Delta_t,\\
\E_I\|\Delta_t^\top\mathcal H_IG_{[t]}\|_F^2
&\le\|\Delta_t\|^2\E_I\|\mathcal H_IG_{[t]}\|_F^2
\le Cqr^2\mathcal L_rs_t\|\Delta_t\|^2,\\
\E_I\|Y^\top\mathcal H_I\Delta_t\|_F^2
&\le\|Y\|^2\E_I\|\mathcal H_I\Delta_t\|_F^2
\le Cqr\mathcal L_rs_t\|\Delta_t\|_F^2.
\end{aligned}
\end{equation}
After averaging, Lemma~\ref{er:feedback} bounds the last two lines by
$\mathcal L_rqr^{5/2}s_t^3$ and $\mathcal L_rqr^2s_t^3$. Lemma~\ref{er:leave-one} bounds the first term.
The empirical forcing covariance gives
$\E_I\|Y^\top G_I\|_F^2\le\|\E_I G_IG_I^\top\|\|Y\|_F^2\le Cr^3\mathcal L_rs_t$.
Insert these estimates into the exact increment
\begin{equation}\label{er:gram-increment}
\Gamma_{t+1}-\Gamma_t=\frac\eta{s_t+\eta}(Y_t^\top D_t+D_t^\top Y_t-\Gamma_t)
+\frac{\eta^2}{s_t+\eta}D_t^\top D_t.
\end{equation}
The squared-norm triangle inequality and the fourth moment of $D_t$ give
$\mathcal L_r(qr^2/t^2+rs_t^2)$ from its final term and the remaining displayed terms from the forcing/Hessian bounds.

Uniform vanishing follows from
\begin{equation*}
qr^2/t^2\le(C_0\Lambda_r^J)^{-2},\qquad
r^2/t\le r/(C_0\Lambda_r^J\sqrt q),\qquad
rs_t^2\le C/r,\qquad \sqrt r\,s_t\le C/\sqrt r,
\end{equation*}
with $q=r^{a_q}$, $a_q>2$, and $J$ sufficiently large.
Lemma~\ref{er:calculus} bounds the entropy Taylor remainder by $C\|\Gamma_{t+1}-\Gamma_t\|_F^2/r=o(\eta)$.
\end{proof}

The final comparison converts the signed drift of the mean Gram matrix into an unconditional expected increment.
\begin{lemma}[Expected increment compared with the derivative at the conditional mean]\label{er:compensation}
Uniformly for $t_0\le t<N$,
\begin{equation*}
\E[\operatorname{er}_T(\Gamma_{t+1})-\operatorname{er}_T(\Gamma_t)]
=\E[\mathbf1_{\mathcal E_\star }D\operatorname{er}_T(\bar \Gamma_t)[\bar \Gamma_{t+1}-\bar \Gamma_t]]+o(\eta).
\end{equation*}
\end{lemma}
\begin{proof}

\textit{Drift decomposition.}
Put $s:=s_t$. Expand the conditional Gram update as
\begin{equation}\label{er:drift-decomposition}
\begin{aligned}
R_{\mathrm{drift}}(Y)&:=\sum_{h=1}^5T_h(Y),\\
T_1(Y)&:=\frac\eta{s+\eta}(\Gamma^{(0)}-\bar\Gamma_t),\\
T_2(Y)&:=\frac\eta{s+\eta}(Y^\top\bar G+\bar G^\top Y),\\
T_3(Y)&:=-\frac\eta{s+\eta}\{Y^\top\overline{\mathcal H}Y+(\overline{\mathcal H}Y)^\top Y\},\\
T_4(Y)&:=-\frac{\eta^2}{s+\eta}\E_I\{G_I^\top\mathcal H_IY+(\mathcal H_IY)^\top G_I\},\\
T_5(Y)&:=\frac{\eta^2}{s+\eta}\E_I(\mathcal H_IY)^\top\mathcal H_IY.
\end{aligned}
\end{equation}
Then \eqref{er:gram-increment} gives
$\E_I(\Gamma_{t+1}-\Gamma_t)=R_{\mathrm{drift}}(Y_t)-\eta(\Gamma_t-\bar\Gamma_t)/(s+\eta)$.
The derivative Hessian bound in Lemma~\ref{er:calculus} yields, on $\mathcal B_t$,
\begin{equation*}
\begin{aligned}
&\left|\{D\operatorname{er}_T(\Gamma_t)-D\operatorname{er}_T(\bar\Gamma_t)\}
[\E_I(\Gamma_{t+1}-\Gamma_t)]\right|\\
&\qquad\le\frac{C\eta}{rs}\|\Gamma_t-\bar\Gamma_t\|_F^2
+\frac Cr\|\Gamma_t-\bar\Gamma_t\|_F\|R_{\mathrm{drift}}(Y_t)\|_F.
\end{aligned}
\end{equation*}
Both reference and current Grams have fixed spectral bounds on this window by Lemmas~\ref{er:affine} and~\ref{er:state}.

\textit{Bounding $T_1$--$T_5$.}
The five terms are bounded together as follows, with $\|Y\|^2\le Cs$ and $\|Y\|_F^2\le Crs$:
\begin{equation*}
\begin{aligned}
\|T_1(Y)\|_F^2/r&\le\frac{C\eta^2}{rs^2}\|\Gamma^{(0)}-\bar\Gamma_t\|_F^2
\stackrel{(i)}{\le}C\eta^2r,\\
\|T_2(Y)\|_F^2/r&\le\frac{C\eta^2}{rs^2}\|Y\|^2\|\bar G\|_F^2
\stackrel{(ii)}{\le}C\eta^2/s,\\
\|T_3(Y)\|_F^2/r&\le\frac{C\eta^2}{rs^2}\|Y\|^2\|\overline{\mathcal H}Y\|_F^2
\stackrel{(ii)}{\le}C\eta^2,\\
\|T_4(Y)\|_F^2/r&\le\frac{C\eta^4}{rs^2}(qr\sqrt{\Lambda_r}\|Y\|_F)^2
\stackrel{(iii)}{\le}C\eta^2\Lambda_r/s,\\
\|T_5(Y)\|_F^2/r&\le\frac{C\eta^4}{rs^2}(qr\mathcal L_r\|Y\|_F^2)^2
\stackrel{(ii)}{\le}C\eta^2r\mathcal L_r^2,\\
\|R_{\mathrm{drift}}(Y)\|_F^2/r&\le5\sum_{h=1}^5\|T_h(Y)\|_F^2/r
\le C\eta^2(\Lambda_r/s+r\mathcal L_r^2+r).
\end{aligned}
\end{equation*}
Here (i) is Lemma~\ref{er:affine}. Bound (ii) uses the mean forcing, mean Hessian, or Hessian second-moment bound in Lemma~\ref{er:training-set}.
Bound (iii) is the empirical map bound in Lemma~\ref{er:cross-map}. All use $\eta qr\asymp1$.

Average the derivative comparison with $\mathbf1_{\mathcal B_t}$.
Cauchy--Schwarz and Lemma~\ref{er:state} bound its error by
\begin{equation*}
C\eta\left\{\frac{\mathcal L_r(s^2+r/t+r/q)}s
+\sqrt{(\Lambda_r/s+r\mathcal L_r^2+r)\mathcal L_r(s^2+r/t+r/q)}\right\}=o(\eta).
\end{equation*}
The required uniform vanishing, also used for rotation, is explicit:
\begin{equation}\label{er:drift-variance-scales}
\begin{aligned}
\Lambda_r\mathcal L_r(s+r/(ts)+r/(qs))
&\le C\Lambda_r\mathcal L_r\{r^{-1}+\Lambda_r^{-2J}+r/(\sqrt q\Lambda_r^J)\}=o(1),\\
r\mathcal L_r^3(s^2+r/t+r/q)
&\le C\mathcal L_r^3\{r^{-1}+r/(\sqrt q\Lambda_r^J)+r^2/q\}=o(1).
\end{aligned}
\end{equation}
These follow from $t\ge C_0r\sqrt q\Lambda_r^J$, $s\le C/r$, and $q=r^{a_q}$, $a_q>2$, choosing $J$ large enough.

\textit{Taylor expansion and tail removal.}
Lemma~\ref{er:one-step} now justifies Taylor expansion about $\Gamma_t$ on $\mathcal B_t$.
The event $\mathcal B_t$ is measurable before the next draw, so the linear Taylor term averages to
$D\operatorname{er}_T(\Gamma_t)[\E_I(\Gamma_{t+1}-\Gamma_t)]$.
The entropy rank is bounded. On $\mathcal E_\star$ the reference derivative times the conditional increment has polynomial growth.
Lemma~\ref{lem:tail-absorption} therefore removes $\mathcal B_t^c$ from the former and $\mathcal E_\star\setminus\mathcal B_t$ from the latter at cost $o(\eta)$.
Finally $\bar\Gamma_t$ and $\mathcal E_\star$ are fixed conditional on initialization, task, and training set, while
$\E_{\mathrm{hist}}\E_I(\Gamma_{t+1}-\Gamma_t)=\bar\Gamma_{t+1}-\bar\Gamma_t$.
This proves the comparison, with the reference derivative extended by zero outside $\mathcal E_\star$.
\end{proof}

\begin{proof}[Proof of Proposition~\ref{prop:first-order-er}]
Lemma~\ref{sp:moment-rates} gives $0<u_L<\omega<u_H<2g\omega$.
Apply Lemma~\ref{er:compensation} to the signed margins in Lemma~\ref{er:mean-phase},
and choose $\kappa_{\mathrm{ER}}>0$ smaller than their minimum over the four runs.
The uniform $o(\eta_j)$ error then gives, for $t_0\le t<N$,
\begin{equation*}
\varsigma_{m,j}\E\{\operatorname{er}_T(\Gamma_{m,j,t+1})-\operatorname{er}_T(\Gamma_{m,j,t})\}
\ge\kappa_{\mathrm{ER}}\eta_j.
\end{equation*}
\end{proof}

\subsection{Proof of Proposition~\ref{rr:rotation-step}}\label{app:rotation}
Fix a rule $m$ and rate $j$, suppress their indices, and put $s:=t\eta$,
$D_t:=G_{I_t}-\mathcal H_{I_t}[Y_t]$ and $P_{Y_t}:=Y_t(Y_t^\top Y_t)^{-1}Y_t^\top$ on the full-rank event.
Write $\operatorname{rot}^0(t,t'):=1-\operatorname{ovlp}_L^0(t,t')$.
Since $(I-P_{Y_t})Y_{t+1}=\eta(I-P_{Y_t})D_t$ and $Y_{t+1}^\top Y_{t+1}=(s+\eta)\Gamma_{t+1}$,
\begin{equation}\label{rr:projector-identity}
\operatorname{rot}^0(t,t+1)=\frac{\eta^2}{r(s+\eta)}
\tr\{\Gamma_{t+1}^{-1}D_t^\top(I-P_{Y_t})D_t\}.
\end{equation}
The column spaces of $L_t^0A_0=Y_tS^{1/2}A_0$ and $Y_t$ agree on Lemma~\ref{er:init}'s event.
Expectations average all randomness unless marked otherwise. Constants follow Section~\ref{app:foundation}.

\subsubsection{Angular coefficient}
Use $M_1,M_2,M_3$, the channel coefficients $\vartheta_{m,\bullet}$, and $\hat u=\eta q\tr S$ from Section~\ref{app:er}.
Define
\begin{equation}\label{rr:angular-coefficient}
\begin{aligned}
\Gamma^{\mathrm{ang}}&:=-\bar G^\top\bar G
 +\tfrac\eta2(M_1+M_1^\top-M_2-M_2^\top)+\tfrac{\eta^2}2M_3,\\
d_{\mathrm{ang}}&:=\tr\{(\Gamma^{(0)})^{-1}\Gamma^{\mathrm{ang}}\}.
\end{aligned}
\end{equation}
The inverse is used only on $\mathcal E_\star$ and its angular traces are extended by zero outside.
Ratios with denominator $\nu_m$ are likewise zero when $\nu_m=0$. On the initialization event $\nu_m\asymp r$.

Comparing \eqref{rr:angular-coefficient} with \eqref{er:gram-drift} explains the sign reversal: coherent fitting pulls adjacent updates toward the same span and therefore contributes $-\bar G^\top\bar G$ to rotation.
The state-feedback term $M_1+M_1^\top$ instead enters $\Gamma^{\mathrm{ang}}$ with a positive sign.
The shared drift moments identify the angular coefficient and its rule-independent collision term.
\begin{lemma}[Angular coefficient under repeated sampling]\label{rr:angular-moments}
For either rule at a fixed dimensionless rate $u\in(0,2)$,
\begin{equation*}
\begin{aligned}
\frac{\E_{\mathrm{train}}d_{\mathrm{ang},m}}r
&=2\vartheta_{m,\mathrm{nuis}}\frac{\tr S}{r}
 -\frac{4\sum_{\bullet\in\{a,o\}}\vartheta_{m,\bullet}^2\tr(S\Upsilon_\bullet)}{\nu_m}
 +\vartheta_{m,\mathrm{nuis}}q\eta\tr(S^2)\\
&\quad-\frac{4\sum_{\bullet\in\{a,o\}}\vartheta_{m,\bullet}^2\tr\Upsilon_\bullet}{q\eta\nu_m r}
 +\frac{-1+\hat u^2/2}{\eta n}+o(1).
\end{aligned}
\end{equation*}
Here $\E_{\mathrm{train}}$ conditions on initialization and task.
The remainder tends to zero in $L^1$ over these remaining variables, uniformly over the fixed parameter class, including fixed $\gamma_n>0$.
\end{lemma}
\begin{proof}

On $\mathcal E_\star$, Lemmas~\ref{er:training-set} and \ref{er:affine} give bounded inverses and
\begin{equation*}
\frac{\|\Gamma^{\mathrm{ang}}\|_*}r
\le\frac{\|\Gamma^{(1)}\|_*+2\|\bar G\|_F^2+2\eta\|M_1\|_*}r\le C,
\qquad \eta\|M_1\|_*\le\eta\|\overline{\mathcal H}\|\E_I\|G_I\|_F^2\le Cr.
\end{equation*}
Thus the inverse identity bounds the error in replacing $(\Gamma^{(0)})^{-1}$ by $(q\eta\nu_m)^{-1}I$,
after normalized trace, by $C\|\Gamma^{(0)}-q\eta\nu_mI\|=o(1)$ in expectation.

Now substitute the four shared moments in \eqref{er:drift-blocks} into \eqref{rr:angular-coefficient} and divide by $q\eta\nu_m$.
Their distinct-pair terms give the first four displayed terms. Replacing $1-n^{-1}$ by $1$ costs $o(1)$.
For repeated pairs the $M_1$ and $M_2$ terms cancel exactly, leaving
$(q\nu_m/n)(-1+\hat u^2/2)I$, which gives the fifth term.
All normalized errors are $o(1)$ since $q\eta\nu_m\asymp1$.
Gaussian forcing moments and Lemma~\ref{lem:tail-absorption} remove the event restrictions.
\end{proof}

\begin{proposition}[Left singular-subspace rotation coefficient]\label{rr:rotation-coefficient}
At every fixed common dimensionless rate $u\in(0,2)$,
\begin{equation*}
\frac{\E(d_{{\mathrm{ang}},C}-d_{{\mathrm{ang}},R})}{r\mu_1}=\Psi(u;h_M)+o(1).
\end{equation*}
Moreover, $\Psi(\cdot;h_M)$ is strictly decreasing on $(0,\infty)$.
Hence $0<u_L<u_H$ and $\Psi(u_H;h_M)>0$ imply $\Psi(u_L;h_M)>\Psi(u_H;h_M)>0$.
\end{proposition}
\begin{proof}

Lemma~\ref{rr:teacher} gives the following limits, with columns corresponding to $R,C$:
\begingroup
\renewcommand{\arraystretch}{1.35}
\begin{equation*}
\begin{array}{c|cc}
\text{Quantity}&R&C\\\hline
\dfrac{\sum_\bullet\vartheta_{m,\bullet}^2\tr(S\Upsilon_\bullet)}{\nu_m}
&\dfrac{\mu_1}{2}(v_H+f)
&\dfrac{\mu_1}{2}\left(gv_H+f\dfrac{\vartheta_a^2}{\bar\vartheta_1}\right)\\[3pt]
\dfrac{\sum_\bullet\vartheta_{m,\bullet}^2\tr\Upsilon_\bullet}{\nu_m}
&\dfrac{h_M+f}{2\gamma_k}
&\dfrac{gh_M+f\vartheta_a^2/\bar\vartheta_1}{2\gamma_k}
\end{array}
\end{equation*}
\endgroup
Insert these limits and $q\eta=u/(\gamma_k\mu_{1,r}r)$ into Lemma~\ref{rr:angular-moments}.
The repeated-pair term cancels between rules at a common rate. The remaining difference is $\Psi(u;h_M)$ from Section~\ref{app:spectrum}.

Moreover $g<1$ and $f\vartheta_a^2/\bar\vartheta_1<f$, so
\begin{equation*}
\Psi'(u;h_M)=-\frac{2\{(1-g)h_M+f-f\vartheta_a^2/\bar\vartheta_1\}}{u^2}
 -\frac{v_H+\gamma_k}{2}<0.
\end{equation*}
\end{proof}

\begin{remark}[Learning-rate dependence]
At fixed adjacent update indices, the leading centered-minus-raw rotation gap is proportional to $u\Psi(u;h_M)$, because $\eta=u/({\gamma_k} \mu_{1,r}qr)$.
For $u\ge1$,
\begin{equation*}
\frac{d}{du}[u\Psi(u;h_M)]=-{\gamma_k} +2\left\{(1-g){v_H} +f-\frac{f\vartheta_a^2}{\bar\vartheta_1}\right\}-({v_H} +{\gamma_k} )u
\le(1-2g){v_H} +2f-2{\gamma_k} <0.
\end{equation*}
For the additional rate construction with $u_L+u_H=(T_R+T_C)/2\to2$ along spectra with ${v_H} \to\infty$, subtraction gives
\begin{equation*}
\begin{aligned}
&u_H\Psi(u_H;h_M)-u_L\Psi(u_L;h_M)\\
&\quad=(u_H-u_L)\left[-{\gamma_k} +2\left\{(1-g){v_H} +f-\frac{f\vartheta_a^2}{\bar\vartheta_1}\right\}-\frac{{v_H} +{\gamma_k}}{2}(u_H+u_L)\right]<0
\end{aligned}
\end{equation*}
for spectra with sufficiently large ${v_H} $ in this construction, since the bracket divided by ${v_H} $ tends to $1-2g<0$.
This comparison concerns the leading coefficients at common
update indices under the additional rate construction above.
\end{remark}

\subsubsection{Adjacent updates}
The following expansion preserves the zero-lag cancellation uniformly on the common geometry window.
\begin{lemma}[One-step overlap on the common geometry window]\label{rr:one-step-overlap}
For either rule and either common rate, if the fixed startup exponent $J$ is sufficiently large,
the first-order response satisfies uniformly for $t_0\le t<N$,
\begin{align}
&\left|\E\operatorname{rot}^0(t,t+1)-\frac1{t+1}
 -\eta\frac{t}{t+1}\frac{\E d_{\mathrm{ang}}}r\right|
\label{rr:one-step-expansion}\\
&\qquad\le C\eta(rs+t^{-1})+o(\eta).
\label{rr:one-step-error}
\end{align}
\end{lemma}
\begin{proof}
\textit{Inverse and projector replacement.}
Start on $\mathcal B_t\cap\mathcal B_{t+1}$ from Lemma~\ref{er:state}, where both Gram inverses are bounded.
The inverse identity, Cauchy--Schwarz, and Lemma~\ref{er:one-step} bound the error from replacing
$\Gamma_{t+1}^{-1}$ by $\Gamma_t^{-1}$ in \eqref{rr:projector-identity} by
\begin{equation*}
C\eta\mathcal L_r
\sqrt{\left(\frac{qr^2}{t^2}+rs^2+\frac{r^2}t+\frac rq+\frac rt+\sqrt r\,s+s\right)
       \left(\frac1{qs^2}+rs^2\right)}.
\end{equation*}
Indeed the unscaled bound is $C\eta^2(rs)^{-1}$ times the product of the localized $L^2$ norm of
$\Gamma_{t+1}-\Gamma_t$ and the localized $L^2$ norm of $\|D_t\|_F^2$.
The projected-direction estimates in the same lemma give
\begin{equation*}
\frac{C\eta^2}{rs^2}\E[\mathbf1_{\mathcal B_t}\|Y_t^\top D_t\|_F^2]
\le C\eta\mathcal L_r\left(\frac{r^2}t+\frac rq+\frac rt+\sqrt r\,s+s\right).
\end{equation*}
This removes $P_{Y_t}$. The event $\mathcal B_t$ is measurable before $I_t$ is drawn.
Removing $\mathcal B_{t+1}$ by the tail estimate below therefore leaves
$\eta\{r(s+\eta)\}^{-1}\E[\mathbf1_{\mathcal B_t}\tr\{\Gamma_t^{-1}\mathcal V(Y_t)\}]$, where
\begin{equation}\label{rr:innovation-decomposition}
\begin{aligned}
\mathcal V(Y)&:=\eta\E_I(G_I-\mathcal H_I[Y])^\top(G_I-\mathcal H_I[Y])\\
&=\Gamma^{(0)}-\eta\E_I\{G_I^\top\mathcal H_I[Y]+\mathcal H_I[Y]^\top G_I\}
 +\eta\E_I\mathcal H_I[Y]^\top\mathcal H_I[Y].
\end{aligned}
\end{equation}

\textit{Inverse compensation.}
Condition on initialization, task and training set in $\mathcal E_\star$, retaining
$\bar\Gamma_t=\E_{\mathrm{hist}}\Gamma_t$ from Section~\ref{app:er}.
The identity $\tr(\Gamma_t^{-1}\Gamma_t)=r$ on $\mathcal B_t$ gives exactly
\begin{equation}\label{rr:inverse-compensation}
\begin{aligned}
&\E_{\mathrm{hist}}[\mathbf1_{\mathcal B_t}\tr\{\Gamma_t^{-1}\mathcal V(Y_t)\}]
 -\tr\{\bar\Gamma_t^{-1}\E_{\mathrm{hist}}\mathcal V(Y_t)\}\\
&=\E_{\mathrm{hist}}[\mathbf1_{\mathcal B_t}
 \tr\{(\Gamma_t^{-1}-\bar\Gamma_t^{-1})(\mathcal V(Y_t)-\Gamma_t)\}]\\
&\quad-\E_{\mathrm{hist}}[\mathbf1_{\mathcal B_t^c}
 \{r+\tr[\bar\Gamma_t^{-1}(\mathcal V(Y_t)-\Gamma_t)]\}].
\end{aligned}
\end{equation}
Only the bounded reference inverse occurs in the last term, handled below.
For the first term return to the full average, keeping $\mathbf1_{\mathcal E_\star}$.
The linear and quadratic terms in \eqref{rr:innovation-decomposition} have Frobenius norms at most
$C\sqrt{\Lambda_r}\|Y\|_F$ and $C\mathcal L_r\|Y\|_F^2$, respectively, by Lemma~\ref{er:training-set}.
Since $\|Y_t\|_F^2\le Crs$ on $\mathcal B_t$ and $\|\bar\Gamma_t-\Gamma^{(0)}\|_*\le Crs$,
\begin{equation*}
\frac1r\E[\mathbf1_{\mathcal B_t}\|\mathcal V(Y_t)-\bar\Gamma_t\|_F^2]
\le C(\Lambda_rs+r\mathcal L_r^2s^2+rs^2).
\end{equation*}
Together with Lemma~\ref{er:state}'s variance bound, the inverse identity and Cauchy--Schwarz therefore bound
the scaled first term of \eqref{rr:inverse-compensation} by
\begin{equation*}
C\eta\mathcal L_r\left\{
\frac{s^2+r/t+r/q}{s}+\sqrt{(s^{-1}+r)(s^2+r/t+r/q)}\right\}.
\end{equation*}
Here and above fixed logarithmic powers are absorbed into $\mathcal L_r$. The drift-variance estimate \eqref{er:drift-variance-scales} in Lemma~\ref{er:compensation} makes this bound $o(\eta)$ uniformly.

\textit{Conditional mean and tail removal.}
Transport the matrix moments of Lemma~\ref{er:moment-recursion} by $Y=L^0S^{-1/2}$.
Lemma~\ref{er:init} bounds this coordinate change. Thus
\begin{equation*}
\E_{\mathrm{hist}}Y_t=s\bar G+O_F(\sqrt r\,s^2),\qquad
\E_{\mathrm{hist}}(Y_t\otimes Y_t)=s\eta\E_I(G_I\otimes G_I)+O_*(rs^2).
\end{equation*}
The map from these two moments to $\E_{\mathrm{hist}}\mathcal V(Y_t)-\Gamma^{(0)}$ is bounded
from the normalized moment norm to $r^{-1}\|\cdot\|_*$: Hessian stability and Cauchy--Schwarz give
\begin{equation*}
\eta\E_I\|\mathcal H_I[Y]\|_F^2\le C\|Y\|_F^2,\qquad
\eta\E_I\|G_I\|_F\|\mathcal H_I[Y]\|_F\le C\sqrt r\|Y\|_F.
\end{equation*}
The quadratic bound extends to signed moments by spectral decomposition.
Consequently, using Lemma~\ref{er:affine},
\begin{equation*}
\begin{aligned}
\E_{\mathrm{hist}}\mathcal V(Y_t)&=\Gamma^{(0)}
 +s\{-\eta(M_2+M_2^\top)+\eta^2M_3\}+O_*(rs^2),\\
\bar\Gamma_t&=\Gamma^{(0)}+(s-\eta)\Gamma^{(1)}+O_*(rs^2),\\
\E_{\mathrm{hist}}\mathcal V(Y_t)-\bar\Gamma_t
 &=s\Gamma^{\mathrm{ang}}+\eta\Gamma^{(1)}+O_*(rs^2).
\end{aligned}
\end{equation*}
Since $\|\bar\Gamma_t-\Gamma^{(0)}\|\le Crs$ and
$\|\Gamma^{\mathrm{ang}}\|_*+\|\Gamma^{(1)}\|_*\le Cr$, another inverse identity yields
\begin{equation*}
\frac1r\tr\{\bar\Gamma_t^{-1}\E_{\mathrm{hist}}\mathcal V(Y_t)\}
=1+s\frac{d_{\mathrm{ang}}}r+O(rs^2+\eta).
\end{equation*}
Multiplying by $\eta/(s+\eta)$ gives the two leading terms in \eqref{rr:one-step-expansion}, with error $C\eta(rs+t^{-1})$.

It remains to remove exceptional events and verify uniformity.
The all-history bounds of Lemma~\ref{er:training-set} give polynomial envelopes on $\mathcal E_\star$ for the removed direction terms
and for $r+\tr\{\bar\Gamma_t^{-1}(\mathcal V(Y_t)-\Gamma_t)\}$.
Lemma~\ref{lem:tail-absorption} and the arbitrarily high confidence in Lemma~\ref{er:state} make their exceptional expectations $o(\eta)$.
Outside $\mathcal E_\star$ use $0\le\operatorname{rot}^0\le1$ and its failure probability. Angular traces are already extended by zero.

Finally $s\asymp t/(qr)$, $t\ge C_0r\sqrt q\Lambda_r^J$, and $rs\le Cc_N$.
Equation~\eqref{er:drift-variance-scales} makes the inverse-compensation bound $o(\eta)$ uniformly.
For the first two error bounds use
\begin{equation*}
\frac1{qs^2},\frac{qr^2}{t^2}\le C\Lambda_r^{-2J},\qquad
rs^2\le C/r,\qquad \frac{r^2}t\le\frac{Cr}{\sqrt q\Lambda_r^J},\qquad
\sqrt r\,s\le C/\sqrt r.
\end{equation*}
The remaining terms are smaller. Assumption~\ref{ass:dimensions} and sufficiently large fixed $J$ make both bounds $o(\eta)$ uniformly.
Combining these estimates proves \eqref{rr:one-step-error}.
\end{proof}

\begin{proof}[Proof of Proposition~\ref{rr:rotation-step}]
Apply Lemma~\ref{rr:one-step-overlap} to the two rules at a common rate. Their $1/(t+1)$ terms cancel.
Proposition~\ref{rr:rotation-coefficient} gives, uniformly on the common window,
\begin{equation*}
\E\operatorname{rot}^0_{C,j}(t,t+1)-\E\operatorname{rot}^0_{R,j}(t,t+1)
=\eta_j\frac{t}{t+1}\{\mu_1\Psi(u_j;h_M)+o(1)\}+O(\eta_jrs)+o(\eta_j).
\end{equation*}
Here $\Psi(u_j;h_M)\ge\Psi(u_H;h_M)>0$, $t/(t+1)\ge1/2$, and $rs\le Cc_N$.
Choose $c_N$ sufficiently small and then $r$ sufficiently large.
The difference is at least $\eta_j\mu_1\Psi(u_H;h_M)/4$, giving the claim with
$\kappa_{\mathrm{rot}}=\mu_1\Psi(u_H;h_M)/4$.
\end{proof}

\subsection{Proof of Proposition~\ref{rr:risk}}\label{app:risk}
Fix a rule and rate. History and next-index expectations condition on initialization, task and training set. Forcing expectations average a fresh pair.
By Fact~\ref{fact:gaussian-isometry}, \eqref{rr:first-order-risk} equals $\lim_{\delta\downarrow0}\delta^{-2}\mathcal R_\delta(W_0+\delta LA_0)$ for fixed $L,A_0$.

The first two matrix moments separate the risk increment into linear gain and quadratic noise cost.
\begin{lemma}[Exact affine risk calculation]\label{rr:affine}
For either rule, on $\mathcal E_\star$ from Lemma~\ref{er:training-set}, the first-order response satisfies uniformly for integers
$t\ge0$ with $s=t\eta\le s_{\max}$, for fixed $s_{\max}>0$,
\begin{equation}\label{rr:affine-risk-increment}
\E_{\mathrm{hist}}[\mathcal R^0(L_{t+1}^0)-\mathcal R^0(L_t^0)]
=-\eta\langle2EW_0A_0^\top,\bar G S^{1/2}\rangle_F
 +\frac{\eta^2}2\E_I\|\mathcal P(\widetilde G_I)\|_F^2+O(r\eta s e^{Cs}).
\end{equation}
\end{lemma}
\begin{proof}

Use the matrix moments $(m_t,V_t)$ of Lemma~\ref{er:moment-recursion}.
With $\mathcal P^*$ denoting the Frobenius adjoint and the trace taken on matrix space,
\begin{equation*}
\E_{\mathrm{hist}}\mathcal R^0(L_t^0)-\mathcal R^0(0)
=-\langle2EW_0A_0^\top,m_t\rangle_F+\tfrac12\tr(\mathcal P^*\mathcal P V_t).
\end{equation*}

This linear functional has norm at most $Cr$ in the normalized moment norm, since
$\|2EW_0A_0^\top\|_F\le C\sqrt r$ and $\|\mathcal P^*\mathcal P\|\le C$.
Apply the exact moment recursion: its affine forcing gives the first two terms of \eqref{rr:affine-risk-increment},
and the bounded generator contributes at most $Cr\eta\|(m_t,V_t)\|\le Cr\eta s e^{Cs}$.
\end{proof}

Choose fixed $s_{\max}\ge c_{\mathrm{risk}}u_H/(\gamma_k\mu_{1,r})$ for all sufficiently large $r$.
Then the lemma applies simultaneously to both rates throughout $t\le c_{\mathrm{risk}}qr$.

\begin{proof}[Proof of Proposition~\ref{rr:risk}]

Condition on Lemma~\ref{er:init}'s event. Gaussian forcing moments and H\"older give
\begin{equation*}
\begin{aligned}
\|y_\alpha\|_{L^8}&\le C\sqrt r,&
\|X_\alpha S^{1/2}\|_{L^8(F)}&\le C\sqrt{qr},\\
\E\|\mathcal P(\widetilde G_I)\|_F^4&\le Cq^2r^4,&
\E|\langle2EW_0A_0^\top,\widetilde G_I\rangle_F|^2&\le Cr^4.
\end{aligned}
\end{equation*}
For the last estimate, $\langle2EW_0A_0^\top,X_\alpha S^{1/2}\rangle_F$ is a centered quadratic form
whose coefficient has bounded norm and rank $O(r)$, hence $L^4$ norm $C\sqrt r$.
Its product with $y_\beta$ has $L^2$ norm at most $Cr$.
Independence of the $n$ stored pairs and Lemma~\ref{rr:source}, with the channel coefficients of Section~\ref{app:er}, imply
\begin{equation*}
\begin{aligned}
\frac{\langle2EW_0A_0^\top,\bar G_mS^{1/2}\rangle_F
 -4\sum_{\bullet\in\{a,o\}}\vartheta_{m,\bullet}\tr(S\Upsilon_\bullet)}r
 &=O_{L^2}(r/\sqrt n),\\
\frac{\E_I\|\mathcal P(\widetilde G_{m,I})\|_F^2}{2qr^2}
 -\frac{\tr(S^2)\nu_m}{r^2}
 &=O_{L^2}(n^{-1/2})+O(r^{-1}+r/q)=o(1).
\end{aligned}
\end{equation*}

Since $n\asymp qr$ and $q\gg r^2$, these errors vanish on a common event of probability $1-o(1)$
over initialization, task and training set. Lemma~\ref{rr:teacher} and $q\eta=u/(\gamma_k\mu_{1,r}r)$ give
\begin{equation*}
q\left\{-4\eta\sum_{\bullet\in\{a,o\}}\vartheta_{m,\bullet}\tr(S\Upsilon_\bullet)
 +\eta^2q\tr(S^2)\nu_m\right\}\longrightarrow p_m(u),
\end{equation*}
where $p_R,p_C$ are exactly \eqref{rr:risk-polynomials}.
Moreover $r\eta s e^{Cs}\le(C/q)t/(qr)$ throughout $0\le t\le c_{\mathrm{risk}}qr$.
Together with \eqref{rr:affine-risk-increment}, this proves the claimed uniform expansion.

The strict rate inequalities give $p_m(u_L),p_m(u_H)<0$.
The difference $p_m(u_H)-p_m(u_L)$ is a positive multiple of $(u_H-u_L)(u_H+u_L-T_m)$, hence is negative for $R$ and positive for $C$.
Choose $c_*>0$ such that
\begin{equation*}
4c_*\le\min\left\{\min_{m\in\{R,C\},\,j\in\{L,H\}}[-p_m(u_j)],\,
p_R(u_L)-p_R(u_H),\,p_C(u_H)-p_C(u_L)\right\}.
\end{equation*}
First choose $c_{\mathrm{risk}}$ small and then $r$ large so that the uniform increment error is at most $c_*/q$.
Each expected increment is then at most $-3c_*/q$, while the raw low-minus-high increment and centered high-minus-low increment
are each at least $2c_*/q$.
Summing from update $0$ to $t-1$, using $L_{m,j,0}^0=0$, proves all endpoint comparisons.
\end{proof}

\subsection{Standard tools}\label{app:tools}
The following facts use local generic variables and their stated dimensions.

\begin{fact}[Gaussian extreme singular values, {\citep[Corollary 5.35]{Vershynin2012}}]
\label{fact:gaussian-extremes}
For $k\ge r$, let $\mathsf G\in\R^{k\times r}$ have independent standard Gaussian entries.
For $u\ge0$,
\begin{equation*}
\Prob\{\sigma_{\max}(\mathsf G)>\sqrt k+\sqrt r+u\}
+\Prob\{\sigma_{\min}(\mathsf G)<\sqrt k-\sqrt r-u\}
\le2\exp(-u^2/2).
\end{equation*}
\end{fact}

\begin{fact}[Gaussian pairings, {\citep{Isserlis1918}}]\label{fact:isserlis}
For jointly centered Gaussian scalar variables $x_1,\ldots,x_{2s}$,
\begin{equation*}
\E\prod_{i=1}^{2s}x_i
=\sum_{\mathfrak p}\prod_{\{i,j\}\in\mathfrak p}\E[x_ix_j],
\end{equation*}
where the sum is over partitions into unordered pairs.
Odd joint moments vanish.
\end{fact}

\begin{fact}[Gaussian quadratic-form isometry and moments, {\citep{Isserlis1918,Janson1997}}]
\label{fact:gaussian-isometry}
For symmetric matrices $D,D'$ of equal dimension and an independent standard Gaussian vector $x$,
with $\Phi(x):=(xx^\top-I)/\sqrt2$,
\begin{equation*}
\E\langle D,\Phi(x)\rangle_F\langle D',\Phi(x)\rangle_F=\tr(DD').
\end{equation*}
Consequently, for the centered quadratic form $\phi_D(x):=x^\top Dx-\tr D$,
$\E\phi_D(x)=0$ and $\E\phi_D(x)^2=2\|D\|_F^2$.
For every $p\ge2$, Gaussian hypercontractivity gives
\begin{equation*}
\|\phi_D(x)\|_{L^p}\le\sqrt2(p-1)\|D\|_F.
\end{equation*}
\end{fact}

\begin{fact}[Hanson--Wright inequality, {\citep{RudelsonVershynin2013}}]
\label{fact:hanson-wright}
Let $x$ have independent centered entries of variance one with sub-Gaussian norms bounded by $b$.
For a fixed symmetric matrix $D$ and $u>0$,
\begin{equation*}
\Prob\{|x^\top Dx-\tr D|>u\}
\le2\exp\left[-c\min\left\{\frac{u^2}{b^4\|D\|_F^2},
\frac{u}{b^2\|D\|}\right\}\right].
\end{equation*}
\end{fact}

\begin{fact}[Gaussian hypercontractivity, {\citep[Chapters 5--6]{Janson1997}}]
\label{fact:hypercontractivity}
If $P$ is a polynomial of degree at most $d$ in independent standard Gaussian variables, then, for $p\ge2$,
\begin{equation*}
\|P\|_{L^p}\le(p-1)^{d/2}\|P\|_{L^2}.
\end{equation*}
Here $P$ denotes a polynomial.
The same estimate holds for a matrix-valued polynomial in the Frobenius norm.
\end{fact}

\begin{fact}[Matrix Bernstein, {\citep[Theorem 6.1(ii)]{Tropp2012}}]
\label{fact:matrix-bernstein}
Let $X_i$ be independent centered self-adjoint matrices of size $d$, with $\|X_i\|\le b$ almost surely.
Set $v:=\|\sum_i\E X_i^2\|$.
For $u>0$,
\begin{equation*}
\Prob\left\{\left\|\sum_iX_i\right\|>u\right\}
\le2d\exp\left\{-\frac{u^2}{2(v+bu/3)}\right\}.
\end{equation*}
\end{fact}

\begin{fact}[Matrix Chernoff, {\citep[Corollary 5.2]{Tropp2012}}]
\label{fact:matrix-chernoff}
Let $X_i$ be independent positive semidefinite matrices of size $d$, with $0\preceq X_i\preceq bI$.
Put $a_-:=\lambda_{\min}(\sum_i\E X_i)$ and $a_+:=\lambda_{\max}(\sum_i\E X_i)$.
For $0<u<1$,
\begin{equation*}
\Prob\left\{\lambda_{\min}\left(\sum_iX_i\right)\le(1-u)a_-\right\}
\le d\left[\frac{e^{-u}}{(1-u)^{1-u}}\right]^{a_-/b}.
\end{equation*}
For $u>0$,
\begin{equation*}
\Prob\left\{\lambda_{\max}\left(\sum_iX_i\right)\ge(1+u)a_+\right\}
\le d\left[\frac{e^u}{(1+u)^{1+u}}\right]^{a_+/b}.
\end{equation*}
\end{fact}

\begin{fact}[Independent-column Gram bound, {\citep[Theorem 3.1]{alpt2011}}]
\label{fact:independent-columns}
Write $\|Z\|_{\psi_1}=\inf\{a>0:\E\exp(|Z|/a)\le2\}$.
Let $x_1,\ldots,x_j\in\R^s$ be independent vectors whose scalar projections have uniformly bounded $\psi_1$ norms.
On the event $\max_i\|x_i\|\le C\sqrt s$, the off-diagonal Gram matrix satisfies
\begin{equation*}
\|((\langle x_i,x_{i'}\rangle\mathbf1_{\{i\ne i'\}}))_{i,i'=1}^j\|
\le Cs\sqrt{j/s}\log(e\sqrt{s/j})
\end{equation*}
outside an event of probability at most $C\exp(-c\sqrt j)$, provided $j/s$ is sufficiently small.
The failure probability of the maximum-norm event is added separately.
\end{fact}

\begin{fact}[Block Schur bound, {\citep{HornJohnson2013}}]\label{fact:block-schur}
For a block operator $K=(K_{i,j})$, one has
\begin{equation*}
\|K\|\le
\left\{\max_i\sum_j\|K_{i,j}\|\right\}^{1/2}
\left\{\max_j\sum_i\|K_{i,j}\|\right\}^{1/2}.
\end{equation*}
\end{fact}

\end{document}